\documentclass[11pt, logo, copyright, colorlinks=true, allcolors=blue]{nvidiatechreport}

\PassOptionsToPackage{numbers, sort&compress}{natbib}
\usepackage[utf8]{inputenc} 
\usepackage[T1]{fontenc}    
\usepackage{url}            
\usepackage{booktabs}       
\usepackage{amsfonts}       
\usepackage{nicefrac}       
\usepackage{microtype}      
\usepackage{xcolor}         
\usepackage{colortbl}       
\usepackage{multirow}       
\usepackage[numbers]{natbib}  

\usepackage{bbm}
\usepackage{pifont}
\usepackage{subcaption}
\usepackage{graphicx}
\usepackage{adjustbox}

\usepackage{amsmath,amssymb,amsthm}
\usepackage{enumitem}
\usepackage[most]{tcolorbox}
\usepackage{mathtools}
\usepackage{xspace}
\usepackage{dsfont}
\usepackage{fvextra}
\usepackage{hyperref}

\usepackage[ruled,linesnumbered]{algorithm2e} 
\usepackage{arydshln}                          

\usepackage[capitalize]{cleveref}
\crefname{section}{Sec.}{Secs.}
\Crefname{section}{Section}{Sections}
\crefname{table}{Tab.}{Tabs.}
\Crefname{table}{Table}{Tables}
\crefname{algocf}{Algorithm}{Algorithms}
\Crefname{algocf}{Algorithm}{Algorithms}
\crefname{appendix}{Appendix}{Appendices}
\Crefname{appendix}{Appendix}{Appendices}
\crefname{subappendix}{Appendix}{Appendices}
\Crefname{subappendix}{Appendix}{Appendices}

\graphicspath{{./}{figures/}{appendix/}}

\theoremstyle{definition}

\theoremstyle{remark}

\definecolor{remarkgreen}{HTML}{2d6a4f}
\definecolor{claimred}{HTML}{c25e3f}
\definecolor{takeawayblue}{HTML}{1f4e8a}
\definecolor{Red}{rgb}{0.768, 0.054, 0.054}
\definecolor{Blue}{rgb}{0.152, 0.294, 0.925}
\definecolor{Green}{rgb}{0,0.4,0.7}
\definecolor{darkgray}{gray}{0.3}
\definecolor{chocolate}{HTML}{D2691E}
\definecolor{maroon}{HTML}{A00000}
\definecolor{indigo}{HTML}{4B0082}
\definecolor{newred}{HTML}{a91e1e}
\definecolor{cadmiumgreen}{rgb}{0.0, 0.42, 0.24}
\definecolor{forestgreen}{rgb}{0.13, 0.55, 0.13}
\definecolor{deltateal}{HTML}{0F766E}
\definecolor{mygray}{gray}{0.92}
\definecolor{myblue}{HTML}{E8F0FB}
\definecolor{linkcolor}{RGB}{190,105,30}

\definecolor{colorful}{RGB}{232,245,233}
\definecolor{posgain}{RGB}{30,128,55}
\definecolor{neggain}{RGB}{195,40,40}
\newcommand{\dpos}[1]{\textcolor{posgain}{$\boldsymbol{+#1}$}}
\newcommand{\dneg}[1]{\textcolor{neggain}{$\boldsymbol{-#1}$}}
\newcommand{\dzero}{$\boldsymbol{0.0}$}

\definecolor{nvidiagreen}{HTML}{76B900}
\definecolor{nvgreen}{RGB}{118,185,0}        
\definecolor{nvgreendark}{RGB}{95,150,0}     
\definecolor{nvgreenpastel}{RGB}{236,248,218} 
\definecolor{findingframe}{RGB}{95,150,0}
\definecolor{findingback}{RGB}{236,248,218}

\newtcolorbox{finding}[1]{%
  enhanced,
  breakable,
  colback=findingback,
  colframe=findingframe,
  boxrule=0.5pt,
  arc=2pt,
  left=8pt, right=8pt, top=4pt, bottom=5pt,
  fonttitle=\bfseries\footnotesize,
  coltitle=white,
  colbacktitle=findingframe,
  title={#1},
  before skip=4pt, after skip=4pt,
}
\definecolor{promptframe}{RGB}{118,185,0}
\definecolor{promptback}{RGB}{236,248,218}
\newtcolorbox{prompttemplate}{%
  enhanced,
  breakable,
  colback=promptback,
  colframe=promptframe,
  boxrule=0.4pt,
  arc=2pt,
  left=8pt, right=8pt, top=4pt, bottom=4pt,
  before skip=4pt, after skip=4pt,
}
\newtcolorbox{prompttemplatefig}{%
  enhanced,
  colback=promptback,
  colframe=promptframe,
  boxrule=0.4pt,
  arc=2pt,
  left=8pt, right=8pt, top=2pt, bottom=2pt,
  before skip=2pt, after skip=2pt,
}
\definecolor{qframe}{RGB}{160,160,160}
\definecolor{qback}{RGB}{246,246,246}
\definecolor{negframe}{RGB}{195,80,80}
\definecolor{negback}{RGB}{253,234,234}
\newtcolorbox{questionbox}{%
  enhanced,
  colback=qback,
  colframe=qframe,
  boxrule=0.4pt,
  arc=2pt,
  left=8pt, right=8pt, top=2pt, bottom=2pt,
  before skip=2pt, after skip=2pt,
}
\newtcolorbox{negcandidatebox}{%
  enhanced,
  colback=negback,
  colframe=negframe,
  boxrule=0.4pt,
  arc=2pt,
  left=8pt, right=8pt, top=2pt, bottom=2pt,
  before skip=2pt, after skip=2pt,
}

\hypersetup{
    colorlinks=true,
    citecolor=teal,
    linkcolor=nvidiagreen,  
    urlcolor=Green,
}

\AtBeginDocument{%
}

\title{Zone of Proximal Policy Optimization:\\ Teacher in Prompts, Not Gradients}

\newcommand{\authorblock}{%
  Byung-Kwan Lee$^{\dagger}$,
  Ximing Lu,
  Shizhe Diao,
  Minki Kang,
  Saurav Muralidharan,
  Karan Sapra,\\
  Andrew Tao,
  Yu-Chiang Frank Wang,
  Ryo Hachiuma%
}
\author{\authorblock}
\correspondingauthor{$^\dagger$Project Lead, \texttt{byungkwanl@nvidia.com}}

\renewcommand{\titlefont}{\color{nvidiagreen}\normalfont\bfseries\fontsize{17}{20}\selectfont}

\makeatletter
\renewcommand{\maketitle}{\bgroup\setlength{\parindent}{0pt}
  \begin{adjustwidth}{0pt}{24pt}
  \begin{flushleft}
    {\raggedright \titlefont \@title\par}%
    \vskip11pt
    {\raggedright \normalfont\bfseries\fontsize{9}{13}\selectfont \authorblock\par}%
    \vskip12pt
    {\normalfont\bfseries\fontsize{9}{13}\selectfont NVIDIA\par}%
    \vskip16pt%
  \end{flushleft}
  \end{adjustwidth}
  \egroup
  \thispagestyle{firststyle}
}
\makeatother

\begin{abstract}
    Knowledge distillation transfers a teacher's competence to a small student but is brittle in the small-student regime: forcing the student to imitate logits from a much larger teacher hurts generalization on benchmark families beyond the training corpus. Reinforcement learning (RL) avoids logit imitation by training on the student's own rollouts. However, questions on which every rollout fails yield zero advantage and are silently discarded, and injecting a stronger teacher's response into the policy gradient to recover them breaks the on-policy assumption and induces drift. We introduce \textbf{Zone of Proximal Policy Optimization} (ZPPO), inspired by Vygotsky's \emph{zone of proximal development}. ZPPO keeps the teacher inside the \emph{prompt} rather than the policy gradient. On hard questions, where the student's mean rollout accuracy is below half, ZPPO constructs two reformulated prompts. A \emph{Binary Candidate-included Question} (BCQ) pairs one correct teacher response with one incorrect student response as anonymized candidates the student uses as references. A \emph{Negative Candidate-included Question} (NCQ) aggregates the student's wrong rollouts into a single prompt to surface their shared failure modes. A prompt replay buffer keeps each hard question eligible for re-sampling until it either graduates (the student's mean rollout accuracy on it reaches half or more) or is FIFO-evicted under finite capacity, amplifying BCQ and NCQ inside the student's current zone of proximal development. We post-train Qwen3.5 students at four scales (0.8B--9B) as VLMs with a 27B teacher and evaluate them on a 31-benchmark suite (16 VLM, 10 LLM, 5 Video); ZPPO outperforms off/on-policy distillation and GRPO, with the largest gains at the smallest scale. Our project page is available at \href{https://byungkwanlee.github.io/ZPPO-page/}{link}.
\end{abstract}

\begin{document}

\maketitle

\begin{center}
    \includegraphics[width=\linewidth]{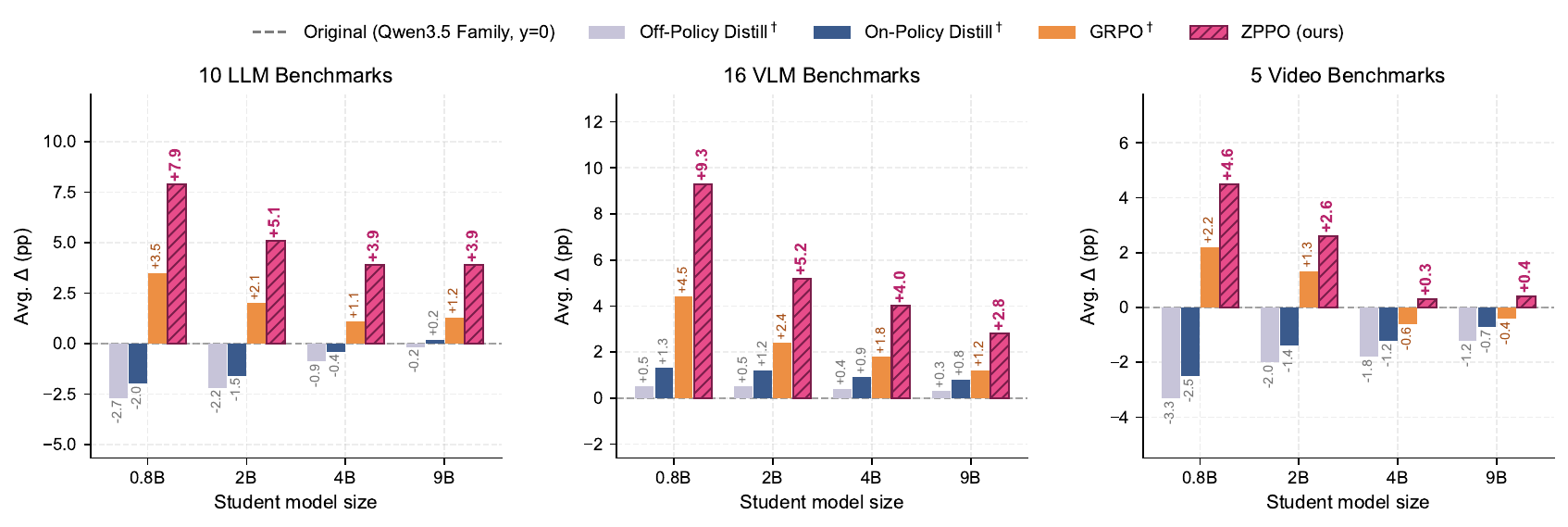}
    \vspace{-3mm}
    \captionof{figure}{Average gain $\Delta$ (pp) on $10$ LLM (left), $16$ VLM (middle), and $5$ Video benchmarks (right), across the four Qwen3.5 student scales. All methods are trained on the same multimodal dataset for vision-language models (VLMs) and evaluated on various benchmarks for generalization. ${\dagger}$ denotes augmentation with the prompt replay buffer.}
    \label{fig:teaser}
\end{center}

\vskip6pt
\section*{Abstract}
{\abscontent}

\section{Introduction}
\label{sec:intro}

\begin{figure*}[t!]
    \vspace{-3mm}
    \centering
    \includegraphics[width=\linewidth]{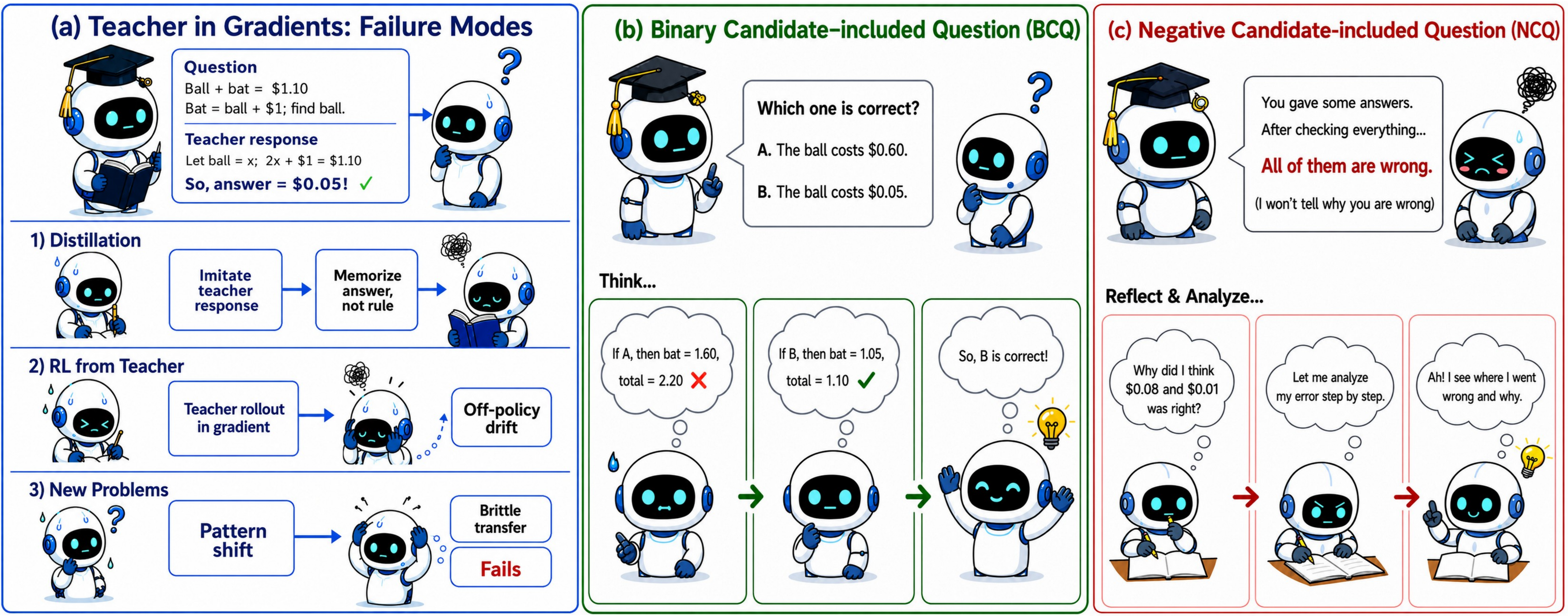}
    \vspace{-5mm}
    \caption{
    (a)~Failure modes that motivate ZPPO. In the small-student regime, matching a much larger teacher's logits is brittle; on the RL side, injecting a teacher response into the policy gradient as if it were on-policy introduces drift.
    (b)~BCQ recovers a learning signal on hard questions by pairing a correct teacher response with a wrong student response as anonymized in-prompt candidates, asking the student to solve the problem using them as references.
    (c)~NCQ aggregates the student's wrong rollouts into a single prompt, surfacing shared errors.}
    \label{fig:concept}
\end{figure*}

Large language models (LLMs) and vision--language models (VLMs) have made rapid progress~\citep{hurst2024gpt,team2023gemini,claude3series2024,qwen3.5,guo2025deepseek,nvidia2026nemotron35} on broad benchmarks, with reinforcement learning post-training~\citep{guo2025deepseek,team2025kimi,liu2025understanding,yu2025dapo,he2025justrl,khatri2025art}. The cost of those gains, however, has scaled with the cost of the model: tens of billions of activated parameters. Models of that size do not fit the regime that matters for deployment on mobile phones, AR/VR glasses, or embodied robots~\citep{Lee_2025_CVPR,NEURIPS2025_e5849736,Lee_2026_CVPR_Masters,marafioti2025smolvlm,vasu2025fastvlm,qu2025mobile}, where compute and memory budgets are much smaller.

Closing this deployment gap requires \emph{small} models that generalize broadly, not narrow specialists. The standard knowledge-transfer recipe is \emph{knowledge distillation}: a small student is trained to imitate the logits, hidden states, or sampled outputs of a strong teacher~\citep{hinton2015distilling,gou2021knowledge,sanh2019distilbert,gu2024minillm,ko2024distillm,wu2024rethinking}. Despite its effectiveness, distillation carries a structural limitation that becomes severe when the student is much smaller than the teacher. Logit imitation is, at its core, supervised fitting on the teacher's distribution: with limited training data, students memorize teacher answers and forget general capabilities~\citep{chu2025sft,chen2025sft,huan2025does}. This limitation grows with the teacher--student capacity gap~\citep{cho2019efficacy,mirzadeh2020improved,zhang-etal-2023-lifting,zhang2023towards}, especially for students of $3$B parameters or fewer~\citep{li2025small,ren2026rethinking}. The result is a student whose generalization on benchmark families beyond the training corpus is brittle, and whose responses rarely venture beyond the paths the teacher has already taken. Our experiments reproduce this pattern (\cref{sec:experiments:main}).

A line of work tries to alleviate these issues by using the student's own rollouts to conduct distillation from the teacher to the student (\emph{on-policy} distillation~\citep{agarwal2024policy,lu2025onpolicydistillation,fu2026revisiting,bousselham2025vold,yang2026learning,zhang2026fast,wu2026lightning}) or by removing the external teacher and letting the student teach itself (\emph{self}-distillation~\citep{shenfeld2026self,zhao2026selfdistilled,hubotter2026reinforcement,yang2026self}). These reduce the teacher--student mismatch, but most share the same fundamental constraint: the training signal is still a logit distribution that the student must match~\citep{li2026rethinking,song2026survey}. For 0.8B or 2B students, model capacity is too limited to absorb such a distribution broadly~\citep{zhang-etal-2023-lifting,zhang2023towards,li2025small}, and the resulting models lose generalization beyond the training corpus~\citep{kim2026does,li2023distilling}.

A natural alternative is reinforcement learning (RL)~\citep{guo2025deepseek,shao2024deepseekmath,hu2025reinforce++,yu2025dapo,liu2025understanding,he2025justrl,khatri2025art,liu2025prorl,hu2025brorl,liu2025dler,liu2026gdpo}, which avoids logit imitation entirely and trains directly on the student's own rollouts. However, RL has its own pathology in the small-model regime: hard questions with zero successful rollouts produce zero advantage and are silently discarded, so the student never learns from the questions where it most needs help. Simply using a stronger teacher's correct response as if it were the student's breaks the on-policy assumption and induces policy drift, because the teacher's response lies far outside the student's rollout distribution. \Cref{fig:concept}a summarizes these two failure modes.

In this paper, we ask a single question: \emph{how can we transfer the teacher's knowledge to the student without ever imitating the teacher's logits or injecting the teacher's response directly into the student's policy gradient?} Our answer, which we call \textbf{Zone of Proximal Policy Optimization} (ZPPO), borrows its name from Vygotsky's \emph{zone of proximal development}~\citep{vygotsky1978mind}, the band of tasks a learner cannot yet solve alone but \emph{can} solve with a small amount of guidance from a more capable peer. Concretely, ZPPO keeps the teacher only inside the \emph{prompt}, never inside the student's policy gradient (\cref{fig:concept}b,c and \cref{fig:method}):

\begin{figure*}[t!]
    \vspace{-3mm}
    \centering
    \includegraphics[width=\linewidth]{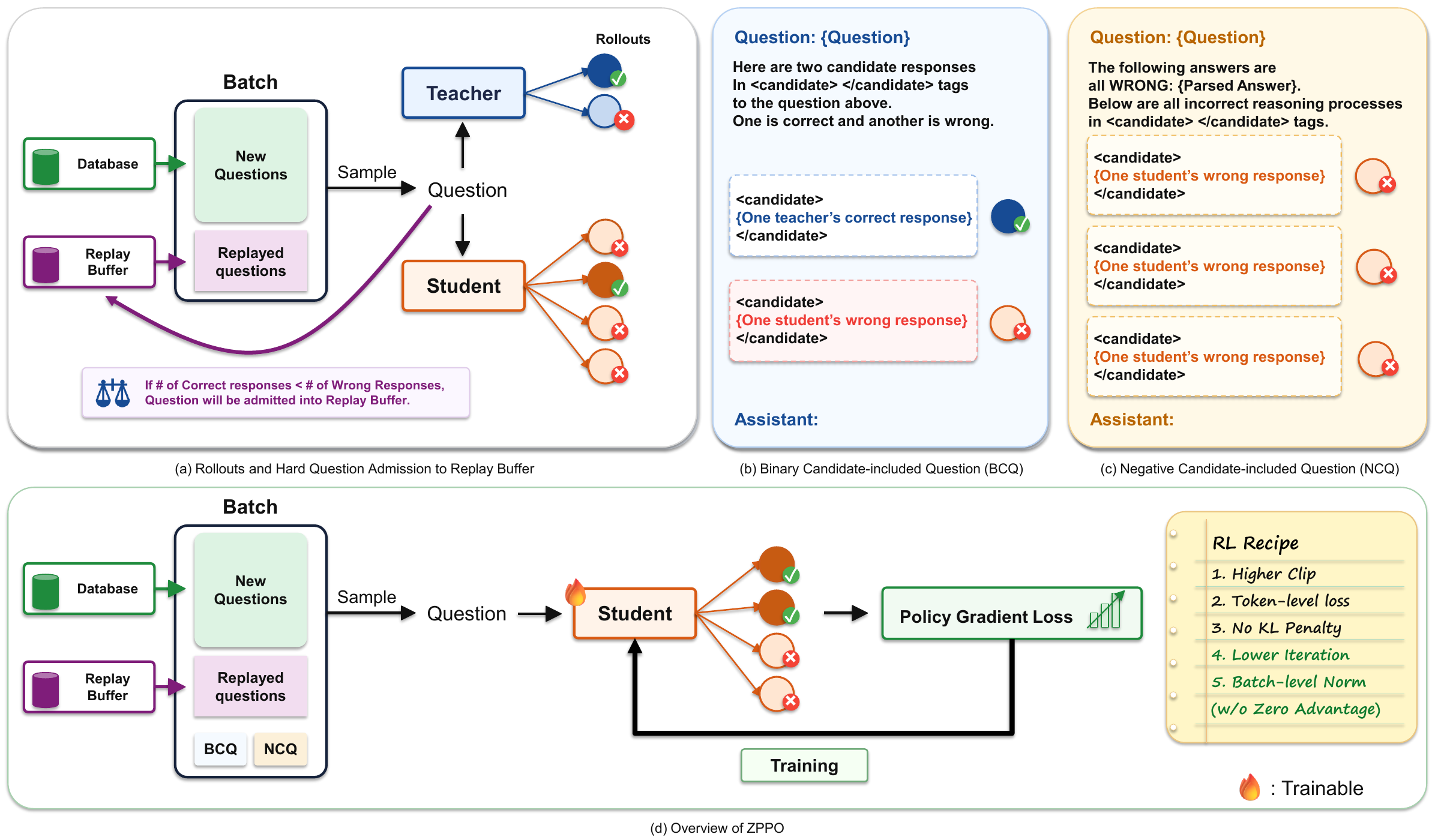}
    \vspace{-5mm}
    \caption{\textbf{Overview of ZPPO.}
    (a)~Hard questions (mean rollout accuracy below half) are admitted to the prompt replay buffer.
    (b)~BCQ pairs one correct teacher response with one wrong student response.
    (c)~NCQ aggregates the student's wrong rollouts into a single prompt.
    (d)~Within one step, the plain groups and their BCQ and NCQ groups enter the same policy gradient update, with the RL recipe.}
    \label{fig:method}
\end{figure*}

\begin{itemize}[leftmargin=12pt,topsep=2pt,itemsep=2pt]
  \item For each hard question (defined as one where the student's mean rollout accuracy is below half), we construct a \emph{Binary Candidate-included Question} (BCQ), a reformulated prompt that appends one correct teacher response and one wrong student response as shuffled, anonymized candidates, and asks the student to solve the problem using them as references. Since the student generates its own response, its policy gradient remains on-policy at the response level.
  \item In parallel, we construct a \emph{Negative Candidate-included Question} (NCQ) that appends \emph{all} of the student's wrong responses, explicitly showing that all of them are wrong. In a standard rollout group each wrong rollout contributes its advantage independently, so NCQ is the place where a question's failed attempts become collectively visible, cueing the student to recognize the shared error patterns and avoid them.
  \item Hard questions are also pushed into a \emph{prompt} replay buffer (no responses are stored). They are re-sampled until they either \emph{graduate} (mean rollout accuracy reaches half, i.e.\ $\bar{r}_x\!\geq\!0.5$) or are FIFO-evicted once the buffer reaches capacity (\cref{sec:method:buffer}, \cref{sec:appendix:zone}). This focuses repeated practice on prompts inside the student's zone of proximal development.
\end{itemize}

These three components (BCQ, NCQ, and prompt replay buffer) are layered on top of GRPO~\citep{shao2024deepseekmath} augmented with three DAPO ingredients~\citep{yu2025dapo}: clip-higher, a token-level policy gradient loss, and no KL penalty. We also adopt two simple but easily mishandled recipe choices that materially affect a small model's training dynamics: \emph{iterations per step} and \emph{batch-level advantage normalization} of REINFORCE++~\citep{hu2025reinforce++}, with zero-advantage groups excluded from the batch statistics. Both are low-cost, yet setting either one incorrectly lowers accuracy (\cref{fig:recipe}).

We instantiate ZPPO on the Qwen3.5 family at four student scales (0.8B, 2B, 4B, and 9B) with a 27B teacher~\citep{qwen3.5}, post-train each student as a VLM on a multimodal RL dataset of image--question pairs, and evaluate on a $31$-benchmark suite spanning 16 VLM, 10 LLM, and 5 Video benchmarks (\cref{sec:appendix:benchmarks}). At every scale, ZPPO outperforms (i)~off- and on-policy distillation and (ii)~GRPO (\cref{fig:teaser}), and ablating any single component (BCQ, NCQ, or the prompt replay buffer) lowers every family's average at every scale (\cref{tab:ablation}, \cref{sec:appendix:main_tables}). The improvements are largest precisely where the small-model gap is widest: measured against the base student, a 0.8B student gains $+9.3$~pp on the VLM benchmarks, while a 9B student gains $+2.8$~pp. Crucially, ZPPO also improves on the LLM and Video benchmark families that lie beyond the training corpus, the very families where distillation hurts.

\begin{table*}[t!]
    \vspace{-3mm}
    \centering
    \caption{Main results on the 16 VLM benchmarks at 0.8B and 2B; $^{\dagger}$ denotes augmentation with the prompt replay buffer, and $\Delta$ reports ZPPO minus the strongest non-ZPPO row in each column on the same model size. Bold marks the column maximum and is shared by every row that attains it; the colored $\Delta$ row is a difference, not a maximum. 4B and 9B results: \cref{sec:appendix:main_tables}. Benchmark abbreviations: \cref{sec:appendix:benchmarks}.}
    \label{tab:main}
    \resizebox{\linewidth}{!}{
    \setlength{\tabcolsep}{1mm}
    \renewcommand{\arraystretch}{1.05}
    \begin{tabular}{l*{17}{c}}
    \toprule
    16 VLM Benchmarks
    & AI2D
    & BabyV
    & CharXiv
    & DynaM
    & EmbSp
    & InfoVQA
    & MVerse
    & MVision
    & MVista
    & MMMU$^\text{Pro}$
    & MM-Vet
    & OCR$^\text{EN}$
    & OCR$^\text{ZH}$
    & VisP
    & VBlind
    & WeMath
    & Avg \\
    \midrule
    Qwen3.5-0.8B & 65.6 & 6.7 & 54.3 & 17.8 & 67.9 & 68.6 & 43.5 & 16.4 & 60.7 & 26.8 & 53.2 & 40.0 & 17.0 & 20.5 & 42.8 & 54.4 & \cellcolor{gray!10}41.0 \\
    \cdashline{1-18}\noalign{\vskip 0.5ex}
    $+$ Off-Distill & 66.3 & 6.7 & 55.1 & 19.4 & 65.8 & 66.0 & 45.3 & 17.3 & 62.0 & 27.6 & 51.5 & 38.9 & 16.2 & 20.0 & 41.8 & 55.9 & \cellcolor{gray!10}41.0 \\
    $+$ Off-Distill$^{\dagger}$ & 66.7 & 6.7 & 56.0 & 20.1 & 65.8 & 66.8 & 45.8 & 17.9 & 62.2 & 29.0 & 51.8 & 39.5 & 15.9 & 20.8 & 42.2 & 56.4 & \cellcolor{gray!10}41.5 \\
    $+$ On-Distill & 67.6 & 7.5 & 56.0 & 20.0 & 66.7 & 67.1 & 45.8 & 19.4 & 62.7 & 28.2 & 52.7 & 39.6 & 16.7 & 21.2 & 41.5 & 57.7 & \cellcolor{gray!10}41.9 \\
    $+$ On-Distill$^{\dagger}$ & 68.6 & 7.8 & 56.4 & 21.1 & 67.1 & 67.6 & 47.4 & 18.7 & 63.6 & 28.8 & 52.5 & 39.1 & 16.7 & 20.9 & 42.3 & 58.6 & \cellcolor{gray!10}42.3 \\
    \cdashline{1-18}\noalign{\vskip 0.5ex}
    $+$ GRPO & 69.5 & 8.6 & 57.0 & 21.3 & 69.2 & 71.0 & 47.7 & 20.1 & 65.2 & 29.9 & 55.6 & 40.9 & 17.0 & 24.2 & 43.2 & 60.6 & \cellcolor{gray!10}43.8 \\
    $+$ GRPO$^{\dagger}$ & 71.2 & 9.8 & 59.9 & 23.6 & 69.4 & 72.4 & 51.1 & 20.9 & 68.3 & 30.5 & 57.5 & 41.3 & 17.5 & 27.8 & 43.6 & 62.5 & \cellcolor{gray!10}45.5 \\
    \rowcolor{colorful}
    $+$ ZPPO & \textbf{76.5} & \textbf{13.9} & \textbf{63.9} & \textbf{31.1} & \textbf{71.5} & \textbf{75.3} & \textbf{59.3} & \textbf{29.2} & \textbf{73.2} & \textbf{37.6} & \textbf{59.9} & \textbf{42.5} & \textbf{18.7} & \textbf{35.0} & \textbf{44.7} & \textbf{71.7} & \cellcolor{gray!10}\textbf{50.3} \\
    \cdashline{1-18}\noalign{\vskip 0.3ex}
    $\Delta$ & \dpos{5.3} & \dpos{4.1} & \dpos{4.0} & \dpos{7.5} & \dpos{2.1} & \dpos{2.9} & \dpos{8.2} & \dpos{8.3} & \dpos{4.9} & \dpos{7.1} & \dpos{2.4} & \dpos{1.2} & \dpos{1.2} & \dpos{7.2} & \dpos{1.1} & \dpos{9.2} & \cellcolor{gray!10}\dpos{4.8} \\
    \midrule
    Qwen3.5-2B & 81.9 & 11.6 & 71.6 & 41.1 & 78.2 & 81.2 & 69.7 & 38.4 & 78.6 & 46.2 & 69.7 & 44.7 & 24.0 & 38.3 & 55.2 & 77.9 & \cellcolor{gray!10}56.8 \\
    \cdashline{1-18}\noalign{\vskip 0.5ex}
    $+$ Off-Distill & 82.5 & 11.9 & 71.0 & 43.2 & 77.2 & 80.2 & 70.8 & 40.4 & 77.9 & 47.4 & 69.2 & 44.4 & 23.0 & 38.4 & 53.7 & 79.5 & \cellcolor{gray!10}56.9 \\
    $+$ Off-Distill$^{\dagger}$ & 82.9 & 12.5 & 71.4 & 43.7 & 77.4 & 80.4 & 71.4 & 41.8 & 77.8 & 47.9 & 69.2 & 44.5 & 23.7 & 37.8 & 54.6 & 79.3 & \cellcolor{gray!10}57.3 \\
    $+$ On-Distill & 82.6 & 13.1 & 71.7 & 44.5 & 77.6 & 81.1 & 71.9 & 41.9 & 78.2 & 48.8 & 69.2 & 45.3 & 23.9 & 38.8 & 54.9 & 79.3 & \cellcolor{gray!10}57.7 \\
    $+$ On-Distill$^{\dagger}$ & 83.4 & 12.8 & 71.7 & 44.9 & 77.5 & 80.9 & 72.0 & 42.7 & 77.9 & 49.3 & 69.4 & 45.5 & 24.5 & 38.8 & 55.6 & 80.6 & \cellcolor{gray!10}58.0 \\
    \cdashline{1-18}\noalign{\vskip 0.5ex}
    $+$ GRPO & 83.5 & 13.7 & 72.8 & 45.6 & 78.3 & 82.4 & 72.3 & 43.1 & 79.0 & 49.3 & 72.3 & 46.1 & 24.9 & 39.2 & 56.5 & 80.4 & \cellcolor{gray!10}58.7 \\
    $+$ GRPO$^{\dagger}$ & 83.8 & 14.4 & 73.0 & 45.9 & 78.7 & 83.1 & 72.8 & 43.4 & 79.3 & 49.6 & 74.0 & 46.4 & 25.2 & 40.3 & 57.0 & 80.7 & \cellcolor{gray!10}59.2 \\
    \rowcolor{colorful}
    $+$ ZPPO & \textbf{85.3} & \textbf{18.6} & \textbf{73.9} & \textbf{52.7} & \textbf{79.5} & \textbf{84.6} & \textbf{76.0} & \textbf{50.5} & \textbf{80.5} & \textbf{53.2} & \textbf{77.1} & \textbf{48.8} & \textbf{26.0} & \textbf{42.0} & \textbf{60.5} & \textbf{82.6} & \cellcolor{gray!10}\textbf{62.0} \\
    \cdashline{1-18}\noalign{\vskip 0.3ex}
    $\Delta$ & \dpos{1.5} & \dpos{4.2} & \dpos{0.9} & \dpos{6.8} & \dpos{0.8} & \dpos{1.5} & \dpos{3.2} & \dpos{7.1} & \dpos{1.2} & \dpos{3.6} & \dpos{3.1} & \dpos{2.4} & \dpos{0.8} & \dpos{1.7} & \dpos{3.5} & \dpos{1.9} & \cellcolor{gray!10}\dpos{2.8} \\
    \bottomrule
    \end{tabular}
    }
\end{table*}

\begin{table*}[t!]
    \vspace{-3mm}
    \centering
    \caption{Evaluation over the 10 LLM and 5 Video benchmarks (probing generalization beyond the training data). $\Delta$ is ZPPO minus the strongest non-ZPPO row per column. Bold marks the column maximum, shared by every row attaining it; the colored $\Delta$ row is a difference. 4B and 9B: \cref{sec:appendix:main_tables}. Benchmark abbreviations: \cref{sec:appendix:benchmarks}.}
    \label{tab:generalization}
    \resizebox{\linewidth}{!}{
    \setlength{\tabcolsep}{1mm}
    \renewcommand{\arraystretch}{1.05}
    \begin{tabular}{lccccccccccccccccc}
    \toprule
    & \multicolumn{11}{c}{10 LLM Benchmarks} & \multicolumn{6}{c}{5 Video Benchmarks} \\
    \cmidrule(lr){2-12}\cmidrule(lr){13-18}
    Method & AIME25 & AIME26 & CEval & GPQA-D & HLE & IMO-AB & MMLU & MMLU-Pro & MMLU-Rd & MultiCh & Avg & MMVU & MVBench & VMME & VMME$^\text{S}$ & VMMMU & Avg \\
    \midrule
    Qwen3.5-0.8B & \textbf{3.3} & 0.0 & 52.3 & 12.1 & 2.2 & 3.3 & 57.7 & 41.7 & 62.5 & 16.6 & \cellcolor{gray!10}25.2 & 39.0 & 51.6 & 50.3 & 51.6 & 48.8 & \cellcolor{gray!10}48.3 \\
    \cdashline{1-18}\noalign{\vskip 0.5ex}
    $+$ Off-Distill$^{\dagger}$ & 0.0 & 0.0 & 47.9 & 10.1 & 2.1 & 2.9 & 52.6 & 38.3 & 56.9 & 14.7 & \cellcolor{gray!10}22.5 & 35.4 & 48.9 & 46.8 & 48.5 & 45.4 & \cellcolor{gray!10}45.0 \\
    $+$ On-Distill$^{\dagger}$ & 0.0 & 0.0 & 49.2 & 10.6 & 2.8 & 4.0 & 53.8 & 38.8 & 58.2 & 14.9 & \cellcolor{gray!10}23.2 & 36.8 & 49.2 & 47.8 & 49.6 & 45.7 & \cellcolor{gray!10}45.8 \\
    $+$ GRPO$^{\dagger}$ & \textbf{3.3} & 0.0 & 54.5 & 25.5 & 3.5 & 7.0 & 59.9 & 47.1 & 65.1 & 20.8 & \cellcolor{gray!10}28.7 & 41.3 & 53.5 & 52.2 & 53.0 & 52.5 & \cellcolor{gray!10}50.5 \\
    \rowcolor{colorful}
    $+$ ZPPO & \textbf{3.3} & 0.0 & \textbf{57.6} & \textbf{42.4} & \textbf{5.0} & \textbf{10.5} & \textbf{63.0} & \textbf{51.7} & \textbf{69.0} & \textbf{28.6} & \cellcolor{gray!10}\textbf{33.1} & \textbf{42.5} & \textbf{56.6} & \textbf{53.9} & \textbf{54.9} & \textbf{56.4} & \cellcolor{gray!10}\textbf{52.9} \\
    \cdashline{1-18}\noalign{\vskip 0.3ex}
    $\Delta$ & \dzero & \dzero & \dpos{3.1} & \dpos{16.9} & \dpos{1.5} & \dpos{3.5} & \dpos{3.1} & \dpos{4.6} & \dpos{3.9} & \dpos{7.8} & \cellcolor{gray!10}\dpos{4.4} & \dpos{1.2} & \dpos{3.1} & \dpos{1.7} & \dpos{1.9} & \dpos{3.9} & \cellcolor{gray!10}\dpos{2.4} \\
    \midrule
    Qwen3.5-2B & \textbf{26.7} & 30.0 & 71.1 & 53.5 & 3.5 & 11.3 & 74.6 & 66.2 & 80.4 & 35.8 & \cellcolor{gray!10}45.3 & 53.7 & 60.7 & 59.9 & 61.2 & 67.6 & \cellcolor{gray!10}60.6 \\
    \cdashline{1-18}\noalign{\vskip 0.5ex}
    $+$ Off-Distill$^{\dagger}$ & \textbf{26.7} & 26.7 & 68.2 & 50.3 & 2.0 & 10.9 & 71.2 & 63.8 & 77.5 & 34.0 & \cellcolor{gray!10}43.1 & 51.5 & 59.2 & 57.9 & 59.0 & 65.4 & \cellcolor{gray!10}58.6 \\
    $+$ On-Distill$^{\dagger}$ & 23.3 & 30.0 & 69.3 & 51.9 & 3.8 & 11.5 & 72.2 & 63.8 & 77.9 & 34.0 & \cellcolor{gray!10}43.8 & 52.4 & 59.9 & 58.6 & 59.4 & 65.5 & \cellcolor{gray!10}59.2 \\
    $+$ GRPO$^{\dagger}$ & \textbf{26.7} & 33.3 & 72.4 & 56.5 & 4.2 & 19.3 & 75.4 & 67.3 & 81.4 & 37.3 & \cellcolor{gray!10}47.4 & 56.7 & 61.7 & 61.2 & 62.6 & 67.3 & \cellcolor{gray!10}61.9 \\
    \rowcolor{colorful}
    $+$ ZPPO & \textbf{26.7} & \textbf{40.0} & \textbf{74.0} & \textbf{60.6} & \textbf{5.1} & \textbf{29.5} & \textbf{76.4} & \textbf{68.2} & \textbf{82.9} & \textbf{40.6} & \cellcolor{gray!10}\textbf{50.4} & \textbf{59.0} & \textbf{62.7} & \textbf{62.4} & \textbf{63.9} & \textbf{68.1} & \cellcolor{gray!10}\textbf{63.2} \\
    \cdashline{1-18}\noalign{\vskip 0.3ex}
    $\Delta$ & \dzero & \dpos{6.7} & \dpos{1.6} & \dpos{4.1} & \dpos{0.9} & \dpos{10.2} & \dpos{1.0} & \dpos{0.9} & \dpos{1.5} & \dpos{3.3} & \cellcolor{gray!10}\dpos{3.0} & \dpos{2.3} & \dpos{1.0} & \dpos{1.2} & \dpos{1.3} & \dpos{0.5} & \cellcolor{gray!10}\dpos{1.3} \\
    \bottomrule
    \end{tabular}
    }
\end{table*}

\begin{finding}{\textsc{Distillation hurts generalization beyond the training corpus; ZPPO does not}}
  \small
  \textcolor{OliveGreen}{\textbf{16 VLM benchmarks (training-aligned):}} Distillation lifts the student only marginally: averaging Off-Distill$^{\dagger}$ and On-Distill$^{\dagger}$ gives $+0.9$/$+0.9$/$+0.7$/$+0.6$~pp at $0.8$B/$2$B/$4$B/$9$B (\cref{tab:main}, \cref{sec:appendix:main_tables}), while ZPPO gains $+9.3$/$+5.2$/$+4.0$/$+2.8$~pp over the untrained base.
  
  \textcolor{OliveGreen}{\textbf{10 LLM and 5 Video benchmarks (beyond the training corpus):}} Distillation \emph{degrades} the same students by $-2.5$/$-1.8$/$-0.9$/$-0.3$~pp (Off-Distill$^{\dagger}$ and On-Distill$^{\dagger}$ averaged; \cref{tab:generalization}, \cref{sec:appendix:benchmarks,sec:appendix:main_tables}), whereas ZPPO \emph{improves} them by $+6.8$/$+4.3$/$+2.7$/$+2.7$~pp.
\end{finding}

\section{Related Work}
\label{sec:related}

Four lines of work bear on ZPPO. \emph{Knowledge distillation}~\citep{gou2021knowledge,agarwal2024policy} turns the teacher's output distribution into the signal the student must match. That signal grows brittle as the teacher--student gap widens~\citep{cho2019efficacy,mirzadeh2020improved,li2025small}, inducing memorization of teacher answers and forgetting of general capabilities~\citep{chu2025sft,chen2025sft,huan2025does}. \emph{RL post-training}~\citep{shao2024deepseekmath,yu2025dapo,guo2025deepseek} trains on the student's own rollouts, but a group of all-wrong rollouts has zero advantage and gives no learning signal. Methods that target such groups reshape advantages~\citep{le2026no,feng2025don}, interleave supervised updates~\citep{ma2026learning,han2026distill}, or feed the teacher back into the gradient through hybrid RL$+$distillation objectives~\citep{xu2025kdrl,xu2026rlkd} or teacher traces added to the group~\citep{NEURIPS2025_a9d5c33e}. All of them change the loss or the target, never the prompt. \emph{Curricula and prompt-side guidance} do act on the prompt. Curricula~\citep{zhang2025clpo,mahrooghi2026goldilocks,zhang2025learninglike,li2026adacurl} reschedule or simplify hard prompts. Hints~\citep{nath2025adaptive,chen2026nudging,wang2026don,xia2026learning} supply a correct cue but no wrong alternative. Partial solutions, or prefixes~\citep{zhang2025bread,zhang2025stephint,li2025staying,huang2025blending}, hand over a reference solution's opening steps. \emph{Learning from mistakes} fine-tunes on corrected solutions, or on correct and wrong answers labeled by an input tag~\citep{an2023learning,wang2024learning,wu2026enhancing}. SocraticPO~\citep{liu2026socraticpo} lets the student continue a wrong attempt after teacher guidance, under a decayed reward. ZPPO instead reformulates the question and samples a fresh group from the student's own policy under the full binary reward. Its replay buffer then amplifies BCQ and NCQ inside the student's zone of proximal development~\citep{vygotsky1978mind}. \cref{sec:related_full} gives the full version.
\section{Zone of Proximal Policy Optimization}
\label{sec:method}

ZPPO is built on top of GRPO~\citep{shao2024deepseekmath} (other policy objectives in \cref{sec:appendix:backbones}). \cref{sec:appendix:algorithm} gives the full training step.

\subsection{Preliminaries: GRPO's Failure Mode}
\label{sec:method:prelim}

Let $x$ denote a question and $y_{\rm S}$ a response sampled from a student policy $\pi_\theta$. For each $x$ we draw a group of $G_{\rm S}$ student rollouts $\{y_{\rm S}^{(g)}\}_{g=1}^{G_{\rm S}} \sim \pi_\theta(\cdot \mid x)$ and assign each an outcome reward $r(x, y_{\rm S}^{(g)}) \in \{0, 1\}$ that signals whether the final answer is correct. Let $\bar{r}_x$ and $\mathrm{std}_x$ denote the within-group mean and standard deviation of $\{r(x, y_{\rm S}^{(g)})\}_{g=1}^{G_{\rm S}}$. The standard group-relative advantage~\citep{shao2024deepseekmath,yu2025dapo} is
\begin{equation}
A^{(g)} \;=\; \frac{r(x, y_{\rm S}^{(g)}) - \bar{r}_x}{\mathrm{std}_x + \epsilon}.
\label{eq:grpo-adv}
\end{equation}
\cref{eq:grpo-adv} is the textbook form ZPPO conceptually builds on; our experiments use the two-step REINFORCE++ variant of \citet{hu2025reinforce++} (Step~1 subtracts the per-group mean; Step~2 normalizes across groups, here the non-trivial groups of one mini-batch), restated in \cref{sec:appendix:algorithm}. The student update applies the PPO surrogate~\citep{schulman2017proximal} on top of $A^{(g)}$. Either form leaves a blind spot for small students. Whenever a rollout group is all-wrong ($\bar{r}_x\!=\!0$) or all-correct ($\bar{r}_x\!=\!1$), every advantage in the group is exactly zero, so the question contributes no gradient signal at all. For a small student, the all-wrong case is exactly where teacher guidance could help, and the wrong approach the group agrees on is not even penalized. ZPPO's goal is to recover a learning signal on those hard questions \emph{without ever placing a teacher response in the student's gradient}. We call $x$ a \emph{hard question} when $\bar{r}_x<\tau$ and use $\tau\!=\!0.5$ throughout, the majority-failure boundary, which admits all-wrong and partially solved groups alike (defined and swept in \cref{sec:appendix:algorithm}).

\subsection{Prompt Reformulation: BCQ and NCQ}
\label{sec:method:bcqncq}

\begin{figure*}[t!]
    \vspace{-3mm}
    \centering
    \includegraphics[width=\linewidth]{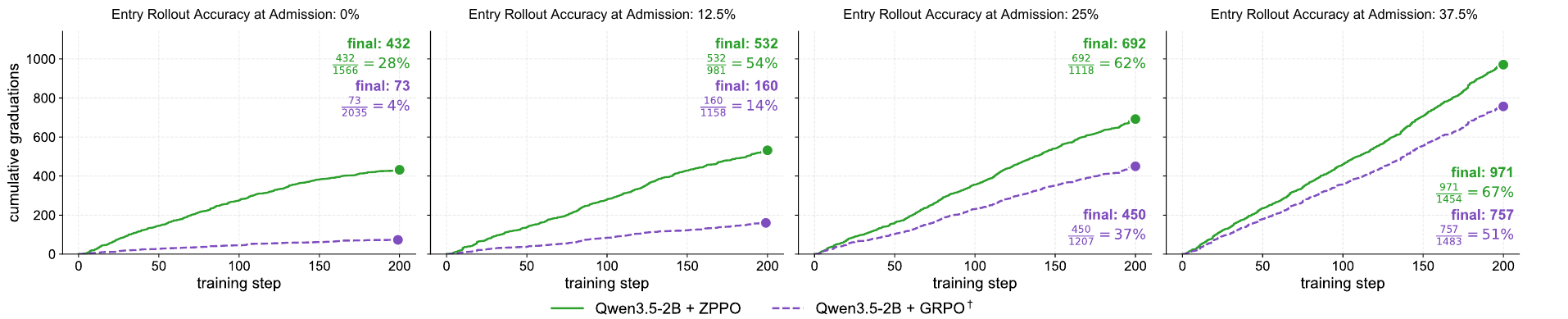}
    \vspace{-5mm}
    \caption{Cumulative graduate counts for ZPPO vs.\ GRPO$^{\dagger}$ at 2B. The printed ratio is graduated over the entries the buffer re-sampled at least once (\cref{sec:appendix:dynamics_2b}).}
    \label{fig:zone}
\end{figure*}

Both BCQ and NCQ start from a hard question $x$ on which we have already drawn $G_{\rm S}$ student rollouts $\{y_{\rm S}^{(g)}\}_{g=1}^{G_{\rm S}}$. In parallel, we sample $G_{\rm T}$ teacher rollouts on $x$ from a frozen teacher policy $\pi_{\rm T}$, score them with the same outcome reward, and keep the correct ones as the pool $\{y_{\rm T}^{(+)}\}$ from which BCQ draws candidates. We use \emph{on-policy} throughout in the \emph{response-level} sense: every gradient-counted token is sampled by the student, never by the teacher. BCQ and NCQ prompts append teacher-compressed text to the original question, but only as input, never as a gradient target. The same checkpoint also judges free-form answers for every method's binary reward (\cref{sec:appendix:hyperparameters:zppo}). Reformulation is also capped: per rollout step, the combined number of BCQ and NCQ instances stays at or below a fraction $\rho_{\rm aug}$ of the new questions (\cref{sec:method:buffer}). On the plain prompt, the original question without candidates, training is therefore ordinary on-policy RL, whether the question is new or replayed. When a question is reformulated, that plain group stays in the same rollout step, and the BCQ or NCQ prompt built from it adds a further group under the same reward and the same parameters. Evaluation always uses the plain prompt (\cref{sec:experiments:discussion}).

\begin{table*}[t!]
    \vspace{-3mm}
    \centering
    \caption{Component ablation of ZPPO over the 16 VLM benchmarks at 0.8B and 2B. Rows add the prompt replay buffer ($^{\dagger}$), BCQ, and NCQ to GRPO, alone or combined. Bold marks the column maximum and is shared by every row that attains it. 4B and 9B: \cref{sec:appendix:main_tables}.}
    \label{tab:ablation}
    \resizebox{\linewidth}{!}{
    \setlength{\tabcolsep}{1mm}
    \renewcommand{\arraystretch}{1.05}
    \begin{tabular}{l*{17}{c}}
    \toprule
    16 VLM Benchmarks
    & AI2D
    & BabyV
    & CharXiv
    & DynaM
    & EmbSp
    & InfoVQA
    & MVerse
    & MVision
    & MVista
    & MMMU$^\text{Pro}$
    & MM-Vet
    & OCR$^\text{EN}$
    & OCR$^\text{ZH}$
    & VisP
    & VBlind
    & WeMath
    & Avg \\
    \midrule
    Qwen3.5-0.8B & 65.6 & 6.7 & 54.3 & 17.8 & 67.9 & 68.6 & 43.5 & 16.4 & 60.7 & 26.8 & 53.2 & 40.0 & 17.0 & 20.5 & 42.8 & 54.4 & \cellcolor{gray!10}41.0 \\
    \cdashline{1-18}\noalign{\vskip 0.5ex}
    $+$ GRPO & 69.5 & 8.6 & 57.0 & 21.3 & 69.2 & 71.0 & 47.7 & 20.1 & 65.2 & 29.9 & 55.6 & 40.9 & 17.0 & 24.2 & 43.2 & 60.6 & \cellcolor{gray!10}43.8 \\
    $+$ GRPO$^{\dagger}$ & 71.2 & 9.8 & 59.9 & 23.6 & 69.4 & 72.4 & 51.1 & 20.9 & 68.3 & 30.5 & 57.5 & 41.3 & 17.5 & 27.8 & 43.6 & 62.5 & \cellcolor{gray!10}45.5 \\
    $+$ GRPO $+$ Both & 71.7 & 9.8 & 58.7 & 23.9 & 69.1 & 72.2 & 50.7 & 21.3 & 67.1 & 31.7 & 56.6 & 40.1 & 17.1 & 26.6 & 43.6 & 63.3 & \cellcolor{gray!10}45.2 \\
    $+$ GRPO$^{\dagger}$ $+$ BCQ & 75.2 & 12.3 & 61.8 & 28.2 & 71.1 & 74.5 & 55.9 & 26.4 & 71.7 & 35.2 & 59.1 & 42.2 & 18.4 & 31.8 & 44.3 & 69.6 & \cellcolor{gray!10}48.6 \\
    $+$ GRPO$^{\dagger}$ $+$ NCQ & 72.0 & 10.2 & 60.3 & 24.3 & 70.0 & 73.0 & 51.5 & 22.7 & 68.7 & 32.1 & 57.9 & 41.6 & 18.1 & 28.2 & 44.0 & 64.6 & \cellcolor{gray!10}46.2 \\
    \rowcolor{colorful}
    $+$ ZPPO & \textbf{76.5} & \textbf{13.9} & \textbf{63.9} & \textbf{31.1} & \textbf{71.5} & \textbf{75.3} & \textbf{59.3} & \textbf{29.2} & \textbf{73.2} & \textbf{37.6} & \textbf{59.9} & \textbf{42.5} & \textbf{18.7} & \textbf{35.0} & \textbf{44.7} & \textbf{71.7} & \cellcolor{gray!10}\textbf{50.3} \\
    \midrule
    Qwen3.5-2B & 81.9 & 11.6 & 71.6 & 41.1 & 78.2 & 81.2 & 69.7 & 38.4 & 78.6 & 46.2 & 69.7 & 44.7 & 24.0 & 38.3 & 55.2 & 77.9 & \cellcolor{gray!10}56.8 \\
    \cdashline{1-18}\noalign{\vskip 0.5ex}
    $+$ GRPO & 83.5 & 13.7 & 72.8 & 45.6 & 78.3 & 82.4 & 72.3 & 43.1 & 79.0 & 49.3 & 72.3 & 46.1 & 24.9 & 39.2 & 56.5 & 80.4 & \cellcolor{gray!10}58.7 \\
    $+$ GRPO$^{\dagger}$ & 83.8 & 14.4 & 73.0 & 45.9 & 78.7 & 83.1 & 72.8 & 43.4 & 79.3 & 49.6 & 74.0 & 46.4 & 25.2 & 40.3 & 57.0 & 80.7 & \cellcolor{gray!10}59.2 \\
    $+$ GRPO $+$ Both & 83.9 & 14.3 & 72.7 & 46.0 & 78.1 & 82.8 & 72.4 & 42.7 & 79.4 & 49.8 & 73.2 & 45.4 & 24.5 & 39.6 & 57.7 & 79.8 & \cellcolor{gray!10}58.9 \\
    $+$ GRPO$^{\dagger}$ $+$ BCQ & 84.7 & 16.7 & 73.7 & 49.6 & 79.2 & 84.0 & 74.3 & 47.3 & 80.1 & 51.4 & 75.8 & 48.1 & 25.6 & 41.1 & 59.1 & 81.8 & \cellcolor{gray!10}60.8 \\
    $+$ GRPO$^{\dagger}$ $+$ NCQ & 84.3 & 15.8 & 73.4 & 48.1 & 79.0 & 83.6 & 73.5 & 45.7 & 79.7 & 50.5 & 74.9 & 47.6 & 25.5 & 40.9 & 58.4 & 81.2 & \cellcolor{gray!10}60.1 \\
    \rowcolor{colorful}
    $+$ ZPPO & \textbf{85.3} & \textbf{18.6} & \textbf{73.9} & \textbf{52.7} & \textbf{79.5} & \textbf{84.6} & \textbf{76.0} & \textbf{50.5} & \textbf{80.5} & \textbf{53.2} & \textbf{77.1} & \textbf{48.8} & \textbf{26.0} & \textbf{42.0} & \textbf{60.5} & \textbf{82.6} & \cellcolor{gray!10}\textbf{62.0} \\
    \bottomrule
    \end{tabular}
    }
\end{table*}

\textbf{(i) Candidate compression:} Before any candidate enters a prompt, the frozen teacher rewrites it under a shared compression prompt and a shared token cap (\cref{sec:appendix:hyperparameters:zppo}). The rewrite keeps only the essential reasoning steps and the final answer, and drops the exploratory text, self-corrections, and retries of the source rollout; it is instructed not to re-derive the problem or add anything the source does not contain. The compressor reads only the source response, never the question or the image.

\textbf{(ii) Binary Candidate-included Question (BCQ):} For each hard question that admits at least one correct teacher response, BCQ uniformly samples one $y_{\rm T}^{(+)}$ and one wrong student rollout $y_{\rm S}^{(-)}$. It teacher-compresses both responses, anonymizes them inside identical \texttt{<candidate>} tags, and randomly shuffles the order. The result is appended to $x$ together with a single instruction (\cref{sec:appendix:hyperparameters:zppo} reproduces the verbatim code-side template, including the per-candidate \texttt{<candidate>} blocks):
\begin{prompttemplate}
\footnotesize
\emph{Here are two candidate responses in \texttt{<candidate>} \texttt{</candidate>} tags to the question above. One is correct and another is wrong. Use these as references to help you solve the problem.}
\end{prompttemplate}
The student then samples a new rollout group $\{y_{\rm BCQ}^{(g)}\}_{g=1}^{G_{\rm S}} \sim \pi_\theta(\cdot \mid x_{\rm BCQ})$ from the reformulated prompt. The intended effect comes from placing a correct teacher response and a wrong student response \emph{side by side}, anonymized, shuffled, and without saying which is correct. The student therefore solves the problem while weighing the two candidates.
\begin{figure*}[t!]
    \vspace{-3mm}
    \centering
    \includegraphics[width=\linewidth]{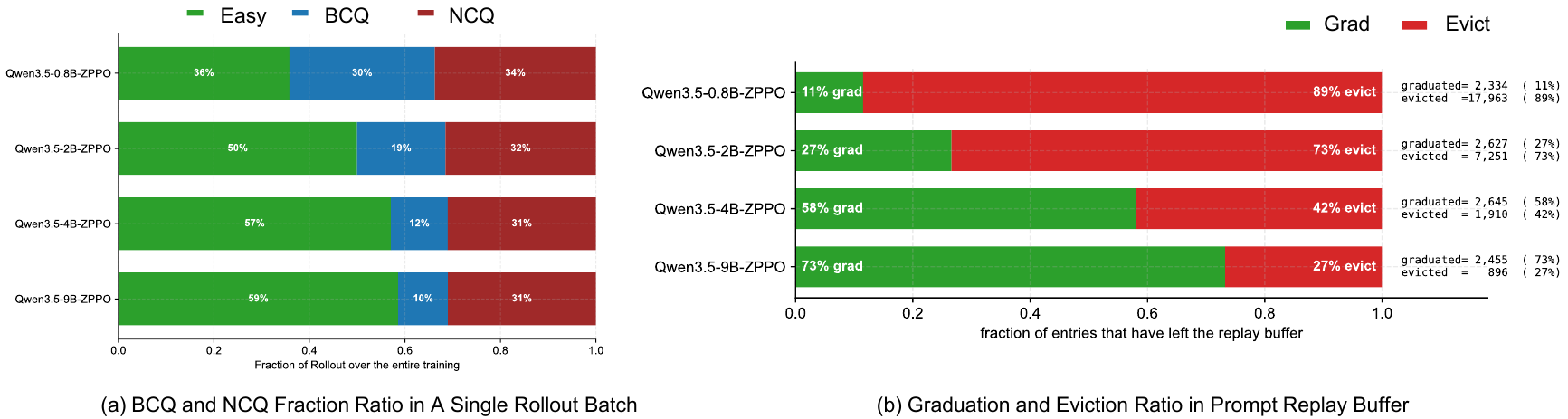}
    \vspace{-5mm}
    \caption{(a)~Easy questions ($\bar{r}_x\!\ge\!0.5$) and the BCQ and NCQ reformulations available on hard ones over training, before the caps, per scale, each as a share of their summed count; a hard question can yield both reformulations (\cref{sec:appendix:algorithm}). (b)~Replay-buffer entries that have left the buffer: graduated vs.\ evicted (\cref{sec:appendix:zone}).}
    \label{fig:system}
\end{figure*}

\textbf{(iii) Negative Candidate-included Question (NCQ):} For each hard question, NCQ collects every wrong student rollout $y_{\rm S}^{(-)}$ in the current group whose final answer parses. It lists the \emph{distinct} parsed answers explicitly inside the prompt, so the answer list is generally shorter than the rollout list. Each teacher-compressed reasoning trace is then appended as a \texttt{<candidate>} block, with the instruction (full block in \cref{sec:appendix:hyperparameters:zppo}):
\begin{prompttemplate}
\footnotesize
\emph{The following answers are all WRONG: $\langle$\,parsed answer\,$\rangle$. Below are the incorrect reasoning processes in \texttt{<candidate>} \texttt{</candidate>} tags.}
\end{prompttemplate}
The student then samples a new rollout group $\{y_{\rm NCQ}^{(g)}\}_{g=1}^{G_{\rm S}} \sim \pi_\theta(\cdot \mid x_{\rm NCQ})$, again entirely student-generated. The role of NCQ, however, is structurally different. In a standard rollout group each wrong rollout contributes its advantage independently, so NCQ is the first place in our training loop where a question's failed attempts become collectively visible. Confronted with them, the student is cued to recognize the shared error patterns and avoid them.

\subsection{Integration with Prompt Replay Buffer}
\label{sec:method:buffer}

In ZPPO, the prompt replay buffer $\mathcal{B}$ serves \emph{to amplify BCQ and NCQ} on questions the student has not yet mastered. It stores only the question $x$ (image and text), never a response.

\textbf{(i) Admission and graduation:} A question $x$ is \emph{admitted} if its $\bar{r}_x\!<\!\tau$. An admitted $x$ leaves $\mathcal{B}$ in one of two ways: it is \emph{graduated} on any later step where $\bar{r}_x\!\geq\!\tau$, or it is \emph{FIFO-evicted} once $|\mathcal{B}|$ exceeds $|\mathcal{B}|_{\max}$. We update $\mathcal{B}$ after each training step, and only the original question is stored, never its BCQ or NCQ reformulation. A replayed question that is still hard on its next visit is again eligible for one or both reformulations. The buffer therefore tracks the questions the student cannot yet solve alone, and graduation marks the ones guidance reaches: the student's \emph{current} zone of proximal development.

\textbf{(ii) Sampling and capacity:}
Each rollout batch combines new questions from the data loader with replay samples drawn uniformly from $\mathcal{B}$, where the replay count is a fixed fraction $\rho_{\rm replay}$ of the new-question count. From the union of new and replayed questions, BCQ and NCQ are constructed on the hardest ones first, ranked by ascending $\bar{r}_x$. The \emph{combined} BCQ+NCQ count per rollout step is then capped at a fraction $\rho_{\rm aug}$ of the new-question count (\cref{sec:appendix:algorithm}). The buffer therefore re-exposes hard questions, each eligible for freshly sampled BCQ/NCQ candidates on every visit.

\section{Experiments}
\label{sec:experiments}

\paragraph{Students and teacher.}
We instantiate ZPPO on Qwen3.5~\citep{qwen3.5} at four student scales ($0.8$B, $2$B, $4$B, $9$B) with a $27$B teacher (\cref{sec:appendix:teacher_capability}), post-training each student as a VLM on the ZPPO-$77$K multimodal RL dataset (\cref{sec:appendix:dataset:corpus}).

\paragraph{Training pipeline.}
ZPPO follows GRPO~\citep{shao2024deepseekmath} with three DAPO ingredients~\citep{yu2025dapo}: clip-higher, a token-level policy gradient loss, and no KL penalty. We use $I\!=\!4$ iterations per step (vs.\ DAPO's $I\!=\!16$) and batch-level advantage normalization~\citep{hu2025reinforce++} with zero-advantage groups excluded from the batch statistics. Every run disables DAPO's dynamic sampling, which drops all-correct/wrong groups and resamples to refill the batch, since the all-wrong groups are exactly what BCQ and NCQ reformulate; \cref{tab:compute_matched} adds it to GRPO$^{\dagger}$ as DAPO$^{\dagger}$. The full configuration is in \cref{sec:appendix:hyperparameters:zppo}, the definition of $I$ and the corresponding budgets in \cref{sec:appendix:baselines}, and the per-run wall-clock in \cref{sec:appendix:hyperparameters:compute}.

\paragraph{Baselines.}
We compare ZPPO against four baselines (\cref{sec:appendix:baselines}) from the first two lines of work in \cref{sec:related}, distillation and RL post-training; \cref{sec:experiments:discussion} covers the third, prompt-side guidance (Hint and Prefix) and curricula, and the fourth, learning from mistakes, along with further RL and imitation methods.
\textbf{(i) \emph{Off-policy distillation}}: imitate fresh teacher rollouts drawn online and filtered by correctness on every step.
\textbf{(ii) \emph{On-policy distillation}}: the student samples its own response, the teacher re-scores it for target logits, and the student imitates them.
\textbf{(iii) \emph{GRPO}}: ZPPO's pipeline without BCQ, NCQ, and the replay buffer.
\textbf{(iv) \emph{GRPO$^{\dagger}$}}: ZPPO's pipeline without BCQ and NCQ. All methods share the decoding and scoring of \cref{sec:appendix:benchmarks}, and seven other judges keep ZPPO above GRPO$^{\dagger}$ and On-Distill$^{\dagger}$ (\cref{sec:appendix:judges}).

\subsection{Main results}
\label{sec:experiments:main}

\Cref{tab:main} ($16$ VLM benchmarks) and \cref{tab:generalization} ($10$ LLM and $5$ Video benchmarks) report the main comparison; the $4$B/$9$B blocks are in \cref{sec:appendix:main_tables}. \Cref{fig:teaser} plots the buffer-augmented rows of both tables as gains over the base.

\textbf{(i)~Distillation \emph{hurts generalization beyond the training corpus}.} The best distillation variant stays close to the original on VLM, while Off- and On-Policy distillation \emph{lose} accuracy on Video and \emph{degrade} the LLM--Video average at every scale (\cref{tab:generalization}; \cref{sec:appendix:baselines,sec:appendix:main_tables}), sharpest at $0.8$B.

\textbf{(ii)~Replay alone is insufficient.} Adding the buffer to either distillation or GRPO yields modest Avg gains in \cref{tab:main}, and re-exposure alone graduates few of the hardest questions (\cref{fig:zone}).

\textbf{(iii)~ZPPO leads at every scale and benchmark family.} The Avg $\Delta$ row is positive across all blocks of \cref{tab:main} and \cref{tab:generalization}, with the largest gains where the small-student gap is widest. ZPPO and GRPO$^{\dagger}$ both use the replay buffer, so their gap comes from BCQ and NCQ, and its bootstrap interval over benchmarks excludes zero at every scale and family (\cref{sec:appendix:stat_reliability}). \cref{sec:appendix:backbone_families} extends the comparison to Ovis2.5-2B~\citep{lu2025ovis2} and InternVL3.5-2B~\citep{wang2025internvl3}.

\begin{figure*}[t!]
    \vspace{-3mm}
    \centering
    \includegraphics[width=\linewidth]{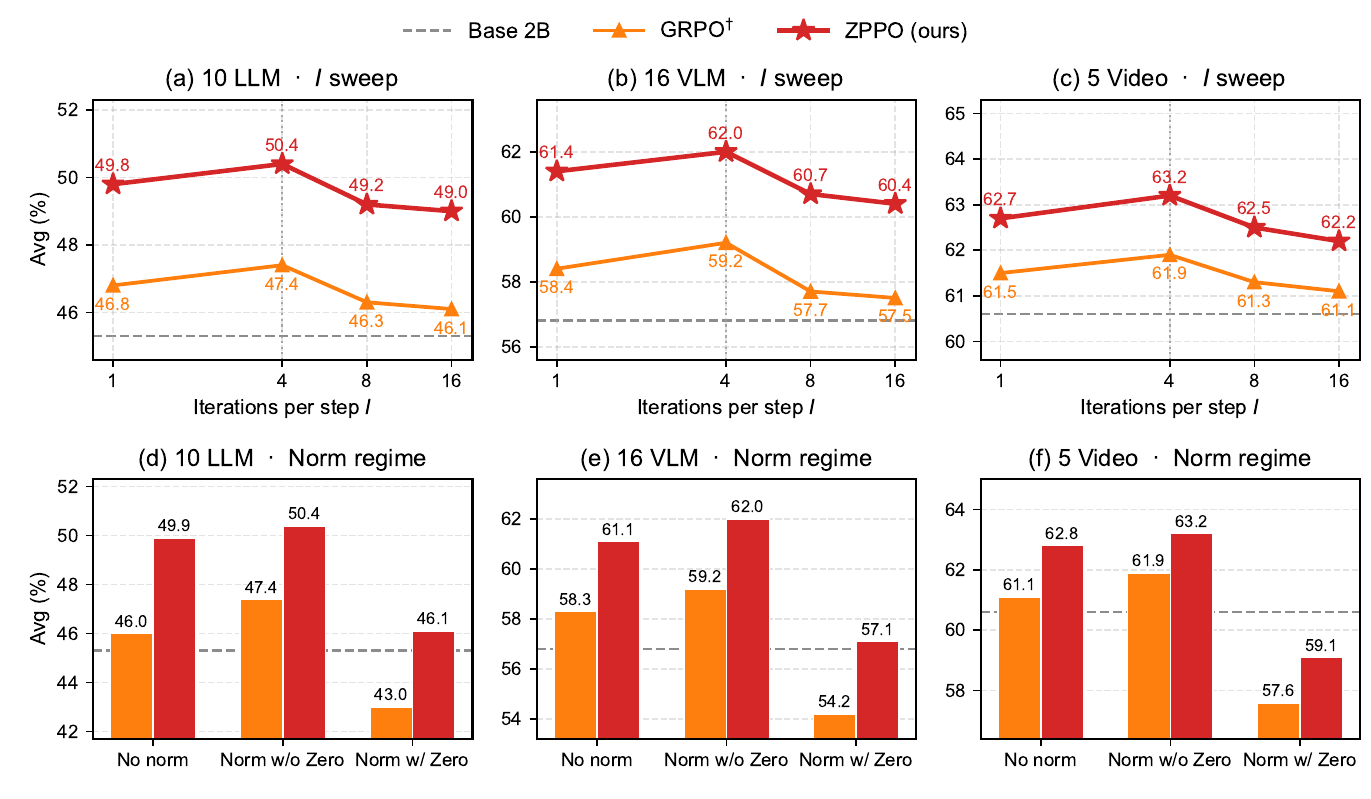}
    \vspace{-5mm}
    \caption{RL recipe ablation at 2B on the 10 LLM (a, d), 16 VLM (b, e), and 5 Video (c, f) benchmarks. \textbf{(Top, a--c)}~Iterations per step $I\!\in\!\{1,4,8,16\}$. \textbf{(Bottom, d--f)}~Batch-level advantage normalization: \emph{No norm}, \emph{Norm w/o Zero} (ZPPO default), and \emph{Norm w/ Zero}. Dashed gray line marks original 2B.}
    \label{fig:recipe}
\end{figure*}

\subsection{Inside the Zone: Admission to Graduation}
\label{sec:experiments:zone}

\textbf{(i)~BCQ and NCQ lift plain-rollout accuracy on the hardest training questions.} \Cref{fig:zone} splits the $2$B buffer entries by plain-rollout accuracy at admission. At $0\%$, where ordinary RL has zero gradient signal, ZPPO graduates $28\%$ of its re-sampled entries ($432$/$1566$) vs.\ GRPO$^{\dagger}$'s $4\%$ ($73$/$2035$); at the next-hardest, $54\%$ vs.\ $14\%$. ZPPO also graduates more entries at the other two admission accuracies. Graduation is scored on plain rollouts of the same training questions.

\textbf{(ii)~Batch and graduation dynamics with scale.} \Cref{fig:system}a gives the share of Easy questions and the share of the BCQ and NCQ reformulations on hard questions. BCQ requires the teacher to succeed where the student fails; as students scale up, the remaining hard questions are increasingly ones the teacher also fails on, so BCQ candidates become scarce (\cref{fig:system}a). In the buffer, most entries are evicted before they graduate at the small scales, and most graduate before they are evicted at the large ones in \cref{fig:system}b. As students grow, ZPPO's guidance thus shifts from BCQ to NCQ (\cref{sec:appendix:algorithm}), and the buffer shifts from eviction to graduation (\cref{sec:appendix:zone}).

\subsection{Component ablation}
\label{sec:experiments:ablation}

\Cref{tab:ablation} measures the ZPPO components on the $16$ VLM benchmarks, adding them to GRPO alone or combined (\emph{Both}\,=\,BCQ\,$+$\,NCQ).

\begin{figure*}[t!]
    \vspace{-3mm}
    \centering
    \includegraphics[width=0.4\linewidth]{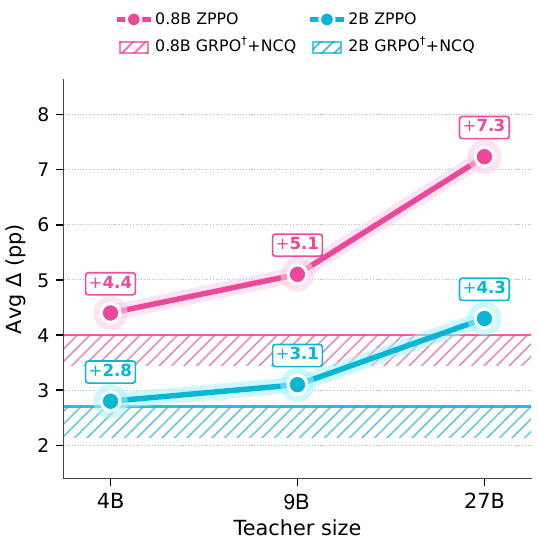}
    \vspace{-3mm}
    \caption{Mean family gain $\Delta$ (pp) across teacher size. The 0.8B and 2B students are trained with 4B, 9B, and 27B teachers (\cref{sec:appendix:teacher_capability}), and each point averages the LLM, VLM, and Video gains over the untrained base. Hatched bands mark GRPO$^{\dagger}\!+\!$NCQ with the 27B teacher (value at the upper edge). A larger teacher widens ZPPO's gain; a 4B teacher brings it close to GRPO$^{\dagger}\!+\!$NCQ.}
    \label{fig:teacher_scale}
\end{figure*}

\textbf{(i)~Neither component suffices alone.} The buffer on its own (GRPO$^{\dagger}$) gains a little over GRPO, and BCQ$+$NCQ without the buffer (GRPO\,$+$\,\emph{Both}) does no better than the buffer alone at any of the four scales (\cref{tab:ablation}, \cref{sec:appendix:main_tables}).

\textbf{(ii)~Replay and reformulation are complementary.} Pairing BCQ \emph{or} NCQ with the buffer beats the buffer alone (GRPO$^{\dagger}$) and the reformulations alone (GRPO\,$+$\,\emph{Both}), since a replayed question can be reformulated again with a freshly sampled candidate set; \cref{sec:appendix:ablation_gen} reports the same ablation on the $10$ LLM and $5$ Video benchmarks.

\textbf{(iii)~The full recipe is strongest.} ZPPO attains the column maximum on every VLM benchmark at all four scales (\cref{tab:ablation}, \cref{sec:appendix:main_tables}), and its two reformulations carry the gain at different scales. BCQ's contribution on VLM shrinks steadily from $0.8$B to $9$B as its candidates thin out (\cref{sec:experiments:zone}), while NCQ's varies much less. With the buffer, BCQ leads NCQ at $0.8$B and $2$B, where the teacher still solves many of the student's hard questions, and NCQ leads at $4$B and $9$B, where it solves fewer. Neither suffices at every scale, so ZPPO uses both. It also relies on the buffer, as (i) and (ii) show. Every component is therefore needed for the best result at every scale. \cref{sec:appendix:stat_reliability} bootstraps both marginals.

\subsection{Additional RL recipe choices}
\label{sec:experiments:recipe}

\begin{table*}[t!]
    \vspace{-3mm}
    \def\tabscale{0.75}
    \begin{minipage}[t]{0.49\linewidth}
        \centering
        \caption{Hint-based guidance methods}
        \label{tab:hint_methods}
        \begin{adjustbox}{scale=\tabscale, max width=\linewidth}
        \setlength{\tabcolsep}{1.2mm}
        \renewcommand{\arraystretch}{0.95}
        \begin{tabular}{lcccccc}
            \toprule
                                          & \multicolumn{3}{c}{Qwen3.5-0.8B}                    & \multicolumn{3}{c}{Qwen3.5-2B}                      \\
            \cmidrule(lr){2-4}\cmidrule(lr){5-7}
            Method                        & LLM             & VLM             & Video           & LLM             & VLM             & Video           \\
            \midrule
            Base                          & $25.2$          & $41.0$          & $48.3$          & $45.3$          & $56.8$          & $60.6$          \\
            \cdashline{1-7}\noalign{\vskip 0.3ex}
            $+$GRPO$^{\dagger}$           & $28.7$          & $45.5$          & $50.5$          & $47.4$          & $59.2$          & $61.9$          \\
            $+$NuRL$^{\dagger}$~\citep{chen2026nudging}& $28.6$          & $46.8$          & $49.2$          & $47.1$          & $59.6$          & $61.4$          \\
            $+$SAGE$^{\dagger}$~\citep{liao2026selfhinting} & $28.3$     & $46.1$          & $49.8$          & $47.4$          & $60.1$          & $61.5$          \\
            $+$HiLL$^{\dagger}$~\citep{xia2026learning}& $28.7$          & $45.9$          & $49.5$          & $46.9$          & $59.8$          & $61.2$          \\
            $+$Guide-GRPO$^{\dagger}$~\citep{nath2025adaptive}& $28.5$          & $46.6$          & $49.9$          & $47.3$          & $59.4$          & $61.3$          \\
            $+$HINT$^{\dagger}$~\citep{wang2026don}    & $28.6$          & $46.4$          & $49.7$          & $47.2$          & $60.2$          & $61.5$          \\
            \cdashline{1-7}\noalign{\vskip 0.3ex}
            $+$GRPO$^{\dagger}$$+$Hint (control)& $28.8$          & $47.2$          & $50.0$          & $47.6$          & $60.4$          & $61.6$          \\
            $+$GRPO$^{\dagger}$$+$BCQ     & $31.5$          & $48.6$          & $52.0$          & $49.1$          & $60.8$          & $62.6$          \\
            \rowcolor{colorful}
            $+$ZPPO                       & $\mathbf{33.1}$ & $\mathbf{50.3}$ & $\mathbf{52.9}$ & $\mathbf{50.4}$ & $\mathbf{62.0}$ & $\mathbf{63.2}$ \\
            \bottomrule
        \end{tabular}
        \end{adjustbox}
    \end{minipage}\hfill
    \begin{minipage}[t]{0.49\linewidth}
        \centering
        \caption{Prefix-based guidance methods}
        \label{tab:prefix_methods}
        \begin{adjustbox}{scale=\tabscale, max width=\linewidth}
        \setlength{\tabcolsep}{1.2mm}
        \renewcommand{\arraystretch}{0.95}
        \begin{tabular}{lcccccc}
            \toprule
                                          & \multicolumn{3}{c}{Qwen3.5-0.8B}                    & \multicolumn{3}{c}{Qwen3.5-2B}                      \\
            \cmidrule(lr){2-4}\cmidrule(lr){5-7}
            Method                        & LLM             & VLM             & Video           & LLM             & VLM             & Video           \\
            \midrule
            Base                          & $25.2$          & $41.0$          & $48.3$          & $45.3$          & $56.8$          & $60.6$          \\
            \cdashline{1-7}\noalign{\vskip 0.3ex}
            $+$GRPO$^{\dagger}$           & $28.7$          & $45.5$          & $50.5$          & $47.4$          & $59.2$          & $61.9$          \\
            $+$LUFFY$^{\dagger}$~\citep{NEURIPS2025_a9d5c33e}& $26.7$          & $45.0$          & $48.9$          & $46.1$          & $59.4$          & $60.8$          \\
            $+$Prefix-RFT$^{\dagger}$~\citep{huang2025blending}& $27.0$          & $45.4$          & $48.5$          & $45.8$          & $59.2$          & $60.9$          \\
            $+$BREAD$^{\dagger}$~\citep{zhang2025bread}& $26.5$          & $44.8$          & $49.0$          & $46.2$          & $59.3$          & $60.7$          \\
            $+$StepHint$^{\dagger}$~\citep{zhang2025stephint}& $26.9$          & $45.3$          & $48.8$          & $46.0$          & $59.5$          & $60.9$          \\
            $+$Hint-GRPO$^{\dagger}$~\citep{huang2025boosting}& $26.6$          & $45.1$          & $48.7$          & $46.1$          & $59.3$          & $60.8$          \\
            \cdashline{1-7}\noalign{\vskip 0.3ex}
            $+$GRPO$^{\dagger}$$+$Prefix (control)& $27.1$          & $45.5$          & $49.1$          & $46.3$          & $59.5$          & $61.0$          \\
            $+$GRPO$^{\dagger}$$+$BCQ     & $31.5$          & $48.6$          & $52.0$          & $49.1$          & $60.8$          & $62.6$          \\
            \rowcolor{colorful}
            $+$ZPPO                       & $\mathbf{33.1}$ & $\mathbf{50.3}$ & $\mathbf{52.9}$ & $\mathbf{50.4}$ & $\mathbf{62.0}$ & $\mathbf{63.2}$ \\
            \bottomrule
        \end{tabular}
        \end{adjustbox}
    \end{minipage}
    \par\medskip
    \begin{minipage}[t]{0.49\linewidth}
        \centering
        \caption{Curriculum or imitation RL}
        \label{tab:curriculum}
        \begin{adjustbox}{scale=\tabscale, max width=\linewidth}
        \setlength{\tabcolsep}{1.2mm}
        \renewcommand{\arraystretch}{0.95}
        \begin{tabular}{lcccccc}
            \toprule
                                          & \multicolumn{3}{c}{Qwen3.5-0.8B}                    & \multicolumn{3}{c}{Qwen3.5-2B}                      \\
            \cmidrule(lr){2-4}\cmidrule(lr){5-7}
            Method                        & LLM             & VLM             & Video           & LLM             & VLM             & Video           \\
            \midrule
            Base                          & $25.2$          & $41.0$          & $48.3$          & $45.3$          & $56.8$          & $60.6$          \\
            \cdashline{1-7}\noalign{\vskip 0.3ex}
            $+$GRPO$^{\dagger}$           & $28.7$          & $45.5$          & $50.5$          & $47.4$          & $59.2$          & $61.9$          \\
            $+$CLPO$^{\dagger}$~\citep{zhang2025clpo}  & $28.6$          & $45.7$          & $50.8$          & $47.8$          & $59.9$          & $61.9$          \\
            $+$Goldilocks RL$^{\dagger}$~\citep{mahrooghi2026goldilocks}& $29.0$          & $45.9$          & $51.0$          & $47.8$          & $59.3$          & $62.0$          \\
            $+$ADCL/EGSR$^{\dagger}$~\citep{zhang2025learninglike} & $28.9$ & $46.3$          & $50.7$          & $47.7$          & $59.4$          & $62.1$          \\
            $+$AdaCuRL$^{\dagger}$~\citep{li2026adacurl} & $29.4$      & $46.2$          & $50.6$          & $47.5$          & $59.3$          & $61.7$          \\
            $+$RIL$^{\dagger}$~\citep{NEURIPS2025_e5849736}& $28.6$          & $46.0$          & $50.7$          & $48.0$          & $59.8$          & $61.8$          \\
            $+$Masters$^{\dagger}$~\citep{Lee_2026_CVPR_Masters}& $28.7$          & $45.7$          & $50.7$          & $47.7$          & $59.3$          & $61.9$          \\
            $+$PRISM$^{\dagger}$~\citep{wang2026prism}& $29.1$          & $46.3$          & $50.8$          & $47.5$          & $59.5$          & $61.9$          \\
            \rowcolor{colorful}
            $+$ZPPO                       & $\mathbf{33.1}$ & $\mathbf{50.3}$ & $\mathbf{52.9}$ & $\mathbf{50.4}$ & $\mathbf{62.0}$ & $\mathbf{63.2}$ \\
            \bottomrule
        \end{tabular}
        \end{adjustbox}
    \end{minipage}\hfill
    \begin{minipage}[t]{0.49\linewidth}
        \centering
        \caption{Learning from mistakes: SFT \& RL}
        \label{tab:mistakes}
        \begin{adjustbox}{scale=\tabscale, max width=\linewidth}
        \setlength{\tabcolsep}{1.2mm}
        \renewcommand{\arraystretch}{0.95}
        \begin{tabular}{lcccccc}
            \toprule
                                          & \multicolumn{3}{c}{Qwen3.5-0.8B}                    & \multicolumn{3}{c}{Qwen3.5-2B}                      \\
            \cmidrule(lr){2-4}\cmidrule(lr){5-7}
            Method                        & LLM             & VLM             & Video           & LLM             & VLM             & Video           \\
            \midrule
            Base                          & $25.2$          & $41.0$          & $48.3$          & $45.3$          & $56.8$          & $60.6$          \\
            \cdashline{1-7}\noalign{\vskip 0.3ex}
            $+$Correction-SFT$^{\dagger}$       & $23.8$    & $41.9$          & $46.7$          & $44.1$          & $57.6$          & $59.3$          \\
            $+$Error-reflection-SFT$^{\dagger}$ & $24.2$    & $42.2$          & $47.0$          & $44.6$          & $57.9$          & $59.6$          \\
            $+$ReLIFT$^{\dagger}$~\citep{ma2026learning}& $28.1$          & $46.9$          & $49.6$          & $47.1$          & $60.4$          & $61.5$          \\
            \cdashline{1-7}\noalign{\vskip 0.3ex}
            $+$GRPO$^{\dagger}$           & $28.7$          & $45.5$          & $50.5$          & $47.4$          & $59.2$          & $61.9$          \\
            $+$RL-ZVP$^{\dagger}$~\citep{le2026no}& $29.0$          & $45.8$          & $50.7$          & $47.5$          & $59.5$          & $62.0$          \\
            $+$LENS$^{\dagger}$~\citep{feng2025don}& $29.3$          & $46.1$          & $50.9$          & $47.7$          & $59.8$          & $62.2$          \\
            $+$GRPO$^{\dagger}$$+$NCQ     & $29.4$          & $46.2$          & $51.0$          & $48.5$          & $60.1$          & $62.3$          \\
            \rowcolor{colorful}
            $+$ZPPO                       & $\mathbf{33.1}$ & $\mathbf{50.3}$ & $\mathbf{52.9}$ & $\mathbf{50.4}$ & $\mathbf{62.0}$ & $\mathbf{63.2}$ \\
            \bottomrule
        \end{tabular}
        \end{adjustbox}
    \end{minipage}
    \par\medskip
    \begin{minipage}[t]{0.49\linewidth}
        \centering
        \caption{Candidates and transfer at $2$B}
        \label{tab:transfer}
        \begin{adjustbox}{scale=\tabscale, max width=\linewidth}
        \setlength{\tabcolsep}{1.2mm}
        \renewcommand{\arraystretch}{0.97}
        \begin{tabular}{lccccc}
            \toprule
                                          &                              &               & \multicolumn{3}{c}{Qwen3.5-2B}                      \\
            \cmidrule(lr){4-6}
            Method                        & Prompt acc                   & Copy          & LLM             & VLM             & Video           \\
            \midrule
            Base                          & --                           & --            & $45.3$          & $56.8$          & $60.6$          \\
            \cdashline{1-6}\noalign{\vskip 0.3ex}
            $+$GRPO$^{\dagger}$           & --                           & --            & $47.4$          & $59.2$          & $61.9$          \\
            \quad $+$BCQ (default)        & $0.23\!\rightarrow\!0.56$    & $0.14$        & $49.1$          & $60.8$          & $62.6$          \\
            \quad $+$correct candidate only& $0.88\!\rightarrow\!0.95$   & $0.87$        & $46.3$          & $57.6$          & $60.2$          \\
            \quad $+$answer-removed       & $0.22\!\rightarrow\!0.55$    & $0.16$        & $48.5$          & $60.0$          & $62.0$          \\
            \quad $+$final answer only    & $0.45\!\rightarrow\!0.48$    & $0.00$        & $47.1$          & $58.9$          & $62.0$          \\
            \quad $+$non-anonymized       & $0.67\!\rightarrow\!0.89$    & $0.79$        & $46.5$          & $57.9$          & $60.3$          \\
            \quad $+$unrelated candidates & $0.03\!\rightarrow\!0.17$    & $0.03$        & $46.9$          & $58.4$          & $61.4$          \\
            \rowcolor{colorful}
            $+$ZPPO                       & $0.24\!\rightarrow\!0.53$    & $0.11$        & $\mathbf{50.4}$ & $\mathbf{62.0}$ & $\mathbf{63.2}$ \\
            \bottomrule
        \end{tabular}
        \end{adjustbox}
    \end{minipage}\hfill
    \begin{minipage}[t]{0.49\linewidth}
        \centering
        \caption{Extra compute over GRPO$^{\dagger}$}
        \label{tab:compute_matched}
        \begin{adjustbox}{scale=\tabscale, max width=\linewidth}
        \setlength{\tabcolsep}{1.2mm}
        \renewcommand{\arraystretch}{0.95}
        \begin{tabular}{lcccccc}
            \toprule
                                          & \multicolumn{3}{c}{Qwen3.5-0.8B}                    & \multicolumn{3}{c}{Qwen3.5-2B}                      \\
            \cmidrule(lr){2-4}\cmidrule(lr){5-7}
            Method                        & LLM             & VLM             & Video           & LLM             & VLM             & Video           \\
            \midrule
            Base                          & $25.2$          & $41.0$          & $48.3$          & $45.3$          & $56.8$          & $60.6$          \\
            \cdashline{1-7}\noalign{\vskip 0.3ex}
            $+$GRPO$^{\dagger}$           & $28.7$          & $45.5$          & $50.5$          & $47.4$          & $59.2$          & $61.9$          \\
            \quad $\times$2 rollouts      & $29.0$          & $46.0$          & $50.6$          & $47.6$          & $59.7$          & $62.0$          \\
            \quad $\times$2 epochs        & $28.6$          & $45.6$          & $50.3$          & $47.2$          & $59.4$          & $61.7$          \\
            \quad $\times$2 rollouts \& 2 epochs& $28.9$          & $45.7$          & $50.7$          & $47.4$          & $59.6$          & $61.9$          \\
            \quad $\times$2 replay buffer batch& $29.1$          & $45.7$          & $50.4$          & $47.6$          & $59.9$          & $62.1$          \\
            $+$On-Distill$^{\dagger}$ $\times$2 epochs& $22.4$          & $42.2$          & $45.1$          & $43.0$          & $57.9$          & $58.6$          \\
            \cdashline{1-7}\noalign{\vskip 0.3ex}
            $+$DAPO$^{\dagger}$ $\times$2 epochs& $28.5$          & $45.5$          & $50.1$          & $47.2$          & $59.3$          & $61.4$          \\
            \rowcolor{colorful}
            $+$ZPPO                       & $\mathbf{33.1}$ & $\mathbf{50.3}$ & $\mathbf{52.9}$ & $\mathbf{50.4}$ & $\mathbf{62.0}$ & $\mathbf{63.2}$ \\
            \bottomrule
        \end{tabular}
        \end{adjustbox}
    \end{minipage}
\end{table*}

We isolate two further recipe choices that materially shape student training at $2$B (\cref{fig:recipe}).

\textbf{(i)~Iterations per step balance off-policy drift against rollout cost; $I\!=\!4$ is the sweet spot.} Larger $I$ increases in-step drift: each mini-batch moves the policy further from the rollout policy. DAPO's default $I\!=\!16$~\citep{yu2025dapo} runs $4\times$ as many updates as $I\!=\!4$ (\cref{fig:recipe}a--c); $I\!=\!1$ removes drift but undertrains at matched step count, and $I\!=\!4$ peaks on all three families. At matched prompt draws and update count, GRPO$^{\dagger}$ at $I\!=\!4$ is still $0.9$/$1.7$/$0.4$~pp above $I\!=\!8$ (\cref{sec:appendix:baselines}).

\textbf{(ii)~Excluding zero-advantage groups from batch normalization (Norm w/o Zero) is critical.} \emph{Norm w/o Zero} (ZPPO default) consistently beats \emph{No norm}, while \emph{Norm w/ Zero} clearly degrades below both (\cref{fig:recipe}d--f). Zero-advantage groups carry no signal, yet replaying hard questions adds many of them to each mini-batch. When their zeros enter the batch statistics, the standard deviation shrinks and every nonzero advantage is inflated. The inflation is largest in the mini-batches with the fewest nonzero groups, whose gradients come from the fewest rollouts, thereby amplifying noise in training (\cref{sec:appendix:algorithm}).

\subsection{Discussion and Analysis}
\label{sec:experiments:discussion}

\textbf{(i) Teacher size matters:} A larger teacher widens ZPPO's gain; a $4$B teacher keeps it at or above GRPO$^{\dagger}\!+\!$NCQ, and the drop is steepest at $0.8$B, where BCQ matters most (\cref{fig:teacher_scale}). This is consistent with BCQ candidates thinning as the teacher's lead shrinks (\cref{fig:system}a).

\textbf{(ii) Hint, Prefix, curricula, and imitation:} \Cref{tab:hint_methods,tab:prefix_methods} compare BCQ with five prior methods each of two guidance families, Hint and Prefix. To isolate the form of guidance, we also build a Hint and a Prefix control from their elements, each as on-policy as its family allows: no teacher token enters the gradient, so no extra off-policy correction is needed (\cref{sec:appendix:baselines}). Each control matches or beats every prior method of its family in every column, LLM and Video included, yet BCQ beats both controls in every column, and ZPPO is higher still. Curricula and imitation RL (\cref{tab:curriculum}) stay close to GRPO$^{\dagger}$, whose buffer is already a difficulty-based curriculum, and all fall below BCQ. BCQ's gain therefore comes from letting the student compare a correct with a wrong candidate, which no cue, opening, curriculum, or imitation objective provides (\cref{sec:appendix:bcq_controls}).

\textbf{(iii) Learning from mistakes (SFT \& RL):} \Cref{tab:mistakes} compares NCQ with methods that learn from the student's errors. Correction-SFT combines the teacher-corrected solutions of LeMa~\citep{an2023learning} with the correct/incorrect input tags of NAT~\citep{wang2024learning}, and Error-reflection-SFT follows ORION~\citep{wu2026enhancing} in refining a teacher solution with the student's errors (\cref{sec:appendix:baselines}). Both lift VLM over the base but lower LLM and Video, and ReLIFT~\citep{ma2026learning}, which fine-tunes on teacher solutions to the questions RL keeps failing, shows the same split against GRPO$^{\dagger}$; RL-ZVP~\citep{le2026no} and LENS~\citep{feng2025don}, which reshape the advantage of wrong rollouts, stay close to GRPO$^{\dagger}$ (\cref{sec:appendix:zerovariance}). NCQ, which shows no correct answer and lists all wrong answers, leads every baseline in \cref{tab:mistakes} on LLM and Video.

\textbf{(iv) Why ZPPO works at inference without BCQ/NCQ:} ZPPO always trains on the plain prompt, the one used at inference. BCQ and NCQ only add rollout groups on top of it as an auxiliary signal, in the same loss (\cref{sec:appendix:algorithm}). As in scaffolding, a question receives them only while its plain-rollout accuracy stays below $\tau$, and it graduates once that accuracy reaches $\tau$. The signal carries over: ZPPO's plain-rollout accuracy rises faster than GRPO$^{\dagger}$'s (\cref{sec:appendix:dynamics_2b}), and ZPPO graduates more hard questions on plain rollouts than GRPO$^{\dagger}$ (\cref{sec:experiments:zone}). Solving BCQ and NCQ prompts helps on the plain prompt when the student derives the answer from the candidates' reasoning: it then learns to solve the question on its own rather than to copy. With \emph{correct candidate only} or \emph{non-anonymized} (\cref{sec:appendix:bcq_controls}), accuracy on these prompts in \cref{tab:transfer} is high from the start, but GPT-6 Astra judges most rollouts as copied from a candidate (Copy $0.87$, $0.79$), and the student falls below GRPO$^{\dagger}$. With BCQ and \emph{answer-removed}, it copies far less (Copy $0.14$, $0.16$), and only these two beat GRPO$^{\dagger}$ on every family. ZPPO, which adds NCQ with no correct answer to copy, copies even less on its BCQ prompts (Copy $0.11$).

\textbf{(v) Extra compute over GRPO$^{\dagger}$:} More rollouts, more epochs, a larger replay batch, or the dynamic sampling of DAPO$^{\dagger}$ leaves every baseline in \cref{tab:compute_matched} below ZPPO (\cref{sec:appendix:budget}). The rows that double rollouts or epochs take longer than ZPPO in wall-clock time (\cref{sec:appendix:hyperparameters:compute}), yet none catches up with ZPPO. Beyond compute, ZPPO also stays above self-distillation and hybrid RL$+$distillation (\cref{sec:appendix:teacher_alternatives}) and dense and multimodal rewards (\cref{sec:appendix:mmrl}).
\section{Conclusion}
\label{sec:conclusion}

ZPPO keeps the teacher inside the prompt, where BCQ and NCQ turn all-wrong questions into a learning signal and the replay buffer revisits them. The flip side is a structural ceiling: questions on which both teacher and student fail receive NCQ but no correct candidate. Extending the student's \emph{zone} beyond what the current teacher covers is, in our view, an important open problem this work leaves behind.

\section*{LLM Usage Statement}

We used GPT-5.6 Sol to check grammar and edit text we wrote, to cross-check the text against
the figures and tables, and to generate \cref{fig:concept}, a conceptual illustration that
gives readers an intuitive overview. We used Claude Opus 5.5 to find bugs in code
we wrote. The authors designed the method, ran every experiment, and reviewed all outputs.

\bibliography{ref}

\clearpage
\appendix
\crefalias{section}{appendix}
\crefalias{subsection}{subappendix}
\section{Full Related Work}
\label{sec:related_full}

This appendix expands \cref{sec:related} along its four lines of work.
\Cref{sec:appendix:baselines} gives the executed step of every baseline and control, and
the other compared rows are in \crefrange{tab:hint_methods}{tab:compute_matched} and
\cref{sec:appendix:additional_comparisons}.

\paragraph{Knowledge distillation.}
Knowledge distillation trains a student to match the logits, hidden states, or sampled
outputs of a teacher~\citep{hinton2015distilling,gou2021knowledge}.
Off-policy distillation trains on responses the teacher wrote, and on-policy distillation
has the teacher re-score the student's own responses~\citep{agarwal2024policy}.
Either way, the student must match the teacher's output
distribution~\citep{song2026survey,li2026rethinking}. This signal grows brittle as the
teacher--student gap widens~\citep{cho2019efficacy,mirzadeh2020improved,zhang-etal-2023-lifting,zhang2023towards},
inducing memorization of teacher answers and forgetting of general
capabilities~\citep{chu2025sft,chen2025sft,huan2025does}. Students of $3$B parameters or
fewer do not gain consistently from a strong reasoner~\citep{li2025small}, and weaker
models tend to imitate its surface verbosity rather than its reasoning
procedure~\citep{ren2026rethinking}. Self-distillation replaces the external
teacher with the same model, often given extra context such as a reference solution or
feedback~\citep{zhao2026selfdistilled,hubotter2026reinforcement,kim2026does}. The student,
whose prompt lacks this context, matches this self-teacher's output distribution. ZPPO also
uses extra context, but puts it in the student's own prompt and trains on the rollouts
sampled from both that prompt and the plain prompt under the binary reward (\cref{sec:appendix:teacher_alternatives}).

Several VLM methods combine distillation with RL. RIL~\citep{NEURIPS2025_e5849736}
pairs RL with adversarial imitation of teacher responses,
Masters~\citep{Lee_2026_CVPR_Masters} masks the teacher's weights before offline RL, and
PRISM~\citep{wang2026prism} adds a black-box adversarial distillation stage between SFT and
RL. All three keep the teacher in the loss. ZPPO instead keeps the teacher in the prompt.
\Cref{tab:curriculum}
compares ZPPO with RIL, Masters, and PRISM.

\paragraph{RL post-training and zero-variance groups.}
PPO, GRPO, DAPO, and related recipes train on the student's own
rollouts~\citep{schulman2017proximal,shao2024deepseekmath,yu2025dapo,guo2025deepseek,hu2025reinforce++}.
When the baseline is the group mean reward, an all-wrong group has zero advantage and
gives no learning signal. Methods that target such
groups without changing the prompt filter them out, add dense rewards, reshape advantages, interleave supervised
updates, or feed the teacher back into the gradient. DAPO's dynamic sampling drops
all-correct and all-wrong groups and resamples to refill the batch~\citep{yu2025dapo}
(DAPO$^{\dagger}$ in \cref{tab:compute_matched}). Dense process rewards give partial credit
inside a rollout: PRIME~\citep{cui2025process} learns an implicit process reward online,
StepGRPO~\citep{zhang2025r1vl} adds rule-based step-wise rewards,
PROGRS~\citep{rezaei2026process} ranks the wrong rollouts of a group with a centered process
bonus, and ProcessThinker~\citep{wu2026processthinker} scores each step by the success rate
of its continuations (\cref{sec:appendix:mmrl}). RL-ZVP~\citep{le2026no} shapes the
advantage of zero-variance groups with token-level entropy, and LENS~\citep{feng2025don}
penalizes a wrong answer more as the student's confidence in it grows.
ReLIFT~\citep{ma2026learning} interleaves RL with fine-tuning on teacher solutions to the
hardest questions, and RSTG~\citep{han2026distill} applies distillation only to all-wrong
groups and fine-tunes on teacher-generated correct trajectories. \Cref{tab:mistakes} and
\cref{sec:appendix:zerovariance} compare these four methods. Hybrid RL$+$distillation
feeds the teacher back into the gradient: KDRL~\citep{xu2025kdrl} adds a KL term toward the
teacher, RLKD~\citep{xu2026rlkd} adds a reward from the teacher's reasoning structure, and
RLAD~\citep{zhang2026reinforcement} anchors its trust region to a teacher--old-policy mixture
(\cref{sec:appendix:teacher_alternatives}). LUFFY~\citep{NEURIPS2025_a9d5c33e} adds
off-policy teacher traces to each GRPO group, with regularized importance sampling
(\cref{tab:prefix_methods}). Other methods add a perception
reward~\citep{xiao2026perception}, a consistency bonus~\citep{chen2026grpo}, or role-aware
token weights~\citep{jiang2026structured} for multimodal reasoning (\cref{sec:appendix:mmrl}).
GSPO~\citep{zheng2025groupsequencepolicyoptimization}, CISPO~\citep{chen2025minimax}, and
DISPO~\citep{karaman2026dispo} change the policy objective (\cref{sec:appendix:backbones}).
All of the methods above change the sampling, the reward, the loss, or the target, never
the student's prompt.

\paragraph{Curricula and prompt-side guidance.}
These methods act on the prompt. Curricula adapt the training distribution to what the
student can currently do~\citep{zhang2025clpo,mahrooghi2026goldilocks}.
AdaCuRL~\citep{li2026adacurl} also revisits past data, and
ADCL/EGSR~\citep{zhang2025learninglike} adds expert solutions that the model reformulates in
its own words. ZPPO's prompt replay buffer is itself a difficulty-based curriculum, and
GRPO$^{\dagger}$ keeps this buffer without BCQ and NCQ. In
\cref{tab:curriculum}, all four curricula fall below
GRPO$^{\dagger}\!+\!$BCQ in every column (\cref{tab:hint_methods}). Hint methods
put a cue in the prompt~\citep{nath2025adaptive,chen2026nudging,xia2026learning,wu2025thought}.
In SAGE~\citep{liao2026selfhinting}, the model writes its own hints, and in
HINT~\citep{wang2026don}, the cue stays abstract and leaves out the answer. A hint gives
a correct cue but no wrong alternative to weigh it against. Prefix methods hand over
the opening steps of a reference
solution~\citep{zhang2025bread,zhang2025stephint,li2025staying,huang2025blending,huang2025boosting}.
\Cref{tab:hint_methods,tab:prefix_methods} compare ZPPO with both families and with our
Hint and Prefix controls (\cref{sec:appendix:baselines}). In both tables, GRPO$^{\dagger}\!+\!$BCQ is above
every prior method and both controls in every column.

\paragraph{Learning from mistakes.}
LeMa~\citep{an2023learning} fine-tunes on corrected solutions, NAT~\citep{wang2024learning}
fine-tunes on correct and wrong trajectories labeled by an input tag, and
ORION~\citep{wu2026enhancing} fine-tunes on teacher chains of thought refined with the
student's own errors. \Cref{tab:mistakes} compares NCQ with ReLIFT, RL-ZVP, and LENS, and
with two controls that turn error information into supervised targets:
Correction-SFT (built on LeMa and NAT) and Error-reflection-SFT (built on ORION). In
SocraticPO~\citep{liu2026socraticpo}, the teacher writes guidance on a wrong attempt, and
the student answers again with the attempt and the guidance in context, over several
rounds. A correct answer after guidance gets a decayed reward. Our Hint control
(\cref{sec:appendix:baselines}) takes two of its elements, teacher-written guidance in the
prompt and an on-policy update under that prompt, and BCQ is above this control in every
column of \cref{tab:hint_methods}.

\paragraph{What ZPPO adds.}
SocraticPO and the hint methods also put guidance in the context. ZPPO differs in three
ways.
\textbf{(i)~What the context holds.} BCQ pairs a correct teacher response with a wrong
student response and does not say which is correct. SocraticPO holds a failed attempt and
the teacher's guidance on it, and a hint holds a correct cue without a wrong alternative.
NCQ lists the student's teacher-compressed wrong rollouts with their distinct
answers marked wrong, and adds no teacher solution or guidance on them.
\textbf{(ii)~How the context enters training.} Each BCQ or NCQ prompt gets its own group of
fresh student rollouts under the full binary reward, beside the plain group. SocraticPO
instead continues one trajectory and decays the reward of answers given after guidance.
\textbf{(iii)~When the context is used.} The prompt replay buffer keeps each hard question
eligible for re-sampling and for fresh candidates on each visit, until it graduates or is
evicted. Replay and reformulation are complementary (\cref{sec:experiments:ablation}).

\section{Training dataset and benchmarks}
\label{sec:appendix:dataset}

\subsection{ZPPO-$77$K training corpus}
\label[appendix]{sec:appendix:dataset:corpus}

We post-train each student as a VLM on ZPPO-$77$K, a multimodal RL dataset of
image--question pairs (\cref{sec:experiments}). Each sample is a triple
$(x_{\rm image},x_{\rm text},a^{\star})$: the image, the question, and the reference
answer that the reward grades against. We build ZPPO-$77$K from two public datasets:
\begin{itemize}[leftmargin=12pt,topsep=2pt,itemsep=2pt]
    \item \texttt{zlab-princeton/Vero-600k}~\citep{sarch2026open}\footnote{HF: \href{https://huggingface.co/datasets/zlab-princeton/Vero-600k}{\texttt{zlab-princeton/Vero-600k}}.}: a $600$K-sample corpus. We use the $29$ sub-datasets listed below.
    \item \nolinkurl{OpenDataArena/MMFineReason-SFT-586K}~\citep{lin2026mmfinereason}\footnote{HF: \href{https://huggingface.co/datasets/OpenDataArena/MMFineReason-SFT-586K-Qwen3-VL-235B-Thinking}{\nolinkurl{OpenDataArena/MMFineReason-SFT-586K-Qwen3-VL-235B-Thinking}}.}: a $586$K-sample corpus with reasoning traces from Qwen3-VL-235B-A22B-Thinking. We keep only the question, the image, and the reference answer, so none of its traces enters any method's loss, target, or prompt. Its released difficulty score is the pass rate of Qwen3-VL-4B-Thinking over four attempts.
\end{itemize}
We split the sub-datasets into two tiers, each with its own cap per sub-dataset:
\begin{itemize}[leftmargin=12pt,topsep=2pt,itemsep=2pt]
    \item \textbf{Tier 1} (cap $2{,}800$ samples per sub-dataset): \texttt{ai2d}~\citep{kembhavi2016diagram}, \texttt{geo170k}~\citep{gao2023g}, \texttt{geomverse}~\citep{kazemi2023geomverse}, \texttt{geoqa\_plus}~\citep{cao2022augmented}, \texttt{mmk12}~\citep{meng2025mmeureka}, \texttt{cosyn\_math}, \texttt{cosyn\_chart}, \texttt{cosyn\_diagram}, \texttt{cosyn\_table}~\citep{yang2025cosyn}, \texttt{wemath20}~\citep{qiao2026wemath2}, \texttt{tqa}~\citep{kembhavi2017you}, \texttt{visualwebinstruct}~\citep{jia2025visualwebinstruct}, \texttt{chartqa}~\citep{masry2022chartqa}, \texttt{arxivqa}~\citep{li2024multimodal}; on the MMFineReason side, \texttt{MMR1}~\citep{leng2025mmr1}, \texttt{Euclid30K}~\citep{lian2026euclid}, \texttt{Geo170K}, \texttt{Geo3K}/\texttt{Geometry3K}~\citep{lu2021inter}, \texttt{mm-openr1}, \texttt{WeMath2}, \texttt{VisualWebInstruct}, \texttt{BMMR}~\citep{xi2025bmmr}, \texttt{TQA}, \texttt{AI2D}, \texttt{ScienceQA}~\citep{lu2022learn}, \texttt{ViRL39K}~\citep{wang2025vlrethinker}.
    \item \textbf{Tier 2} (cap $1{,}400$ samples per sub-dataset): \texttt{pathvqa}~\citep{he2020pathvqa}, \texttt{vqarad}~\citep{lau2018vqarad}, \texttt{raven}~\citep{zhang2019raven}, \texttt{evochart}~\citep{huang2024evochart}, \texttt{infographic\_vqa}~\citep{mathew2022infographicvqa}, \texttt{aokvqa}~\citep{schwenk2022aokvqa}, \texttt{gqa}~\citep{hudson2019gqa}, \texttt{iconqa}~\citep{lu2021iconqa}, \texttt{vqav2}~\citep{balanced_vqa_v2}, \texttt{visual7w}~\citep{zhu2016visual7w}, \texttt{tallyqa}~\citep{acharya2019tallyqa}, \texttt{pixelreasoner}~\citep{wang2025pixelreasoner}, \texttt{multihop}, \texttt{pixmo}~\citep{deitke2024molmo}, \texttt{visual\_probe}; on the MMFineReason side, \texttt{Raven}, \texttt{VisualSphinx}~\citep{feng2025visualsphinx}, \texttt{PuzzleQA}~\citep{chia2024puzzlevqa}, \texttt{LLaVA-CoT}~\citep{xu2024llavacot}, \texttt{Zebra-CoT}~\citep{li2025zebracot}.
\end{itemize}
A name map merges the sub-datasets that appear in both sources, and Vero takes priority.
From MMFineReason, we keep only the samples that Qwen3-VL-4B-Thinking solves in at most one of its four attempts. After the
tier caps, three
per-sample filters apply: answer length $\leq\!512$ characters, image side
$\min(h,w)\!\geq\!100$ pixels, and the $4{,}096$-token prompt budget of plain rollouts
(\cref{tab:appendix:hyperparameters}), which includes the image tokens and the
instruction of \cref{fig:prompt-format}. Every method trains on the same filtered dataset.

AI2D and InfoVQA appear both in the corpus and as VLM settings
(\cref{tab:appendix:vlm_benchmarks}): the corpus holds samples from their training splits,
and we evaluate them on their held-out splits. \texttt{wemath20}/\texttt{WeMath2} are We-Math 2.0, not the WeMath
benchmark.

\subsection{Evaluation benchmarks}
\label[appendix]{sec:appendix:benchmarks}

The $31$-benchmark suite of \cref{sec:experiments} has $16$ VLM, $10$ LLM, and
$5$ Video settings.
The LLM settings are text-only except HLE, which also includes a small share of image-bearing questions.
ZPPO-$77$K holds only image--question pairs, so the LLM and Video families lie beyond the
training corpus. Each family average (Avg) and the $15$-setting LLM$+$Video summary in
\cref{sec:intro} weight every setting equally.
\Cref{tab:appendix:vlm_benchmarks}, \cref{tab:appendix:llm_benchmarks}, and
\cref{tab:appendix:video_benchmarks} give each setting's abbreviation, full name,
scoring, and data source.

\textsc{Exact} is the \texttt{\textbackslash boxed\{\}} grader of
\cref{sec:appendix:hyperparameters:zppo}, and \textsc{Rule} is the benchmark's own matching
rule. \textsc{Exact+Judge} and \textsc{Rule+Judge} call the judge only when that scorer
gives $0$. \textsc{Det.} is a deterministic scorer with no judge. \textsc{BenchJudge} runs
the benchmark's own judge prompt on our judge checkpoint (\emph{Benchmark-specific judge
prompts} below).

Five suites report an aggregate score instead of per-item accuracy. DynaMath counts each of its
$501$ seeds as correct only when all $10$ variants are correct. VBlind averages the $7$
BlindTest tasks. CharXiv (validation split) is the mean of its Reasoning and Descriptive
accuracies. MM-Vet uses its graded $[0,1]$ score. OCR$^\text{EN}$ and OCR$^\text{ZH}$
are the official OCRBench v2 aggregates, which average the per-category scores within each
language. Their \textsc{Det.} scorer is the official OCRBench v2 evaluator. The
\textsc{Det.} scorer of InfoVQA is its official ANLS
metric~\citep{mathew2022infographicvqa}, which gives partial credit by normalized edit
distance.

\begin{table}[!htbp]
    \centering
    \caption{The $16$ VLM evaluation settings (\cref{tab:main} and \cref{tab:ablation}).}
    \label{tab:appendix:vlm_benchmarks}
    \resizebox{\linewidth}{!}{%
    \renewcommand{\arraystretch}{1.05}
    \begin{tabular}{l p{0.34\linewidth} l p{0.26\linewidth}}
        \toprule
        Abbreviation & Full name & Scoring & Data source (HF) \\
        \midrule
        AI2D                  & AI2 Diagrams~\citep{kembhavi2016diagram}                                  & \textsc{Exact+Judge} & \texttt{lmms-lab/ai2d} \\
        BabyV                 & BabyVision~\citep{chen2026babyvision}                                    & \textsc{Exact+Judge} & \texttt{UnipatAI/BabyVision} \\
        CharXiv               & CharXiv (\texttt{validation}; mean of Reasoning and Descriptive)~\citep{wang2024charxiv} & \textsc{BenchJudge} & \texttt{princeton-nlp/CharXiv} \\
        DynaM                 & DynaMath ($501$ seeds $\times$ $10$ variants; worst-case accuracy)~\citep{zou2024dynamath}& \textsc{Exact+Judge} & \texttt{DynaMath/DynaMath\_Sample} \\
        EmbSp                 & EmbSpatial-Bench~\citep{du2024embspatial}                              & \textsc{Exact+Judge} & \texttt{FlagEval/EmbSpatial-Bench} \\
        InfoVQA               & InfographicVQA~\citep{mathew2022infographicvqa}                                & \textsc{Det.} & \texttt{lmms-lab/DocVQA} (\texttt{InfographicVQA}) \\
        MVerse                & MathVerse (\texttt{testmini})~\citep{zhang2024mathverse}                & \textsc{Exact+Judge} & \texttt{CaraJ/MathVerse-lmmseval} \\
        MVision               & MathVision~\citep{wang2024measuring}                                    & \textsc{Exact+Judge} & \texttt{MathLLMs/MathVision} \\
        MVista                & MathVista (\texttt{testmini})~\citep{lu2023mathvista}                & \textsc{Rule+Judge} & \texttt{AI4Math/MathVista} \\
        MMMU$^\text{Pro}$     & MMMU-Pro (all $3$ configs)~\citep{yue2024mmmupro}            & \textsc{Exact+Judge} & \texttt{MMMU/MMMU\_Pro} \\
        MM-Vet                & MM-Vet (v$1$)~\citep{yu2023mm}                                  & \textsc{BenchJudge} & \texttt{lmms-lab/MMVet} \\
        OCR$^\text{EN}$       & OCRBench v2 (English aggregate of category scores)~\citep{fu2026ocrbench} & \textsc{Det.} & \texttt{ling99/OCRBench\_v2} \\
        OCR$^\text{ZH}$       & OCRBench v2 (Chinese aggregate of category scores)~\citep{fu2026ocrbench} & \textsc{Det.} & \texttt{ling99/OCRBench\_v2} \\
        VisP                  & VisualPuzzles~\citep{song2025visualpuzzles}                                 & \textsc{Exact+Judge} & \texttt{neulab/VisualPuzzles} \\
        VBlind                & VLMs-are-Blind (BlindTest, all $7$ tasks)~\citep{rahmanzadehgervi2024vision}                                & \textsc{Exact+Judge} & \texttt{XAI/vlmsareblind} \\
        WeMath                & WeMath (\texttt{testmini})~\citep{qiao2024we}                   & \textsc{Exact+Judge} & \texttt{We-Math/We-Math} \\
        \bottomrule
    \end{tabular}}
\end{table}

\begin{table}[!htbp]
    \centering
    \caption{The $10$ LLM evaluation settings (\cref{tab:generalization}, LLM block).}
    \label{tab:appendix:llm_benchmarks}
    \resizebox{\linewidth}{!}{%
    \renewcommand{\arraystretch}{1.05}
    \begin{tabular}{l p{0.34\linewidth} l p{0.26\linewidth}}
        \toprule
        Abbreviation & Full name & Scoring & Data source (HF) \\
        \midrule
        AIME25            & AIME 2025 (\texttt{AIME2025-I/II}, $30$ problems)                                        & \textsc{Exact+Judge} & \texttt{opencompass/AIME2025} \\
        AIME26            & AIME 2026 ($30$ problems)                                                                & \textsc{Exact+Judge} & \texttt{MathArena/aime\_2026} \\
        CEval             & C-Eval (\texttt{val}, all $52$ subjects)~\citep{huang2023c}            & \textsc{Exact+Judge} & \texttt{ceval/ceval-exam} \\
        GPQA-D            & GPQA-Diamond ($198$ rows)~\citep{rein2023gpqa}                   & \textsc{Exact+Judge} & \texttt{Idavidrein/gpqa} \\
        HLE               & Humanity's Last Exam (mostly text; some questions with images)~\citep{phan2025humanity} & \textsc{BenchJudge} & \texttt{cais/hle} \\
        IMO-AB            & IMO-AnswerBench~\citep{luong2025towards}                                    & \textsc{Exact+Judge} & \texttt{OpenEvals/IMO-AnswerBench} \\
        MMLU              & MMLU (\texttt{test}, all subjects)~\citep{hendrycks2020measuring}                 & \textsc{Exact+Judge} & \texttt{cais/mmlu} \\
        MMLU-Pro          & MMLU-Pro~\citep{wang2024mmlu}                                           & \textsc{Exact+Judge} & \texttt{TIGER-Lab/MMLU-Pro} \\
        MMLU-Rd           & MMLU-Redux 2.0 (all $57$ subjects)~\citep{gema2024are}                 & \textsc{Exact+Judge} & \texttt{edinburgh-dawg/mmlu-redux-2.0} \\
        MultiCh           & MultiChallenge~\citep{sirdeshmukh2025multichallenge}                                     & \textsc{BenchJudge} & \texttt{ScaleAI/MultiChallenge} \\
        \bottomrule
    \end{tabular}}
\end{table}

\begin{table}[!htbp]
    \centering
    \caption{The $5$ Video evaluation settings (\cref{tab:generalization}, Video block).}
    \label{tab:appendix:video_benchmarks}
    \resizebox{0.8\linewidth}{!}{%
    \renewcommand{\arraystretch}{1.05}
    \begin{tabular}{l p{0.34\linewidth} l p{0.26\linewidth}}
        \toprule
        Abbreviation & Full name & Scoring & Data source (HF) \\
        \midrule
        MMVU              & MMVU (\texttt{validation}, $1{,}000$ rows)~\citep{zhao2025mmvu}         & \textsc{BenchJudge} & \texttt{yale-nlp/MMVU} \\
        MVBench           & MVBench ($20$ task configs)~\citep{li2024mvbench}                        & \textsc{Rule+Judge} & \texttt{OpenGVLab/MVBench} \\
        VMME              & Video-MME (without subtitles)~\citep{fu2024video}                      & \textsc{Rule+Judge} & \texttt{lmms-lab/Video-MME} \\
        VMME$^\text{S}$   & Video-MME (with subtitles)~\citep{fu2024video}                         & \textsc{Rule+Judge} & \texttt{lmms-lab/Video-MME} \\
        VMMMU             & Video-MMMU~\citep{hu2025video}                                         & \textsc{Rule+Judge} & \texttt{lmms-lab/VideoMMMU} \\
        \bottomrule
    \end{tabular}}
\end{table}

\paragraph{Decoding configuration.}
Training-time student and teacher rollouts sample at \texttt{temperature=1.0} and
\texttt{top-p=1.0}, with no top-$k$ and no penalty terms. Evaluation uses one
configuration for every setting, method, and Qwen3.5 checkpoint:
\texttt{temperature=0.6},
\texttt{top-p=0.95},
\texttt{top-k=20},
\texttt{min-p=0},
\texttt{presence-penalty=1.5},
\texttt{repetition-penalty=1.0},
\texttt{max-new-tokens=12288},
\texttt{min-pixels=256$\times$32$\times$32},
\texttt{max-pixels=1280$\times$32$\times$32}.
Video inputs use a per-suite frame budget: $16$ frames for MVBench, $32$ for
Video-MMMU, and up to $64$ at $1$~fps for MMVU. Video-MME takes $16$ to $32$ frames, set by
its clip-length split (short, medium, long). Every method and checkpoint uses the same
frame budgets.

\paragraph{Prompt and judge templates.}
For every setting, we replace the benchmark's own instructions on reasoning and answer
format (e.g., ``Think step by step'' or ``put your final answer within
\verb|\boxed{}|'') with the training instruction of \cref{fig:prompt-format}. Evaluation
never uses a BCQ or NCQ prompt (\cref{sec:experiments:discussion}(iv)). Correction-SFT$^{\dagger}$ keeps its training-time \texttt{<correct>} input tag at evaluation
(\cref{sec:appendix:baselines}).
\begin{figure}[!htbp]
\begin{prompttemplatefig}
\small
\emph{You FIRST think about the reasoning process as an internal monologue and then provide the final answer. The reasoning process MUST BE enclosed within \texttt{<think>}\,\texttt{</think>} tags. The final answer MUST BE put in \texttt{\textbackslash boxed\{\}}.}
\end{prompttemplatefig}
\caption{Instruction appended once, after the question, at training and evaluation.}
\label{fig:prompt-format}
\end{figure}

The shared judge runs only after \textsc{Exact} or \textsc{Rule} gives $0$. It reads the extracted
snippet when a \texttt{\textbackslash boxed\{\}} span parses but does not match, and, as a
fallback, the full response when none parses. The judge is Qwen3.5-$27$B-FP8, the same checkpoint that
generates the teacher text. It decodes at \texttt{temperature=0.0}, \texttt{top-p=1.0}, and
\texttt{max-new-tokens=512}, with the template of \cref{fig:prompt-judge-shared}. Its
parser accepts only \texttt{\{"verdict": "correct"\}} or \texttt{\{"verdict": "wrong"\}}
and scores any other output $0$. The checkpoint, template, parser, and fallback are the
same for every method and for the untrained base. The teacher-size study
(\cref{sec:appendix:teacher_capability}) changes only the teacher and keeps this judge.
On the settings that use this template, \cref{sec:appendix:judges} rescores the same responses with seven other judges.
\begin{figure}[!htbp]
\begin{prompttemplatefig}
\small
\emph{You are an expert judge evaluating whether a model's response is correct according to the ground truth.}

\textbf{\emph{\#\# Your Role (READ FIRST)}}\\
\emph{You are a COMPARATOR, not a solver. Your ONLY job is to decide how well the model's response matches the provided GROUND TRUTH.}
\begin{itemize}[leftmargin=12pt,topsep=1pt,itemsep=1pt]
    \item \emph{Do NOT attempt to solve the problem yourself.}
    \item \emph{Do NOT infer or derive what the ``real'' answer should be.}
    \item \emph{Do NOT second-guess the ground truth, even if it seems wrong.}
    \item \emph{The ``Model Response'' section below MAY be either (a) the model's full trajectory including any \texttt{<think>...</think>} reasoning and a concluding \texttt{\textbackslash boxed\{...\}} answer, OR (b) a short EXTRACTED snippet that was already pulled out as the candidate final answer (e.g.\ a bare letter like \texttt{B}, a number like \texttt{42}, an expression like \texttt{x = 5/3}). Treat both forms uniformly: the snippet (when given) IS the model's committed final answer, and a full trajectory's final answer is the content inside the last \texttt{\textbackslash boxed\{...\}} or the concluding sentence AFTER any \texttt{</think>} tag.}
    \item \emph{If the final answer is missing due to answer format error, truncation, or other reasons, look inside thinking or reasoning.}
    \item \emph{You should OUTPUT JSON with the format \texttt{\{"verdict": "correct" or "wrong"\}} based on the model's response and the ground truth.}
\end{itemize}

\textbf{\emph{\#\# Question}}\\
\emph{\texttt{\{\{ question \}\}}}

\textbf{\emph{\#\# Answer Options}} (rendered only when non-empty)\\
\emph{\texttt{\{\{ options \}\}}}

\textbf{\emph{\#\# Model Response}}\\
\emph{\texttt{\{\{ model\_response \}\}}}

\textbf{\emph{\#\# Ground Truth Answer}}\\
\emph{\texttt{\{\{ ground\_truth \}\}}}

\emph{Output ONLY a valid JSON object with exactly this format (no prose, no code fences, no trailing commentary):} \texttt{\{"verdict": "correct" or "wrong"\}}
\end{prompttemplatefig}
\caption{Shared judge template.}
\label{fig:prompt-judge-shared}
\end{figure}

\paragraph{Benchmark-specific judge prompts.}
Five suites keep their own judge prompts, which we run on our judge checkpoint.
MM-Vet~\citep{yu2023mm} grades on $[0,1]$ against few-shot exemplars with the dataset's
\texttt{<AND>}/\texttt{<OR>} tags. MultiChallenge~\citep{sirdeshmukh2025multichallenge}
applies its YES/NO verifier to the text after \texttt{</think>}, or to the boxed answer if
that text is empty. MMVU~\citep{zhao2025mmvu} uses separate
templates for multiple-choice and open-ended questions. HLE~\citep{phan2025humanity} uses
the CAIS template, which extracts the final answer and returns yes or no with a
confidence. CharXiv~\citep{wang2024charxiv} uses its Reasoning template with four rubrics
and its Descriptive template with seven question classes, each returning a binary score.

\section{ZPPO algorithm}
\label{sec:appendix:algorithm}

\Cref{alg:zppo} gives one ZPPO training step, where $x$ denotes a question.
Training is on-policy in the response-level sense of \cref{sec:method:bcqncq}:
every gradient-counted token is sampled by the student. Each group is sampled under its
own prompt by the student policy of the current rollout step ($\pi_{\rm old}$ in
\cref{eq:ppo-token-ratio}). The frozen teacher $\pi_{\rm T}$
reaches the student through the prompt: its correct rollouts form the BCQ pool, and it
compresses every BCQ and NCQ candidate. The same checkpoint (Qwen3.5-$27$B-FP8) also judges free-form answers
for every method's binary reward (\cref{sec:appendix:hyperparameters:zppo,sec:appendix:benchmarks}). No teacher
response is a gradient target.
We call $x$ a \emph{hard question} when its plain-rollout accuracy
$\bar{r}_x$ is below $\tau$. Every run outside the sweep of
\cref{tab:appendix:tau_sweep} uses $\tau\!=\!0.5$.

ZPPO is built on GRPO~\citep{shao2024deepseekmath}: a group-relative advantage and a
PPO-style clipped surrogate~\citep{schulman2017proximal}. It adds three DAPO
ingredients~\citep{yu2025dapo}: clip-higher with DAPO's
$(\epsilon_{\rm low},\epsilon_{\rm high})\!=\!(0.20,0.28)$, a token-level policy gradient
loss, and no KL penalty. It disables DAPO's dynamic sampling, which would drop the
all-wrong groups that BCQ and NCQ reformulate. It uses the two-step REINFORCE++
estimator~\citep{hu2025reinforce++} of \crefrange{eq:adv-step1}{eq:adv-step2} with
\emph{Norm w/o Zero}, which excludes zero-advantage groups from the batch statistics, and
runs $I\!=\!4$ iterations per step. The dual clip of
\citet{ye2020mastering} with $c_{\rm dual}\!=\!10$ (\cref{eq:dualclip}) bounds the
surrogate of a negative-advantage token. GRPO and GRPO$^{\dagger}$ use all of these
settings (\cref{sec:appendix:baselines}), and DAPO$^{\dagger}$ adds dynamic sampling to
GRPO$^{\dagger}$ (\cref{tab:compute_matched}).

\begin{algorithm}[!htbp]
\small
\SetAlgoLined
\DontPrintSemicolon
\KwIn{Student $\pi_\theta$, frozen teacher $\pi_{\rm T}$, dataset $\mathcal{D}$, prompt replay buffer $\mathcal{B}$; $G_{\rm S}, G_{\rm T}, \rho_{\rm replay}, \rho_{\rm aug}, |\mathcal{B}|_{\max}, \tau\!=\!0.5$.}
\KwOut{Updated $\pi_\theta$, $\mathcal{B}$.}
\textbf{// 1. Plain rollouts.}\;
$X \leftarrow$ new questions $X_{\rm new}\!\sim\!\mathcal{D}$ and up to $\rho_{\rm replay}|X_{\rm new}|$ questions replayed from $\mathcal{B}$\;
\lForEach{$x \in X$}{draw $G_{\rm S}$ rollouts from $\pi_\theta(\cdot|x)$, score them, and compute $\bar{r}_x$}
\textbf{// 2. Hard questions.}\;
$X_{\rm aug} \leftarrow$ the at most $\rho_{\rm aug}|X_{\rm new}|$ hardest $x \in X$ with $\bar{r}_x\!<\!\tau$\;
\textbf{// 3. Teacher rollouts, drawn in parallel with Step~1.}\;
\lForEach{$x \in X$}{draw $G_{\rm T}$ rollouts from $\pi_{\rm T}(\cdot|x)$ and keep the correct ones}
\textbf{// 4. BCQ and NCQ prompts.}\;
\ForEach{$x \in X_{\rm aug}$}{
    \lIf{a teacher rollout on $x$ is correct and a wrong student rollout on $x$ parses}{$x_{\rm BCQ} \leftarrow x$ $+$ one correct teacher and one parsed wrong student rollout, compressed, shuffled, and unlabeled}
    \lIf{a wrong student rollout on $x$ parses}{$x_{\rm NCQ} \leftarrow x$ $+$ every parsed wrong student rollout, compressed and marked wrong}
}
\textbf{// 5. Reformulated rollouts.}\;
keep at most $\rho_{\rm aug}|X_{\rm new}|$ BCQ and NCQ prompts, hardest first, and draw $G_{\rm S}$ rollouts on each\;
\textbf{// 6. Update and buffer refresh.}\;
update $\pi_\theta$ with the GRPO objective on every plain, BCQ, and NCQ group\;
refresh $\mathcal{B}$ from the plain-rollout $\bar{r}_x$: admit $x$ with $\bar{r}_x\!<\!\tau$, graduate $x$ with $\bar{r}_x\!\geq\!\tau$, and FIFO-evict beyond $|\mathcal{B}|_{\max}$\;
\caption{One ZPPO training step.}
\label{alg:zppo}
\end{algorithm}

\begin{table}[!htbp]
    \centering
    \caption{ZPPO at $2$B under five hard-question thresholds $\tau$, averaged over the $10$ LLM, $16$ VLM, and $5$ Video benchmarks. Each ZPPO row also uses its $\tau$ for buffer admission and graduation (Step~6 of \cref{alg:zppo}). GRPO has no $\tau$, and GRPO$^{\dagger}$ uses $\tau\!=\!0.5$.}
    \label{tab:appendix:tau_sweep}
    \resizebox{0.45\linewidth}{!}{
    \setlength{\tabcolsep}{3mm}
    \renewcommand{\arraystretch}{1.05}
    \begin{tabular}{llccc}
        \toprule
        Method & $\tau$ & LLM & VLM & Video \\
        \midrule
        Base     & --      & $45.3$ & $56.8$ & $60.6$ \\
        GRPO     & --      & $46.7$ & $58.7$ & $61.2$ \\
        GRPO$^{\dagger}$ & $0.5$ & $47.4$ & $59.2$ & $61.9$ \\
        \midrule
        \multirow{5}{*}{ZPPO}
                 & $0.1$   & $47.9$     & $59.9$     & $62.1$     \\
                 & $0.3$   & $49.8$     & $61.5$     & $62.5$     \\
                 & \cellcolor{colorful}$0.5$ & \cellcolor{colorful}$50.4$ & \cellcolor{colorful}$62.0$ & \cellcolor{colorful}$63.2$ \\
                 & $0.7$   & $50.5$     & $61.4$     & $63.4$     \\
                 & $0.9$   & $50.1$     & $62.3$     & $62.8$     \\
        \bottomrule
    \end{tabular}
    }
\end{table}

\paragraph{Properties of the loop.}
BCQ needs a correct teacher rollout and a parsed wrong student rollout, so a question the
teacher fails on gets NCQ only (\cref{fig:teacher_scale} varies the teacher size). NCQ needs
only a parsed wrong student rollout, so Step~4 builds an NCQ instance for every question in
$X_{\rm aug}$ that has one. Each visit draws
new teacher rollouts and takes the student candidates from the current
plain group, so the candidates are fresh on every visit. Only plain rollouts drive
admission and graduation, and the buffer stores only the question $x$.

\paragraph{How a reformulated group enters the loss.}
Plain, BCQ, and NCQ groups enter the same loss: they share the clipped surrogate and the
optimizer mini-batches, each under the prompt that sampled it. A BCQ or NCQ prompt adds a
further group, and the plain group of the same question stays in the update. The teacher
text in a reformulated prompt is input only, and the only ratio is the PPO ratio of
\cref{eq:ppo-token-ratio}, between the current parameters and those that sampled the
group. A teacher rollout placed in a group would instead need an off-policy correction,
as in LUFFY~\citep{NEURIPS2025_a9d5c33e}. Evaluation uses the plain prompt
(\cref{sec:experiments:discussion}).

\paragraph{What one rollout step contains.}
One step draws $384$ new questions and up to $96$ replayed ones, and keeps at most $96$
BCQ and NCQ instances (\cref{tab:appendix:hyperparameters}). This gives at most $576$
groups of $G_{\rm S}\!=\!8$ rollouts, so BCQ and NCQ are at most one group in five, or one
in six with full replay. Step~2 caps the hard \emph{questions} and Step~5 caps the
\emph{instances}, each at $96$, so a question with both a BCQ and an NCQ instance fills
two slots. All groups are split across the $I\!=\!4$
optimizer mini-batches. \Cref{sec:appendix:replay_matched} raises the replay fraction of
GRPO$^{\dagger}$ to $\rho_{\rm replay}\!=\!0.50$, which gives it the same $576$-group cap.

\paragraph{What \cref{fig:system}a reports.}
\Cref{fig:system}a covers all $200$ rollout steps at each scale, before the
caps. \emph{Easy} counts questions with $\bar{r}_x\!\ge\!0.5$, and \emph{BCQ} and
\emph{NCQ} count the reformulations available on hard questions. A hard question can yield both,
so each share is a fraction of the summed count. Only BCQ needs the teacher to solve the
question, so its share falls with scale while the NCQ share stays near one third.

\paragraph{Two-step advantage estimator (REINFORCE++).}
We use \emph{REINFORCE++ w/ Baseline}~\citep{hu2025reinforce++}. Step~1 subtracts the
group mean, and Step~2 divides by a standard deviation taken across groups. We take it
over one optimizer mini-batch and only over its non-trivial groups (\emph{Norm w/o Zero}). Let $\mathcal{G}$
be the groups of one mini-batch, plain, BCQ, and NCQ alike, $r^{(g,i)}$ the $i$-th reward
of group $g$, $\bar{r}_g$ and $\mathrm{std}_g$ its mean and standard deviation, and
$\mathcal{G}^{\star}\!=\!\{g\!:\!\mathrm{std}_g\!>\!0\}$ the non-trivial groups. All-wrong
and all-correct groups are trivial. Step~1 is
\begin{equation}
A'^{(g,i)} \;=\; r^{(g,i)} - \bar{r}_g,
\label{eq:adv-step1}
\end{equation}
which is zero on every trivial group, and Step~2 normalizes over a subset
$\mathcal{S}\!\subseteq\!\mathcal{G}$ with population statistics and a small
$\epsilon\!>\!0$:
\begin{align}
A^{(g,i)} &\;=\; \frac{A'^{(g,i)} - \mu_{\mathcal{S}}}{\sigma_{\mathcal{S}} + \epsilon}, \nonumber \\
\mu_{\mathcal{S}} &\;=\; \mathrm{mean}_{(g,i):g\in\mathcal{S}}\,A'^{(g,i)}, \nonumber \\
\sigma_{\mathcal{S}} &\;=\; \mathrm{std}_{(g,i):g\in\mathcal{S}}\,A'^{(g,i)}.
\label{eq:adv-step2}
\end{align}
Every group sums to zero after Step~1, so $\mu_{\mathcal{S}}\!=\!0$. The three settings
of \cref{sec:experiments:recipe}(ii) differ only in Step~2. \emph{No norm} skips it,
which is \cref{eq:grpo-adv} without its denominator. \emph{Norm w/o Zero}, the
default of ZPPO, GRPO, and GRPO$^{\dagger}$, uses $\mathcal{S}\!=\!\mathcal{G}^{\star}$, so trivial groups keep $A\!=\!0$.
\emph{Norm w/ Zero} uses $\mathcal{S}\!=\!\mathcal{G}$. With
$f_{\rm nt}\!=\!|\mathcal{G}^{\star}|/|\mathcal{G}|$ the fraction of non-trivial groups,
the zeros of the trivial groups shrink the standard deviation to
$\sigma_{\mathcal{G}}\!=\!\sqrt{f_{\rm nt}}\,\sigma_{\mathcal{G}^{\star}}$,
so every non-trivial advantage is rescaled,
\begin{equation}
A^{(g,i)}_{\text{w/Zero}}
  \;=\; \frac{\sigma_{\mathcal{G}^{\star}}+\epsilon}{\sqrt{f_{\rm nt}}\,\sigma_{\mathcal{G}^{\star}}+\epsilon}\,A^{(g,i)}_{\text{w/o\,Zero}} \quad (g\in\mathcal{G}^{\star}),
\label{eq:wzero-amplify}
\end{equation}
by a factor in $[1,1/\sqrt{f_{\rm nt}}]$ recomputed on each mini-batch. AdamW cancels a
scale that is the same on every mini-batch, but not one that changes with $f_{\rm nt}$.
The update size relative to \emph{Norm w/o Zero} therefore depends on how many trivial groups a mini-batch holds, and it
is largest where the gradient comes from the fewest rollouts and is the noisiest. The
buffer raises the share of hard questions in a mini-batch, so $f_{\rm nt}$ varies across mini-batches.

\paragraph{Per-token clipped surrogate.}
For response token $t$, the ratio to the policy that sampled the
rollout~\citep{schulman2017proximal} is
\begin{equation}
\rho_t=\frac{\pi_\theta(y_t\mid x,y_{<t})}{\pi_{\rm old}(y_t\mid x,y_{<t})},
\label{eq:ppo-token-ratio}
\end{equation}
where $x$ is the reformulated prompt for a BCQ or NCQ group. Only student response tokens
enter the loss, which is aggregated over tokens. With the rollout's advantage $A$ from
\cref{eq:adv-step2} and
$q(\rho_t,A)=\min\!\big(\rho_t A,\,\mathrm{clip}(\rho_t,1\!-\!\epsilon_{\rm low},1\!+\!\epsilon_{\rm high})A\big)$,
the surrogate adds the dual clip of \citet{ye2020mastering}:
\begin{equation}
\ell(\rho_t,A)=
\begin{cases}
\max\!\big(q(\rho_t,A),\,c_{\rm dual}A\big), & A<0,\\
q(\rho_t,A), & A\geq 0.
\end{cases}
\label{eq:dualclip}
\end{equation}

\paragraph{The hard-question threshold.}
With binary rewards, the variance of a group is fixed by its mean,
\begin{equation}
\mathrm{std}_g^{2} \;=\; \frac{1}{G_{\rm S}}\sum_{i=1}^{G_{\rm S}}\bigl(r^{(g,i)}-\bar{r}_g\bigr)^{2} \;=\; \bar{r}_g\,\bigl(1-\bar{r}_g\bigr),
\label{eq:group-var}
\end{equation}
which is zero at $\bar{r}_g\!\in\!\{0,1\}$ and peaks at $0.5$. \Cref{eq:adv-step2} divides
every group by the same $\sigma_{\mathcal{S}}$, so below $\tau\!=\!0.5$ the plain
group's signal shrinks as the question gets harder and vanishes at $\bar{r}_x\!=\!0$.
The threshold also admits partially solved questions
($\bar{r}_x\!\in\!\{\tfrac{1}{8},\tfrac{2}{8},\tfrac{3}{8}\}$), where a reformulated
group adds to an existing signal. Every ZPPO row of \cref{tab:appendix:tau_sweep} stays
above GRPO$^{\dagger}$ on all three families.

\section{Hyperparameters and baseline implementations}
\label{sec:appendix:hyperparameters}

\subsection{ZPPO hyperparameters}
\label[appendix]{sec:appendix:hyperparameters:zppo}

\Cref{tab:appendix:hyperparameters} lists the optimization, rollout, and replay settings
of ZPPO.

\begin{table}[!htbp]
    \centering
    \caption{Training hyperparameters of ZPPO. Settings shared with GRPO and GRPO$^{\dagger}$ apply to every RL row, and a row that changes a setting or adds a term names it. Distillation and supervised rows also split each rollout step's $384$ new prompts into $I\!=\!4$ optimizer mini-batches of $96$.}
    \label{tab:appendix:hyperparameters}
    \adjustbox{width=\linewidth,max totalheight=0.88\textheight}{%
    \renewcommand{\arraystretch}{0.88}
    \begin{tabular}{l l p{0.50\linewidth}}
        \toprule
        Group & Hyperparameter & Value \\
        \midrule
        \multirow{4}{*}{Student / teacher}
            & Student initialization                      & \texttt{Qwen/Qwen3.5-\{0.8,2,4,9\}B} (HF) \\
            & Teacher                                     & \texttt{Qwen/Qwen3.5-27B-FP8} (HF), frozen, for teacher rollouts, compression, and judging; \texttt{Qwen/Qwen3.5-27B} (HF) for the baselines that read teacher logits \\
            & GPUs per node (student / frozen models)     & $6\,/\,2$ in every run \\
            & Total GPUs                                  & $64$ NVIDIA H100-$80$\,GB ($8$ nodes $\times$ $8$) \\
        \midrule
        \multirow{11}{*}{Optimization}
            & Optimizer                                   & AdamW (\texttt{AnyPrecisionAdamW}) \\
            & Learning rate                               & $1.0\!\times\!10^{-6}$ (constant for the full run) \\
            & LR scheduler                                & constant, no warmup, no decay \\
            & AdamW $(\beta_1, \beta_2)$                  & $(0.9,\,0.999)$ \\
            & AdamW $\epsilon$                            & $10^{-8}$ \\
            & Weight decay                                & $0.1$ (decoupled) \\
            & Gradient clipping (max grad norm)           & $1.0$ \\
            & PPO inner epochs (over a global batch)      & $1$ (each global batch is used once) \\
            & Mixed precision                             & bfloat16 \\
            & KL penalty                                  & $0$ (following DAPO) \\
            & Dynamic sampling                            & disabled \\
        \midrule
        \multirow{7}{*}{Rollouts}
            & Student group size $G_{\rm S}$              & $8$ \\
            & Teacher group size $G_{\rm T}$ (BCQ pool)   & $4$ \\
            & Tensor-parallel size (student / teacher)    & $1\,/\,1$ \\
            & Sampling temperature (student / teacher)    & $1.0\,/\,1.0$ (training-time rollouts) \\
            & Top-$p$ (student / teacher)                 & $1.0\,/\,1.0$ (no nucleus truncation) \\
            & Max prompt length                           & $4{,}096$ tokens for plain prompts \\
            & Max response length (student / teacher)     & $12{,}288$ / $12{,}288$ tokens \\
        \midrule
        \multirow{6}{*}{GRPO backbone}
            & Reward                                      & binary $\in\{0,1\}$ (grader, then shared judge) \\
            & Advantage estimator                         & two-step, REINFORCE++ w/ Baseline~\citep{hu2025reinforce++} \\
            & Clip-lower $\epsilon_{\rm low}$             & $0.20$ \\
            & Clip-higher $\epsilon_{\rm high}$           & $0.28$ (DAPO clip-higher) \\
            & Dual clip ratio                             & $10.0$ \\
            & Loss aggregation                            & token-level (DAPO) \\
        \midrule
        \multirow{2}{*}{Recipe choices}
            & Iterations per step $I$                     & $4$ optimizer mini-batches per rollout step \\
            & Batch advantage normalization               & on (\emph{Norm w/o Zero}) \\
        \midrule
        \multirow{6}{*}{Replay \& BCQ/NCQ}
            & Hard-question threshold $\tau$              & $0.5$ on $\bar{r}_x$ (swept in \cref{tab:appendix:tau_sweep}) \\
            & Replay fraction $\rho_{\rm replay}$         & $0.25$ of $|X_{\rm new}|$ \\
            & Prompt replay buffer capacity $|\mathcal{B}|_{\max}$ & $10{,}000$ prompts \\
            & Augmentation fraction $\rho_{\rm aug}$      & $0.25$ of $|X_{\rm new}|$ \\
            & Eviction policy                             & FIFO once $|\mathcal{B}|\!>\!|\mathcal{B}|_{\max}$ \\
            & Candidate compression                      & max $512$ tokens per candidate \\
        \midrule
        \multirow{8}{*}{Schedule \& batching}
            & Data schedule                               & $200\!\times\!384=76{,}800$ draws from ZPPO-$77$K \\
            & Micro-batch size per device                 & $4$ \\
            & Global batch size                            & $96$ \emph{new} prompts per update ($4$ accum.\ $\times\,4$ micro $\times\,48$ GPUs $/\,G_{\rm S}$); up to $144$ with replayed and BCQ/NCQ prompts \\
            & Rollout batch size                          & $384$ new, up to $96$ replayed, up to $96$ BCQ/NCQ \\
            & Gradient accumulation per update            & $4$ per device at $96$ prompts, up to $6$ at $144$ \\
            & Total rollout steps per run                 & $200$ \\
            & Optimizer mini-batches per run              & $800$ ($I\!\cdot\!200$ over the $200$ rollout steps) \\
            & Padding-free training                       & on \\
        \bottomrule
    \end{tabular}}
\end{table}

\paragraph{What the ``Base'' rows are.}
Every Qwen3.5 run starts from the released checkpoint of each size, which is natively
multimodal and already post-trained by its developers. The \emph{Base} row is this
checkpoint before our post-training, evaluated as in \cref{sec:appendix:benchmarks}.

\paragraph{The binary reward.}
Each answer is first graded by
\href{https://github.com/hiyouga/MathRuler}{\texttt{mathruler.grader}}:
\texttt{extract\_boxed\_content} reads the last \texttt{\textbackslash boxed\{\dots\}}
span, and \texttt{grade\_answer} compares it with the reference answer $a^{\star}$,
accepting LaTeX and numerical equivalents such as \texttt{1/2} and \texttt{0.5}. If the
comparison fails, it is retried once after removing a degree or percent sign that both
sides carry. If it still fails, the answer goes to the shared LLM-as-a-judge~\citep{zheng2023judging},
with the same setup as at evaluation (\cref{sec:appendix:benchmarks}). The judge enters
training only through the binary reward it assigns, never as a gradient target. Every
method uses this routing.

\paragraph{Candidate compression.}
Once the candidates are chosen (\cref{sec:method:bcqncq}), the frozen teacher compresses each one under
the prompt of \cref{fig:prompt-compression}, with the same $512$-token cap for teacher and
student candidates.

\begin{figure}[!htbp]
\begin{prompttemplatefig}
\small
\emph{Compress the response below into a summary (in 5 lines max).}

\emph{Rules:}
\begin{itemize}[leftmargin=12pt,topsep=1pt,itemsep=1pt]
    \item \emph{Response is in \texttt{<response>}...\texttt{</response>} tags}
    \item \emph{Summary should be in \texttt{<summary>}...\texttt{</summary>} tags and should be in 5 lines max}
    \item \emph{Keep ONLY the essential reasoning steps and the final answer}
    \item \emph{Remove ALL exploratory text, self-corrections, retries, and filler}
    \item \emph{Do NOT re-derive or add new information}
    \item \emph{End with the final answer in \texttt{\textbackslash boxed\{\}} format}
\end{itemize}
\emph{\texttt{<response>}\\
$\langle$\emph{candidate response to be compressed}: $y_{\rm T}^{(+)}$ or $y_{\rm S}^{(-)}$ for BCQ, $y_{\rm S}^{(-)}$ for NCQ$\rangle$\\
\texttt{</response>}}
\end{prompttemplatefig}
\caption{Candidate compression prompt. The frozen teacher applies it, with a $512$-token cap, to every BCQ and NCQ candidate before the candidate enters a prompt.}
\label{fig:prompt-compression}
\end{figure}

\paragraph{Reformulated prompts (BCQ and NCQ).}
\Cref{fig:prompt-bcq} and \cref{fig:prompt-ncq} give the verbatim templates of
\cref{sec:method:bcqncq} that build $x_{\rm BCQ}$ and $x_{\rm NCQ}$. A wrong rollout becomes a candidate only if its final
\texttt{\textbackslash boxed\{\}} span parses. BCQ's
two blocks carry no label, letter, or fixed order. In NCQ,
$1\!\leq\!K\!\leq\!G_{\rm S}$ is the number of parsed wrong rollouts and $M\!\leq\!K$ the number of
distinct parsed answers: the prompt lists each of the $M$ answers once, joined with
``\texttt{, }'', and keeps all $K$ compressed traces as \texttt{<candidate>} blocks.
\begin{figure}[!htbp]
\begin{prompttemplatefig}
\small
$\langle$\emph{original question} $x$$\rangle$

\emph{Here are two candidate responses in \texttt{<candidate> </candidate>} tags to the question above. One is correct and another is wrong. Use these as references to help you solve the problem.}\\
\texttt{<candidate>}\\
$\langle$\emph{candidate$_1$: compressed $y_{\rm T}^{(+)}$ or $y_{\rm S}^{(-)}$, shuffled}$\rangle$\\
\texttt{</candidate>}\\
\texttt{<candidate>}\\
$\langle$\emph{candidate$_2$: the other one}$\rangle$\\
\texttt{</candidate>}
\end{prompttemplatefig}
\caption{BCQ reformulated-prompt template. \Cref{sec:method:bcqncq} quotes its instruction in full. The two shuffled blocks hold one compressed correct teacher rollout and one compressed wrong student rollout.}
\label{fig:prompt-bcq}
\end{figure}
\begin{figure}[!htbp]
\begin{prompttemplatefig}
\small
$\langle$\emph{original question} $x$$\rangle$

\emph{The following answers are all WRONG: \texttt{\textbackslash boxed\{}$a_1$\texttt{\}}, \texttt{\textbackslash boxed\{}$a_2$\texttt{\}}, \dots, \texttt{\textbackslash boxed\{}$a_M$\texttt{\}}. Below are the incorrect reasoning processes in \texttt{<candidate> </candidate>} tags.}\\
\texttt{<candidate>}\\
$\langle$\emph{compressed wrong rollout $1$}$\rangle$\\
\texttt{</candidate>}\\
\texttt{<candidate>}\\
$\langle$\emph{compressed wrong rollout $2$}$\rangle$\\
\texttt{</candidate>}\\
\dots\\
\texttt{<candidate>}\\
$\langle$\emph{compressed wrong rollout $K$}$\rangle$\\
\texttt{</candidate>}
\end{prompttemplatefig}
\caption{NCQ reformulated-prompt template. The prompt lists $M$ distinct wrong answers and $K$ candidate blocks.}
\label{fig:prompt-ncq}
\end{figure}

The instruction of \cref{fig:prompt-format} is appended last to both templates. Plain prompts are filtered
to $4{,}096$ tokens (\cref{sec:appendix:dataset}), and each candidate is capped at $512$
tokens. The candidate text thus adds at most $1{,}024$ tokens for BCQ and $4{,}096$ for
NCQ.

\subsection{Baseline implementations}
\label[appendix]{sec:appendix:baselines}

We give the executed step of every baseline and control. Off-Distill, On-Distill, GRPO,
and their $^{\dagger}$ rows are the baselines. The Hint and Prefix controls,
Correction-SFT, and Error-reflection-SFT are controls built from elements of their cited
methods (\cref{sec:related_full}) and run on our shared pipeline. At the default budget,
all rows share the initialization, corpus, rollout sizes, constant learning rate, and $200$
rollout steps of \cref{tab:appendix:hyperparameters}, so each draws the same $76{,}800$ new
prompts. All rows also share the evaluation setup of \cref{sec:appendix:benchmarks},
except that Correction-SFT$^{\dagger}$ keeps its training-time \texttt{<correct>} tag at
evaluation. Each cited method in \cref{tab:hint_methods,tab:prefix_methods,tab:curriculum}
runs on the GRPO$^{\dagger}$ pipeline, prompt replay buffer included, and changes only what
that method changes (\cref{sec:related_full}). It runs at this default budget and uses the
$27$B teacher wherever it needs one.

The ablation rows run \cref{alg:zppo} with components switched off and nothing else
changed. GRPO$^{\dagger}\!+\!$BCQ skips the NCQ branch of Step~4, GRPO$^{\dagger}\!+\!$NCQ
skips the BCQ branch, and GRPO\,$+$\,Both keeps both branches at
$\rho_{\rm replay}\!=\!0$, so its hard questions come from the current batch alone.

\paragraph{Replay rule.} Every $^{\dagger}$ row draws
$\min(\rho_{\rm replay}|X_{\rm new}|,|\mathcal{B}|)$ replay prompts uniformly without
replacement \emph{before} the buffer refresh of \cref{alg:zppo}. The refresh reads
$\bar r_x$ from the plain rollouts alone, never from a BCQ, NCQ, hint, or prefix group.
Every algorithm below uses this rule, and $\operatorname{concat}(X_{\rm new},X_{\rm replay})$
joins the two prompt lists.

\paragraph{Hard questions used by the controls.} The Hint and Prefix controls,
Correction-SFT, and Error-reflection-SFT generate their teacher output online, on at most
the $\rho_{\rm aug}|X_{\rm new}|\!=\!96$ hardest questions with $\bar r_x\!<\!\tau$ per
step. A question with no usable teacher output adds no group or target.

\paragraph{Off-policy distillation.}
Off-Distill draws $G_{\rm T}\!=\!4$ fresh teacher rollouts per prompt, grades them, and
imitates the correct ones under the per-token generalized JSD of GKD~\citep{agarwal2024policy}
at $\beta\!=\!0.5$. As in all
training-time teacher generation, the rollouts come from Qwen3.5-$27$B-FP8, and the JSD reads the
teacher's token distributions from the unquantized Qwen3.5-$27$B. A prompt with no
correct rollout adds no target. No teacher rollout is cached across
replay visits (\cref{alg:offpolicy}).

\begin{algorithm}[!htbp]
\SetAlgoLined
\DontPrintSemicolon
\KwIn{Student $\pi_\theta$, frozen teacher $\pi_{\rm T}$, dataset $\mathcal{D}$, buffer $\mathcal{B}$ ($^{\dagger}$ only); $G_{\rm T}\!=\!4$, $G_{\rm S}$, $\beta\!=\!0.5$, $I$; $\rho_{\rm replay},|\mathcal{B}|_{\max},\tau$ ($^{\dagger}$ only).}
Sample $X_{\rm new}\!\sim\!\mathcal{D}$. For $^{\dagger}$, draw $X_{\rm replay}$ from $\mathcal{B}$ by the replay rule; otherwise set $X_{\rm replay}\!=\!\emptyset$. Set $X\!\leftarrow\!\operatorname{concat}(X_{\rm new},X_{\rm replay})$.\;
\ForEach{$x \in X$}{
    Draw $G_{\rm S}$ student rollouts $\{y_{\rm S}^{(g)}\}_{g=1}^{G_{\rm S}}\!\sim\!\pi_\theta(\cdot|x)$ and score them; the $^{\dagger}$ row keeps their mean $\bar r_x$.\;
    Draw $G_{\rm T}$ fresh teacher rollouts $\{y_{\rm T}^{(g)}\}_{g=1}^{G_{\rm T}}\!\sim\!\pi_{\rm T}(\cdot|x)$, grade them, and keep the $N_x$ correct ones.\;
    \If{$N_x>0$}{Accumulate the average per-token generalized JSD over the $N_x$ correct teacher rollouts.\;}
}
Split $X$ into $I$ optimizer mini-batches of $96$ new prompts each, as in the RL rows. Apply one AdamW update per mini-batch on its JSD, and skip a mini-batch with no correct teacher rollout.\;
\textbf{($^{\dagger}$ only)} Update $\mathcal{B}$ from $\bar r_x$: admit, graduate, and FIFO-evict.\;
\Return{$\pi_\theta$ (and $\mathcal{B}$ for $^{\dagger}$).}
\caption{Executed Off-Distill step. The student imitates online teacher rollouts filtered by correctness. The $^{\dagger}$ row adds prompt replay.}
\label{alg:offpolicy}
\end{algorithm}

\paragraph{On-policy distillation.}
On-Distill is the on-policy case of GKD~\citep{agarwal2024policy}. The frozen, unquantized
Qwen3.5-$27$B re-scores each of the $G_{\rm S}\!=\!8$ student rollouts token by token under
the same JSD at $\beta\!=\!0.5$. The $^{\dagger}$ row adds prompt replay only (\cref{alg:onpolicy}).
\Cref{tab:appendix:distill_lr} also runs both $^{\dagger}$ distillation rows at $10^{-5}$
and $5\!\times\!10^{-5}$. Both larger rates lower LLM and Video in every cell, further at
$5\!\times\!10^{-5}$ than at $10^{-5}$, and raise VLM by at most $0.5$~pp, in line with the
finding that a larger learning rate increases forgetting in strong-to-weak
distillation~\citep{piskorz2026policy}.

\begin{algorithm}[!htbp]
\SetAlgoLined
\DontPrintSemicolon
\KwIn{Student $\pi_\theta$, frozen teacher $\pi_{\rm T}$, dataset $\mathcal{D}$, buffer $\mathcal{B}$ ($^{\dagger}$ only); $G_{\rm S}$, $\beta\!=\!0.5$, $I$; $\rho_{\rm replay},|\mathcal{B}|_{\max},\tau$ ($^{\dagger}$ only).}
Sample $X_{\rm new}\!\sim\!\mathcal{D}$. For $^{\dagger}$, draw $X_{\rm replay}$ from $\mathcal{B}$ by the replay rule; otherwise set $X_{\rm replay}\!=\!\emptyset$. Set $X\!\leftarrow\!\operatorname{concat}(X_{\rm new},X_{\rm replay})$.\;
\ForEach{$x \in X$}{
    Draw $G_{\rm S}$ student rollouts $\{y_{\rm S}^{(g)}\}_{g=1}^{G_{\rm S}}\!\sim\!\pi_\theta(\cdot|x)$.\;
    Forward them through $\pi_{\rm T}$, and accumulate the per-token generalized JSD between $\pi_{\rm T}$ and $\pi_\theta$ at every position of those student rollouts.\;
    \textbf{($^{\dagger}$ only)} Score the group with the binary reward and keep its mean $\bar r_x$.\;
}
Split $X$ into $I$ optimizer mini-batches of $96$ new prompts each, as in the RL rows, and apply one AdamW update per mini-batch on its JSD.\;
\textbf{($^{\dagger}$ only)} Update $\mathcal{B}$ from $\bar r_x$: admit, graduate, and FIFO-evict.\;
\Return{$\pi_\theta$ (and $\mathcal{B}$ for $^{\dagger}$).}
\caption{Executed On-Distill step. The student imitates the teacher's re-scoring of its own rollouts. The $^{\dagger}$ row adds prompt replay.}
\label{alg:onpolicy}
\end{algorithm}

\begin{table}[!htbp]
    \centering
    \caption{Distillation baselines at three learning rates. The $10^{-6}$ rows are the defaults of \cref{tab:main,tab:generalization}, and every other setting stays at its default (\cref{tab:appendix:hyperparameters}); $^{\dagger}$ denotes augmentation with the prompt replay buffer.}
    \label{tab:appendix:distill_lr}
    \resizebox{0.7\linewidth}{!}{
    \setlength{\tabcolsep}{2mm}
    \renewcommand{\arraystretch}{1.05}
    \begin{tabular}{llcccccc}
        \toprule
        \multirow{2}{*}{Method} & \multirow{2}{*}{Learning rate} & \multicolumn{3}{c}{Qwen3.5-0.8B} & \multicolumn{3}{c}{Qwen3.5-2B} \\
        \cmidrule(lr){3-5}\cmidrule(lr){6-8}
                 &     & LLM  & VLM  & Video & LLM  & VLM  & Video \\
        \midrule
        Base                     & --                    & $25.2$ & $41.0$ & $48.3$ & $45.3$ & $56.8$ & $60.6$ \\
        \cdashline{1-8}\noalign{\vskip 0.3ex}
        Off-Distill$^{\dagger}$  & $10^{-6}$ (default)   & $22.5$ & $41.5$ & $45.0$ & $43.1$ & $57.3$ & $58.6$ \\
        Off-Distill$^{\dagger}$  & $10^{-5}$             & $21.8$ & $41.9$ & $44.7$ & $42.0$ & $57.6$ & $57.2$ \\
        Off-Distill$^{\dagger}$  & $5\!\times\!10^{-5}$  & $21.3$ & $41.7$ & $44.2$ & $41.7$ & $57.4$ & $57.1$ \\
        \cdashline{1-8}\noalign{\vskip 0.3ex}
        On-Distill$^{\dagger}$   & $10^{-6}$ (default)   & $23.2$ & $42.3$ & $45.8$ & $43.8$ & $58.0$ & $59.2$ \\
        On-Distill$^{\dagger}$   & $10^{-5}$             & $22.3$ & $42.8$ & $45.5$ & $42.3$ & $58.4$ & $58.5$ \\
        On-Distill$^{\dagger}$   & $5\!\times\!10^{-5}$  & $22.2$ & $42.4$ & $45.2$ & $42.0$ & $58.2$ & $58.3$ \\
        \bottomrule
    \end{tabular}
    }
\end{table}

\paragraph{GRPO and GRPO$^{\dagger}$.}
GRPO runs only Steps~1 and~6 of \cref{alg:zppo}, without the buffer.
GRPO$^{\dagger}$ keeps the buffer at $\rho_{\rm replay}\!=\!0.25$. Both follow the training
pipeline of \cref{sec:experiments}: clip-higher, a token-level loss, no KL penalty,
\emph{Norm w/o Zero}, $I\!=\!4$, and no dynamic sampling. DAPO$^{\dagger}$ adds
DAPO's dynamic sampling~\citep{yu2025dapo} to GRPO$^{\dagger}$, and
\cref{tab:compute_matched} runs it for two epochs.

\paragraph{Iterations per step $I$ and the corresponding budgets.}
$I$ is the number of optimizer mini-batches that one rollout batch is split into. Every
rollout enters exactly one mini-batch, and the PPO inner-epoch count is $1$, so a run has
$200I$ updates. At the default $I\!=\!4$ and one epoch over ZPPO-$77$K, every RL,
distillation, and supervised row takes $800$ mini-batches of $96$ new prompts
(\cref{tab:appendix:hyperparameters}).
DAPO's batch and mini-batch sizes correspond to $I\!=\!16$~\citep{yu2025dapo}, the largest
value in \cref{fig:recipe}a--c. The new-prompt batch is $96I$, so in the sweep
the prompt draws grow with the update count: $19{,}200$ to $307{,}200$ draws at
$I\!\in\!\{1,4,8,16\}$. At $I\!=\!8$ and $I\!=\!16$, training runs for two and four
epochs. GRPO$^{\dagger}$ at $I\!=\!8$ over $200$ rollout steps and GRPO$^{\dagger}$ at
$I\!=\!4$ over $400$ steps (the $\times$2 epochs row of \cref{tab:compute_matched}) both
draw $153{,}600$ new prompts and run $1{,}600$ updates. At $2$B, $I\!=\!4$ over $400$ steps stays above
$I\!=\!8$ on all three families ($47.2$/$59.4$/$61.7$ vs.\ $46.3$/$57.7$/$61.3$ on
LLM/VLM/Video).

\paragraph{Why build the Hint and Prefix controls.}
The cited methods differ from ZPPO in more than the form of guidance: gates, schedules,
reward shaping, importance weights, offline traces, and auxiliary models. Each control keeps
only its family's form of guidance and runs it on the ZPPO pipeline, under the same per-step caps
and update rule, so that its gap to BCQ isolates the form of
guidance. Both controls also keep teacher tokens out of the gradient, so they need no
importance weight, the correction that a teacher token in the gradient would require
(\cref{sec:appendix:algorithm}). The Hint control is therefore on-policy in the same sense as
BCQ, because its teacher text sits in the prompt. The Prefix control is as on-policy as a
prefix allows: its continuations start from a teacher-written opening inside the
student's own response. In \cref{tab:hint_methods,tab:prefix_methods}, BCQ is above both controls in
every column.

\paragraph{Hint control (\cref{tab:hint_methods}).}
Hint methods give an answer-free cue abstracted from a
solution~\citep{nath2025adaptive,chen2026nudging,wang2026don,xia2026learning,wu2025thought}.
Our control runs on GRPO$^{\dagger}$ and combines elements of six methods. The cue is
answer-free (NuRL~\citep{chen2026nudging}, HINT~\citep{wang2026don}) and sits after the
question, in training prompts only (NuRL). A teacher writes it online
(HiLL~\citep{xia2026learning}, SocraticPO~\citep{liu2026socraticpo}) from a correct solution
(HiLL), and it gives high-level guidance rather than the solution
(TAPO~\citep{wu2025thought}). The student samples extra rollouts under the cue on a hard
question (Guide-GRPO~\citep{nath2025adaptive}), and that group trains under the prompt
that sampled it (HiLL, SocraticPO). It leaves out every other part of these methods, such
as the all-wrong gates, Guide-GRPO's pooled advantage and importance weight, NuRL's
reference-answer input and two-stage schedule, HINT's Affinity weighting, HiLL's trained
hinter and hint search, and TAPO's thought library.

Concretely, the frozen $27$B teacher draws $G_{\rm T}\!=\!4$ rollouts on each of at most
the $96$ hardest questions with $\bar r_x\!<\!\tau$, graded as for the BCQ pool. The
teacher then writes the cue from one uniformly chosen graded-correct rollout and, as in
candidate compression, reads only that rollout, never the question or the image. The teacher's
prompt forbids revealing the final answer or writing a \texttt{\textbackslash boxed\{\}} span
(\cref{fig:prompt-hint}). A question with no correct teacher rollout gets no cue. The cue
goes inside \texttt{<hint>}, after the question and before the instruction of
\cref{fig:prompt-format}. The student samples $G_{\rm S}$ rollouts on that prompt as a new
group under the full binary reward. The plain group stays in the step, and each group
trains under its own prompt with the two-step estimator and the clipped surrogate. The
buffer is updated as in ZPPO. Evaluation uses no \texttt{<hint>} block
(\cref{alg:hint}).
\begin{figure}[!htbp]
\begin{prompttemplatefig}
\small
\emph{Read the solved response below and write a concise hint that helps solve the original question. Do NOT reveal the final answer, do NOT include any \texttt{\textbackslash boxed\{\}} expression, and do NOT copy the full solution. Keep only high-level guidance or the key intermediate idea.}

\emph{\texttt{<response>}\\
$y_{\rm T}^{(+)}$\\
\texttt{</response>}}
\end{prompttemplatefig}
\caption{Hint-generation prompt for \cref{alg:hint}. The frozen teacher reads the solved response.}
\label{fig:prompt-hint}
\end{figure}
\begin{algorithm}[!htbp]
\SetAlgoLined
\DontPrintSemicolon
\KwIn{Student $\pi_\theta$, frozen teacher $\pi_{\rm T}$, dataset $\mathcal{D}$, buffer $\mathcal{B}$; $G_{\rm T}\!=\!4$, $G_{\rm S}$, $I$, $\rho_{\rm replay},\rho_{\rm aug},|\mathcal{B}|_{\max},\tau$.}
Sample $X_{\rm new}\!\sim\!\mathcal{D}$, draw $X_{\rm replay}$ from $\mathcal{B}$ by the replay rule, and set $X\!\leftarrow\!\operatorname{concat}(X_{\rm new},X_{\rm replay})$.\;
\ForEach{$x \in X$}{
    Draw $G_{\rm S}$ student rollouts $\{y_{\rm S}^{(g)}\}_{g=1}^{G_{\rm S}}\!\sim\!\pi_\theta(\cdot|x)$, score them, and compute $\bar r_x$.\;
}
Select $X_{\rm aug}$: at most the $\rho_{\rm aug}|X_{\rm new}|$ hardest questions with $\bar r_x\!<\!\tau$, ranked by ascending $\bar r_x$.\;
\ForEach{$x \in X_{\rm aug}$}{
    Draw $G_{\rm T}$ fresh teacher rollouts $\{y_{\rm T}^{(g)}\}_{g=1}^{G_{\rm T}}\!\sim\!\pi_{\rm T}(\cdot|x)$, grade them, and keep the $N_x$ correct ones.\;
    \If{$N_x>0$}{
        Choose one correct rollout $y_{\rm T}^{(+)}$ uniformly and query $\pi_{\rm T}$ with \cref{fig:prompt-hint} on $y_{\rm T}^{(+)}$ to obtain the cue $h$.\;
        Form $x_{\rm Hint}$ by inserting \texttt{<hint>}$\,h\,$\texttt{</hint>} after the question and before the instruction of \cref{fig:prompt-format}.\;
        Draw $G_{\rm S}$ student rollouts $\{y_{\rm Hint}^{(g)}\}_{g=1}^{G_{\rm S}}\!\sim\!\pi_\theta(\cdot|x_{\rm Hint})$ and score them as a new group.\;
    }
}
Compute \cref{eq:adv-step1} within each plain and hint group, and split the groups across $I$ optimizer mini-batches. In each mini-batch, apply \cref{eq:adv-step2} to the non-trivial groups, keep trivial-group advantages at zero, and apply the token-level clipped surrogate of \cref{eq:dualclip}, each group under the prompt that sampled it. Skip a mini-batch whose advantages are all zero.\;
Update $\mathcal{B}$ from the plain-rollout $\bar r_x$: admit, graduate, and FIFO-evict.\;
\Return{$\pi_\theta$, $\mathcal{B}$.}
\caption{Executed Hint-control step. The cue enters training prompts only.}
\label{alg:hint}
\end{algorithm}

\paragraph{Prefix control (\cref{tab:prefix_methods}).}
Prefix methods give the opening part of the solution itself, and the student continues
from it~\citep{zhang2025bread,zhang2025stephint,li2025staying,huang2025blending,huang2025boosting}.
Our control runs on GRPO$^{\dagger}$ and combines elements of six methods. The source is a
correct rollout of a stronger model, used in the same step as the student's plain rollouts
(LUFFY~\citep{NEURIPS2025_a9d5c33e}, StepHint~\citep{zhang2025stephint}). The prefix opens
the response while the question stays unchanged (Hint-GRPO~\citep{huang2025boosting},
StepHint), and it is cut at a fraction of the length (Prefix-RFT~\citep{huang2025blending}),
with $\alpha\!=\!0.4$. Student rollouts branch from it on hard questions
(BREAD~\citep{zhang2025bread}), and no policy-gradient term touches the prefix (BREAD,
SEELE~\citep{li2025staying}). It leaves out every other part of these methods, such as
offline traces, step splits, adaptive or scheduled prefix lengths, prefixes on every
problem, replacement of the plain group, BREAD's all-wrong gate, and any loss on prefix or
teacher tokens (e.g., SEELE's supervised term and LUFFY's reshaped ratio).

Concretely, the frozen $27$B teacher draws $G_{\rm T}\!=\!4$ rollouts on each of at most
the $96$ hardest questions with $\bar r_x\!<\!\tau$. One uniformly chosen graded-correct
rollout $y_{\rm T}^{(+)}\!=\!(t_1,\dots,t_L)$ gives the prefix
$p\!=\!(t_1,\dots,t_{\lfloor\alpha L\rfloor})$. A question with no correct teacher rollout
gets no prefix. The student samples $G_{\rm S}$ continuations from $p$ as a new group, and
the plain group stays in the step. The reward grades prefix and continuation together,
and every prefix token is masked from the loss.
The update and the buffer follow ZPPO. Evaluation uses no prefix (\cref{alg:prefix}).
\begin{algorithm}[!htbp]
\SetAlgoLined
\DontPrintSemicolon
\KwIn{Student $\pi_\theta$, frozen teacher $\pi_{\rm T}$, dataset $\mathcal{D}$, buffer $\mathcal{B}$; $G_{\rm T}\!=\!4$, $G_{\rm S}$, $\alpha\!=\!0.4$, $I$, $\rho_{\rm replay},\rho_{\rm aug},|\mathcal{B}|_{\max},\tau$.}
Sample $X_{\rm new}\!\sim\!\mathcal{D}$, draw $X_{\rm replay}$ from $\mathcal{B}$ by the replay rule, and set $X\!\leftarrow\!\operatorname{concat}(X_{\rm new},X_{\rm replay})$.\;
\ForEach{$x \in X$}{
    Draw $G_{\rm S}$ student rollouts $\{y_{\rm S}^{(g)}\}_{g=1}^{G_{\rm S}}\!\sim\!\pi_\theta(\cdot|x)$, score them, and compute $\bar r_x$.\;
}
Select $X_{\rm aug}$: at most the $\rho_{\rm aug}|X_{\rm new}|$ hardest questions with $\bar r_x\!<\!\tau$, ranked by ascending $\bar r_x$.\;
\ForEach{$x \in X_{\rm aug}$}{
    Draw $G_{\rm T}$ fresh teacher rollouts $\{y_{\rm T}^{(g)}\}_{g=1}^{G_{\rm T}}\!\sim\!\pi_{\rm T}(\cdot|x)$, grade them, and keep the $N_x$ correct ones.\;
    \If{$N_x>0$}{
        Choose one correct rollout $y_{\rm T}^{(+)}=(t_1,\dots,t_L)$ uniformly and set the response prefix $p=(t_1,\dots,t_{\lfloor\alpha L\rfloor})$.\;
        Draw $G_{\rm S}$ student continuations $\{c^{(g)}\}_{g=1}^{G_{\rm S}}\!\sim\!\pi_\theta(\cdot|x,p)$ from the shared prefix.\;
        Score the responses $\{p\,\Vert\,c^{(g)}\}$ as a new group, and mask every token of $p$ from the loss (no policy-gradient or supervised term on $p$).\;
    }
}
Compute \cref{eq:adv-step1} within each plain and prefix group, and split the groups across $I$ optimizer mini-batches. In each mini-batch, apply \cref{eq:adv-step2} to the non-trivial groups, keep trivial-group advantages at zero, and apply the token-level clipped surrogate of \cref{eq:dualclip} to the unmasked response tokens. Skip a mini-batch whose advantages are all zero.\;
Update $\mathcal{B}$ from the plain-rollout $\bar r_x$: admit, graduate, and FIFO-evict.\;
\Return{$\pi_\theta$, $\mathcal{B}$.}
\caption{Executed Prefix-control step. The prefix opens the response, and every prefix token is masked from the loss.}
\label{alg:prefix}
\end{algorithm}

\paragraph{Correction-SFT control (\cref{tab:mistakes}).}
Correction-SFT takes the correction format and the filter of LeMa~\citep{an2023learning}:
a corrector names the wrong step, explains why it is wrong, and rewrites the solution from
there, and only a correction with a correct final answer is kept. It takes the input tags of
NAT~\citep{wang2024learning}: a tag in the input marks the response as correct or
incorrect, and both train under the same cross-entropy. The corrector is the frozen $27$B
teacher. For each selected hard question, it reads the question and one uniformly sampled
wrong rollout (\cref{fig:prompt-correction}). Each graded-correct correction gives two
instances at equal count: the correction under \texttt{<correct>} and the wrong rollout
under \texttt{<incorrect>}. The input is the question, the tag, and the instruction of
\cref{fig:prompt-format}; as in LeMa, the input of the \texttt{<correct>} instance also holds
the wrong rollout that its correction fixes. Cross-entropy covers response tokens only. The teacher writes
the correction, and the refinement target below, in the think-then-boxed format of
\cref{fig:prompt-format}. As in
NAT, evaluation uses the positive tag.
\begin{figure}[!htbp]
\begin{prompttemplatefig}
\small
\emph{Below is a question and a student's incorrect attempt. Identify the first step that is wrong, state briefly why it is wrong, then write a complete corrected solution from that point onward. Do not restate the incorrect attempt. End with the final answer in \texttt{\textbackslash boxed\{\}} format.}

\emph{\texttt{<question>}\\
$x$\\
\texttt{</question>}}

\emph{\texttt{<incorrect\_attempt>}\\
$y_{\rm S}^{(-)}$\\
\texttt{</incorrect\_attempt>}}
\end{prompttemplatefig}
\caption{Correction-generation prompt for \cref{alg:correctionsft}. The frozen teacher reads the question and one wrong student rollout.}
\label{fig:prompt-correction}
\end{figure}
\begin{algorithm}[!htbp]
\SetAlgoLined
\DontPrintSemicolon
\KwIn{Student $\pi_\theta$, frozen teacher $\pi_{\rm T}$, dataset $\mathcal{D}$, buffer $\mathcal{B}$; $G_{\rm S}$, $I$, $\rho_{\rm replay},\rho_{\rm aug},|\mathcal{B}|_{\max},\tau$.}
Sample $X_{\rm new}\!\sim\!\mathcal{D}$, draw $X_{\rm replay}$ from $\mathcal{B}$ by the replay rule, and set $X\!\leftarrow\!\operatorname{concat}(X_{\rm new},X_{\rm replay})$.\;
\ForEach{$x \in X$}{
    Draw $G_{\rm S}$ student rollouts $\{y_{\rm S}^{(g)}\}_{g=1}^{G_{\rm S}}\!\sim\!\pi_\theta(\cdot|x)$, score them, and compute $\bar r_x$.\;
}
Select $X_{\rm aug}$: at most the $\rho_{\rm aug}|X_{\rm new}|$ hardest questions with $\bar r_x\!<\!\tau$, ranked by ascending $\bar r_x$.\;
\ForEach{$x \in X_{\rm aug}$}{
    Choose one wrong rollout $y_{\rm S}^{(-)}$ uniformly, query $\pi_{\rm T}$ with \cref{fig:prompt-correction} to obtain the correction $y_{\rm corr}$, and grade it.\;
    \If{$y_{\rm corr}$ is correct}{
        Form $(x\,\Vert\,y_{\rm S}^{(-)}\,\Vert\,$\texttt{<correct>}$,\,y_{\rm corr})$ and $(x\,\Vert\,$\texttt{<incorrect>}$,\,y_{\rm S}^{(-)})$, with each tag placed before the instruction of \cref{fig:prompt-format}.\;
        Accumulate cross-entropy on both responses at equal instance count, with prompt and tag tokens masked.\;
    }
}
Split $X$ into $I$ optimizer mini-batches of $96$ new prompts each, as in the RL rows. Apply one AdamW update per mini-batch on its cross-entropy, and skip a mini-batch with no verified pair.\;
Update $\mathcal{B}$ from the plain-rollout $\bar r_x$: admit, graduate, and FIFO-evict.\;
\Return{$\pi_\theta$, $\mathcal{B}$.}
\caption{Executed Correction-SFT step. The frozen teacher is the corrector, and its correction is graded before it becomes a target.}
\label{alg:correctionsft}
\end{algorithm}

\paragraph{Error-reflection-SFT control (\cref{tab:mistakes}).}
Error-reflection-SFT takes the refinement and the filter of ORION~\citep{wu2026enhancing}:
a teacher solution is refined with the student's own reasoning errors, only a refinement
with a correct final answer is kept, and the student is fine-tuned on the result. As in
Correction-SFT, the frozen teacher writes this target (in ORION, the student refines); it
refines once per question with all wrong attempts at once, and the refinement holds only the
revised solution. On each selected hard question, the
teacher writes $G_{\rm T}\!=\!4$ drafts without seeing student errors. It then revises one
graded-correct draft in a single call, using all $K_x$ current wrong rollouts
(\cref{fig:prompt-reflection}).
A refinement graded correct becomes the target, with cross-entropy on its response tokens
only (\cref{alg:errorreflectionsft}).
\begin{figure}[!htbp]
\begin{prompttemplatefig}
\small
\emph{Below is a question, your own draft solution, and several incorrect attempts by a weaker student. Revise your draft so that it makes explicit the reasoning the student is missing: expand the steps the attempts get wrong, and keep the rest concise. Output only the revised solution, and end it with the final answer in \texttt{\textbackslash boxed\{\}} format.}

\emph{\texttt{<question>}\\
$x$\\
\texttt{</question>}}

\emph{\texttt{<your\_draft>}\\
$y_{\rm T}^{\rm draft}$\\
\texttt{</your\_draft>}}

\emph{\texttt{<student\_attempts>}\\
$y_{\rm S}^{(-,1)}$, $\dots$, $y_{\rm S}^{(-,K_x)}$\\
\texttt{</student\_attempts>}}
\end{prompttemplatefig}
\caption{Refinement prompt for \cref{alg:errorreflectionsft}. The frozen teacher revises one of its graded-correct drafts using the current wrong student rollouts.}
\label{fig:prompt-reflection}
\end{figure}
\begin{algorithm}[!htbp]
\SetAlgoLined
\DontPrintSemicolon
\KwIn{Student $\pi_\theta$, frozen teacher $\pi_{\rm T}$, dataset $\mathcal{D}$, buffer $\mathcal{B}$; $G_{\rm T}\!=\!4$, $G_{\rm S}$, $I$, $\rho_{\rm replay},\rho_{\rm aug},|\mathcal{B}|_{\max},\tau$.}
Sample $X_{\rm new}\!\sim\!\mathcal{D}$, draw $X_{\rm replay}$ from $\mathcal{B}$ by the replay rule, and set $X\!\leftarrow\!\operatorname{concat}(X_{\rm new},X_{\rm replay})$.\;
\ForEach{$x \in X$}{
    Draw $G_{\rm S}$ student rollouts $\{y_{\rm S}^{(g)}\}_{g=1}^{G_{\rm S}}\!\sim\!\pi_\theta(\cdot|x)$, score them, and compute $\bar r_x$.\;
}
Select $X_{\rm aug}$: at most the $\rho_{\rm aug}|X_{\rm new}|$ hardest questions with $\bar r_x\!<\!\tau$, ranked by ascending $\bar r_x$.\;
\ForEach{$x \in X_{\rm aug}$}{
    Draw $G_{\rm T}$ fresh teacher drafts $\{y_{\rm T}^{(g)}\}_{g=1}^{G_{\rm T}}\!\sim\!\pi_{\rm T}(\cdot|x)$, grade them, and keep the $N_x$ correct ones.\;
    Collect the $K_x$ wrong student rollouts $\{y_{\rm S}^{(-,m)}\}_{m=1}^{K_x}$ from $\{y_{\rm S}^{(g)}\}$.\;
    \If{$N_x>0$}{
        Choose one correct draft $y_{\rm T}^{\rm draft}$ uniformly and query $\pi_{\rm T}$ with \cref{fig:prompt-reflection} on $x$, $y_{\rm T}^{\rm draft}$, and all $K_x$ wrong rollouts to obtain the refinement $y_{\rm T}^{\rm ref}$.\;
        Grade $y_{\rm T}^{\rm ref}$ with the shared outcome reward.\;
        \If{$y_{\rm T}^{\rm ref}$ is correct}{Accumulate cross-entropy on $(x,y_{\rm T}^{\rm ref})$ over the refined response tokens only.\;}
    }
}
Split $X$ into $I$ optimizer mini-batches of $96$ new prompts each, as in the RL rows. Apply one AdamW update per mini-batch on its cross-entropy, and skip a mini-batch with no verified target.\;
Update $\mathcal{B}$ from the plain-rollout $\bar r_x$: admit, graduate, and FIFO-evict.\;
\Return{$\pi_\theta$, $\mathcal{B}$.}
\caption{Executed Error-reflection-SFT step. The frozen teacher is the refiner, and its refinement is graded before it becomes a target.}
\label{alg:errorreflectionsft}
\end{algorithm}

\subsection{Compute cost per run}
\label[appendix]{sec:appendix:hyperparameters:compute}

\Cref{tab:appendix:compute} reports the measured end-to-end wall-clock time per
run on the hardware of \cref{tab:appendix:hyperparameters}. \Cref{sec:appendix:budget}
compares these hours with accuracy.

\begin{table}[!htbp]
    \centering
    \caption{Measured end-to-end wall-clock hours per training run, rounded to the nearest hour. Each run uses its own $64\!\times\!$NVIDIA H100-$80$\,GB allocation. The time covers student generation and optimization, teacher generation and candidate compression where used, teacher re-scoring in the distillation rows, reward judging, synchronization, and data movement.}
    \label{tab:appendix:compute}
    \resizebox{0.5\linewidth}{!}{%
    \renewcommand{\arraystretch}{1.05}
    \begin{tabular}{lcccc}
        \toprule
        Method & $0.8$B & $2$B & $4$B & $9$B \\
        \midrule
        Off-Distill & $85$ & $91$ & $95$ & $104$ \\
        Off-Distill$^{\dagger}$ & $88$ & $96$ & $100$ & $108$ \\
        On-Distill & $75$ & $78$ & $82$ & $88$ \\
        On-Distill$^{\dagger}$ & $78$ & $82$ & $83$ & $90$ \\
        On-Distill$^{\dagger}$, $\times$2 epochs & $157$ & $165$ & $168$ & $183$ \\
        GRPO & $61$ & $69$ & $75$ & $90$ \\
        GRPO$^{\dagger}$ & $61$ & $69$ & $76$ & $93$ \\
        GRPO$^{\dagger}$, $\times$2 rollouts & $119$ & $123$ & $146$ & $173$ \\
        GRPO$^{\dagger}$, $\times$2 epochs & $124$ & $141$ & $153$ & $188$ \\
        GRPO$^{\dagger}$, $\times$2 rollouts \& 2 epochs & $223$ & $259$ & $292$ & $366$ \\
        GRPO$^{\dagger}$, $\times$2 replay buffer batch & $65$ & $72$ & $80$ & $99$ \\
        DAPO$^{\dagger}$, $\times$2 epochs & $96$ & $110$ & $132$ & $167$ \\
        \rowcolor{colorful}ZPPO & $68$ & $79$ & $92$ & $110$ \\
        \bottomrule
    \end{tabular}}
\end{table}

\section{Extended main-paper tables}
\label{sec:appendix:main_tables}

\Cref{tab:main_4b9b}, \cref{tab:generalization_4b9b}, and \cref{tab:ablation_4b9b}
give the $4$B and $9$B blocks of \cref{tab:main}, \cref{tab:generalization}, and
\cref{tab:ablation}, in the same format. At $4$B and $9$B, as at $0.8$B and $2$B,
Off-Distill$^{\dagger}$ and On-Distill$^{\dagger}$ stay close to the base on VLM but fall
below it on Video and on the LLM--Video average. For ZPPO's gain over GRPO$^{\dagger}$,
Off-Distill$^{\dagger}$, and On-Distill$^{\dagger}$, the paired bootstrap interval over
benchmark settings excludes zero at every scale and family
(\cref{sec:appendix:stat_reliability}).

\begin{table}[!htbp]
    \centering
    \caption{Main results on the 16 VLM benchmarks at 4B and 9B; $^{\dagger}$ denotes augmentation with the prompt replay buffer, and $\Delta$ reports ZPPO minus the strongest non-ZPPO row in each column on the same model size. Bold marks the column maximum and is shared by every row that attains it; the colored $\Delta$ row is a difference, not a maximum. 0.8B and 2B results: \cref{tab:main}. Benchmark abbreviations: \cref{sec:appendix:benchmarks}.}
    \label{tab:main_4b9b}
    \resizebox{\linewidth}{!}{
    \setlength{\tabcolsep}{0.9mm}
    \renewcommand{\arraystretch}{1.05}
    \begin{tabular}{l*{17}{c}}
    \toprule
    16 VLM Benchmarks
    & AI2D
    & BabyV
    & CharXiv
    & DynaM
    & EmbSp
    & InfoVQA
    & MVerse
    & MVision
    & MVista
    & MMMU$^\text{Pro}$
    & MM-Vet
    & OCR$^\text{EN}$
    & OCR$^\text{ZH}$
    & VisP
    & VBlind
    & WeMath
    & Avg \\
    \midrule
    Qwen3.5-4B & 88.7 & 16.0 & 78.7 & 59.3 & 82.1 & 88.5 & 83.1 & 64.3 & 83.9 & 64.0 & 78.4 & 50.8 & 28.9 & 47.4 & 66.4 & 89.2 & \cellcolor{gray!10}66.9 \\
    \cdashline{1-18}\noalign{\vskip 0.5ex}
    $+$ Off-Distill & 87.2 & 17.5 & 80.3 & 60.9 & 80.6 & 86.4 & 83.7 & 67.0 & 83.0 & 64.8 & 76.8 & 51.0 & 29.4 & 47.6 & 66.9 & 90.2 & \cellcolor{gray!10}67.1 \\
    $+$ Off-Distill$^{\dagger}$ & 87.0 & 17.7 & 80.0 & 60.7 & 81.1 & 86.8 & 83.7 & 67.5 & 82.9 & 65.1 & 77.7 & 52.0 & 28.3 & 48.8 & 67.4 & 90.6 & \cellcolor{gray!10}67.3 \\
    $+$ On-Distill & 87.1 & 17.5 & 80.7 & 61.5 & 81.0 & 87.2 & 83.9 & 67.8 & 84.0 & 65.6 & 78.2 & 51.3 & 29.5 & 48.2 & 67.8 & 90.6 & \cellcolor{gray!10}67.6 \\
    $+$ On-Distill$^{\dagger}$ & 88.2 & 17.4 & 80.5 & 61.6 & 80.5 & 87.7 & 84.5 & 67.6 & 83.8 & 65.6 & 78.2 & 52.0 & 30.0 & 48.7 & 67.8 & 91.4 & \cellcolor{gray!10}67.8 \\
    \cdashline{1-18}\noalign{\vskip 0.5ex}
    $+$ GRPO & 88.4 & 18.9 & 80.5 & 61.5 & 81.3 & 88.0 & 84.6 & 68.3 & 84.1 & 66.0 & 78.7 & 52.2 & 30.6 & 49.2 & 68.4 & 90.8 & \cellcolor{gray!10}68.2 \\
    $+$ GRPO$^{\dagger}$ & 88.5 & 20.9 & 81.4 & 61.4 & 82.0 & 88.1 & 84.5 & 68.3 & 85.3 & 65.5 & 78.4 & 52.4 & 30.4 & 51.0 & 69.4 & 91.8 & \cellcolor{gray!10}68.7 \\
    \rowcolor{colorful}
    $+$ ZPPO & \textbf{89.6} & \textbf{27.8} & \textbf{82.8} & \textbf{63.9} & \textbf{82.4} & \textbf{88.8} & \textbf{85.7} & \textbf{72.2} & \textbf{85.4} & \textbf{67.1} & \textbf{79.8} & \textbf{54.6} & \textbf{33.0} & \textbf{54.5} & \textbf{74.9} & \textbf{92.6} & \cellcolor{gray!10}\textbf{70.9} \\
    \cdashline{1-18}\noalign{\vskip 0.3ex}
    $\Delta$ & \dpos{0.9} & \dpos{6.9} & \dpos{1.4} & \dpos{2.3} & \dpos{0.3} & \dpos{0.3} & \dpos{1.1} & \dpos{3.9} & \dpos{0.1} & \dpos{1.1} & \dpos{1.1} & \dpos{2.2} & \dpos{2.4} & \dpos{3.5} & \dpos{5.5} & \dpos{0.8} & \cellcolor{gray!10}\dpos{2.2} \\
    \midrule
    Qwen3.5-9B & 90.2 & 22.9 & 82.9 & 62.5 & \textbf{82.1} & 87.8 & 86.9 & 70.8 & 86.3 & 67.7 & 83.0 & 50.0 & 31.7 & 55.3 & 71.7 & 90.9 & \cellcolor{gray!10}70.2 \\
    \cdashline{1-18}\noalign{\vskip 0.5ex}
    $+$ Off-Distill & 89.1 & 23.0 & 82.0 & 64.7 & 80.9 & 88.1 & 85.7 & 73.2 & 85.5 & 67.9 & 81.4 & 51.1 & 32.4 & 54.4 & 73.4 & 91.0 & \cellcolor{gray!10}70.2 \\
    $+$ Off-Distill$^{\dagger}$ & 89.2 & 23.4 & 82.6 & 64.8 & 80.7 & 88.6 & 86.1 & 72.8 & 85.5 & 68.6 & 81.1 & 51.6 & 32.2 & 55.2 & 73.7 & 91.5 & \cellcolor{gray!10}70.5 \\
    $+$ On-Distill & 89.5 & 24.3 & 82.2 & 64.6 & 81.0 & 89.5 & 86.5 & 72.9 & 86.4 & 69.1 & 82.8 & 51.6 & 32.6 & 54.1 & 73.3 & 91.5 & \cellcolor{gray!10}70.7 \\
    $+$ On-Distill$^{\dagger}$ & 89.1 & 24.0 & 83.0 & 64.6 & 80.9 & 89.5 & 86.4 & 73.5 & 86.4 & 69.6 & 82.9 & 52.0 & 32.5 & 55.6 & 73.6 & 91.8 & \cellcolor{gray!10}71.0 \\
    \cdashline{1-18}\noalign{\vskip 0.5ex}
    $+$ GRPO & 89.8 & 24.7 & 83.4 & 65.3 & 80.4 & 89.6 & 87.4 & 74.3 & 86.0 & 69.1 & 83.2 & 52.2 & 32.9 & 55.9 & 73.3 & 91.8 & \cellcolor{gray!10}71.2 \\
    $+$ GRPO$^{\dagger}$ & 90.2 & 24.0 & 83.2 & 65.1 & 81.4 & 89.6 & 87.9 & 75.0 & 87.3 & 68.9 & 83.0 & 52.2 & 32.9 & 55.7 & 74.3 & 92.2 & \cellcolor{gray!10}71.4 \\
    \rowcolor{colorful}
    $+$ ZPPO & \textbf{90.8} & \textbf{26.3} & \textbf{83.7} & \textbf{66.7} & \textbf{82.1} & \textbf{91.4} & \textbf{88.1} & \textbf{75.9} & \textbf{87.5} & \textbf{70.2} & \textbf{83.9} & \textbf{56.7} & \textbf{35.6} & \textbf{56.4} & \textbf{79.4} & \textbf{92.9} & \cellcolor{gray!10}\textbf{73.0} \\
    \cdashline{1-18}\noalign{\vskip 0.3ex}
    $\Delta$ & \dpos{0.6} & \dpos{1.6} & \dpos{0.3} & \dpos{1.4} & \dzero & \dpos{1.8} & \dpos{0.2} & \dpos{0.9} & \dpos{0.2} & \dpos{0.6} & \dpos{0.7} & \dpos{4.5} & \dpos{2.7} & \dpos{0.5} & \dpos{5.1} & \dpos{0.7} & \cellcolor{gray!10}\dpos{1.6} \\
    \bottomrule
    \end{tabular}
    }
\end{table}

\begin{table}[!htbp]
    \centering
    \caption{Evaluation over the 10 LLM and 5 Video benchmarks at 4B and 9B (probing generalization beyond the training data). $\Delta$ is ZPPO minus the strongest non-ZPPO row per column. Bold marks the column maximum, shared by every row attaining it; the colored $\Delta$ row is a difference. 0.8B and 2B: \cref{tab:generalization}. Benchmark abbreviations: \cref{sec:appendix:benchmarks}.}
    \label{tab:generalization_4b9b}
    \resizebox{\linewidth}{!}{
    \setlength{\tabcolsep}{0.9mm}
    \renewcommand{\arraystretch}{1.05}
    \begin{tabular}{lccccccccccccccccc}
    \toprule
    & \multicolumn{11}{c}{10 LLM Benchmarks} & \multicolumn{6}{c}{5 Video Benchmarks} \\
    \cmidrule(lr){2-12}\cmidrule(lr){13-18}
    Method & AIME25 & AIME26 & CEval & GPQA-D & HLE & IMO-AB & MMLU & MMLU-Pro & MMLU-Rd & MultiCh & Avg & MMVU & MVBench & VMME & VMME$^\text{S}$ & VMMMU & Avg \\
    \midrule
    Qwen3.5-4B & 53.3 & 63.3 & 83.6 & 69.7 & 7.8 & 35.5 & 85.3 & 79.4 & 89.5 & 48.9 & \cellcolor{gray!10}61.6 & 69.2 & \textbf{70.2} & 68.1 & 69.9 & \textbf{80.3} & \cellcolor{gray!10}71.5 \\
    \cdashline{1-18}\noalign{\vskip 0.5ex}
    $+$ Off-Distill$^{\dagger}$ & 53.3 & 63.3 & 83.0 & 68.2 & 6.3 & 35.7 & 83.7 & 77.3 & 88.1 & 48.1 & \cellcolor{gray!10}60.7 & 67.9 & 67.4 & 66.4 & 68.8 & 78.2 & \cellcolor{gray!10}69.7 \\
    $+$ On-Distill$^{\dagger}$ & 53.3 & 63.3 & 82.6 & 69.3 & 7.8 & 35.8 & 85.0 & 77.8 & 88.6 & 48.2 & \cellcolor{gray!10}61.2 & 68.7 & 67.8 & 67.0 & 69.3 & 78.9 & \cellcolor{gray!10}70.3 \\
    $+$ GRPO$^{\dagger}$ & 53.3 & 63.3 & 84.6 & 73.5 & 7.1 & 40.6 & 85.9 & 79.2 & 88.9 & 50.7 & \cellcolor{gray!10}62.7 & 69.2 & 69.1 & 68.2 & 68.7 & 79.4 & \cellcolor{gray!10}70.9 \\
    \rowcolor{colorful}
    $+$ ZPPO & \textbf{56.7} & \textbf{66.7} & \textbf{86.1} & \textbf{77.3} & \textbf{8.3} & \textbf{48.0} & \textbf{86.0} & \textbf{79.7} & \textbf{90.1} & \textbf{56.0} & \cellcolor{gray!10}\textbf{65.5} & \textbf{70.6} & 69.6 & \textbf{68.5} & \textbf{70.2} & 80.2 & \cellcolor{gray!10}\textbf{71.8} \\
    \cdashline{1-18}\noalign{\vskip 0.3ex}
    $\Delta$ & \dpos{3.4} & \dpos{3.4} & \dpos{1.5} & \dpos{3.8} & \dpos{0.5} & \dpos{7.4} & \dpos{0.1} & \dpos{0.3} & \dpos{0.6} & \dpos{5.3} & \cellcolor{gray!10}\dpos{2.8} & \dpos{1.4} & \dneg{0.6} & \dpos{0.3} & \dpos{0.3} & \dneg{0.1} & \cellcolor{gray!10}\dpos{0.3} \\
    \midrule
    Qwen3.5-9B & 60.0 & 70.0 & 84.0 & 77.3 & 8.9 & 38.8 & 87.3 & 81.1 & 90.5 & 50.8 & \cellcolor{gray!10}64.9 & 73.8 & 71.4 & \textbf{70.5} & 72.4 & 80.9 & \cellcolor{gray!10}73.8 \\
    \cdashline{1-18}\noalign{\vskip 0.5ex}
    $+$ Off-Distill$^{\dagger}$ & 60.0 & 70.0 & 84.5 & 76.8 & 8.4 & 39.3 & 87.5 & 79.9 & 89.5 & 51.0 & \cellcolor{gray!10}64.7 & 72.5 & 69.6 & 68.5 & 71.5 & 80.8 & \cellcolor{gray!10}72.6 \\
    $+$ On-Distill$^{\dagger}$ & 60.0 & 70.0 & 84.2 & 77.4 & 9.2 & 39.4 & 87.2 & 81.2 & 90.5 & 52.1 & \cellcolor{gray!10}65.1 & 73.0 & 70.4 & 69.6 & 71.6 & 80.7 & \cellcolor{gray!10}73.1 \\
    $+$ GRPO$^{\dagger}$ & 63.3 & 70.0 & 85.3 & 77.8 & 8.8 & 43.6 & 87.4 & 81.3 & 90.9 & 53.0 & \cellcolor{gray!10}66.1 & 73.1 & 71.5 & 69.3 & 71.2 & 82.0 & \cellcolor{gray!10}73.4 \\
    \rowcolor{colorful}
    $+$ ZPPO & \textbf{70.0} & \textbf{73.3} & \textbf{87.2} & \textbf{78.8} & \textbf{9.8} & \textbf{49.8} & \textbf{88.1} & \textbf{82.0} & \textbf{91.7} & \textbf{56.8} & \cellcolor{gray!10}\textbf{68.8} & \textbf{74.2} & \textbf{71.7} & 70.2 & \textbf{72.8} & \textbf{82.1} & \cellcolor{gray!10}\textbf{74.2} \\
    \cdashline{1-18}\noalign{\vskip 0.3ex}
    $\Delta$ & \dpos{6.7} & \dpos{3.3} & \dpos{1.9} & \dpos{1.0} & \dpos{0.6} & \dpos{6.2} & \dpos{0.6} & \dpos{0.7} & \dpos{0.8} & \dpos{3.8} & \cellcolor{gray!10}\dpos{2.7} & \dpos{0.4} & \dpos{0.2} & \dneg{0.3} & \dpos{0.4} & \dpos{0.1} & \cellcolor{gray!10}\dpos{0.4} \\
    \bottomrule
    \end{tabular}
    }
\end{table}

\begin{table}[!htbp]
    \centering
    \caption{Component ablation of ZPPO over the 16 VLM benchmarks at 4B and 9B. Rows add the prompt replay buffer ($^{\dagger}$), BCQ, and NCQ to GRPO, alone or combined. Bold marks the column maximum and is shared by every row that attains it. 0.8B and 2B: \cref{tab:ablation}. Benchmark abbreviations: \cref{sec:appendix:benchmarks}.}
    \label{tab:ablation_4b9b}
    \resizebox{\linewidth}{!}{
    \setlength{\tabcolsep}{0.9mm}
    \renewcommand{\arraystretch}{1.05}
    \begin{tabular}{l*{17}{c}}
    \toprule
    16 VLM Benchmarks
    & AI2D
    & BabyV
    & CharXiv
    & DynaM
    & EmbSp
    & InfoVQA
    & MVerse
    & MVision
    & MVista
    & MMMU$^\text{Pro}$
    & MM-Vet
    & OCR$^\text{EN}$
    & OCR$^\text{ZH}$
    & VisP
    & VBlind
    & WeMath
    & Avg \\
    \midrule
    Qwen3.5-4B & 88.7 & 16.0 & 78.7 & 59.3 & 82.1 & 88.5 & 83.1 & 64.3 & 83.9 & 64.0 & 78.4 & 50.8 & 28.9 & 47.4 & 66.4 & 89.2 & \cellcolor{gray!10}66.9 \\
    \cdashline{1-18}\noalign{\vskip 0.5ex}
    $+$ GRPO & 88.4 & 18.9 & 80.5 & 61.5 & 81.3 & 88.0 & 84.6 & 68.3 & 84.1 & 66.0 & 78.7 & 52.2 & 30.6 & 49.2 & 68.4 & 90.8 & \cellcolor{gray!10}68.2 \\
    $+$ GRPO$^{\dagger}$ & 88.5 & 20.9 & 81.4 & 61.4 & 82.0 & 88.1 & 84.5 & 68.3 & 85.3 & 65.5 & 78.4 & 52.4 & 30.4 & 51.0 & 69.4 & 91.8 & \cellcolor{gray!10}68.7 \\
    $+$ GRPO $+$ Both & 89.0 & 20.7 & 81.3 & 61.0 & 82.1 & 87.8 & 84.2 & 67.0 & 84.7 & 65.9 & 79.0 & 51.2 & 30.6 & 50.5 & 69.5 & 91.3 & \cellcolor{gray!10}68.5 \\
    $+$ GRPO$^{\dagger}$ $+$ BCQ & 89.5 & 24.3 & 81.8 & 61.9 & 81.9 & 88.2 & 85.1 & 69.7 & 85.2 & 65.9 & 79.3 & 53.9 & 31.8 & 52.3 & 72.3 & 91.0 & \cellcolor{gray!10}69.6 \\
    $+$ GRPO$^{\dagger}$ $+$ NCQ & 89.3 & 24.9 & 81.4 & 62.7 & 81.7 & 88.7 & 85.2 & 70.2 & 84.9 & 66.3 & 79.4 & 54.2 & 32.1 & 52.2 & 72.8 & 92.1 & \cellcolor{gray!10}69.9 \\
    \rowcolor{colorful}
    $+$ ZPPO & \textbf{89.6} & \textbf{27.8} & \textbf{82.8} & \textbf{63.9} & \textbf{82.4} & \textbf{88.8} & \textbf{85.7} & \textbf{72.2} & \textbf{85.4} & \textbf{67.1} & \textbf{79.8} & \textbf{54.6} & \textbf{33.0} & \textbf{54.5} & \textbf{74.9} & \textbf{92.6} & \cellcolor{gray!10}\textbf{70.9} \\
    \midrule
    Qwen3.5-9B & 90.2 & 22.9 & 82.9 & 62.5 & \textbf{82.1} & 87.8 & 86.9 & 70.8 & 86.3 & 67.7 & 83.0 & 50.0 & 31.7 & 55.3 & 71.7 & 90.9 & \cellcolor{gray!10}70.2 \\
    \cdashline{1-18}\noalign{\vskip 0.5ex}
    $+$ GRPO & 89.8 & 24.7 & 83.4 & 65.3 & 80.4 & 89.6 & 87.4 & 74.3 & 86.0 & 69.1 & 83.2 & 52.2 & 32.9 & 55.9 & 73.3 & 91.8 & \cellcolor{gray!10}71.2 \\
    $+$ GRPO$^{\dagger}$ & 90.2 & 24.0 & 83.2 & 65.1 & 81.4 & 89.6 & 87.9 & 75.0 & 87.3 & 68.9 & 83.0 & 52.2 & 32.9 & 55.7 & 74.3 & 92.2 & \cellcolor{gray!10}71.4 \\
    $+$ GRPO $+$ Both & 90.3 & 24.3 & 83.3 & 64.5 & 80.9 & 89.8 & 87.7 & 73.5 & 86.0 & 69.3 & 83.6 & 52.7 & 32.5 & 56.0 & 74.5 & 92.2 & \cellcolor{gray!10}71.3 \\
    $+$ GRPO$^{\dagger}$ $+$ BCQ & 90.3 & 25.3 & 83.3 & 65.4 & 82.0 & 90.6 & 87.3 & 74.6 & 86.5 & 70.1 & 83.5 & 55.0 & 34.0 & 56.0 & 76.2 & 92.4 & \cellcolor{gray!10}72.0 \\
    $+$ GRPO$^{\dagger}$ $+$ NCQ & 90.7 & 26.0 & 82.9 & 66.4 & 81.6 & 90.7 & 87.9 & 75.4 & 86.9 & 70.0 & 83.3 & 56.2 & 35.2 & 56.2 & 78.4 & 92.8 & \cellcolor{gray!10}72.5 \\
    \rowcolor{colorful}
    $+$ ZPPO & \textbf{90.8} & \textbf{26.3} & \textbf{83.7} & \textbf{66.7} & \textbf{82.1} & \textbf{91.4} & \textbf{88.1} & \textbf{75.9} & \textbf{87.5} & \textbf{70.2} & \textbf{83.9} & \textbf{56.7} & \textbf{35.6} & \textbf{56.4} & \textbf{79.4} & \textbf{92.9} & \cellcolor{gray!10}\textbf{73.0} \\
    \bottomrule
    \end{tabular}
    }
\end{table}

\paragraph{Setting-level win counts over all four scales.}
We compare ZPPO with GRPO$^{\dagger}$ in each of the
$4\!\times\!31\!=\!124$ (scale, setting) cells of \cref{tab:main},
\cref{tab:generalization}, \cref{tab:main_4b9b}, and \cref{tab:generalization_4b9b},
using the printed one-decimal values. ZPPO is above GRPO$^{\dagger}$ in $121$
cells, tied in $3$, and never below. The ties are AIME25 and AIME26 at $0.8$B and AIME25
at $2$B. At $4$B and $9$B, ZPPO is above GRPO$^{\dagger}$ in all
$31$ settings.

\paragraph{Component ablation on the LLM and Video benchmarks.}
\label{sec:appendix:ablation_gen}
\Cref{tab:ablation_gen} repeats the component ablation of \cref{tab:ablation} on
the $10$ LLM and $5$ Video benchmarks at all four scales.

\begin{table*}[t!]
    \vspace{-3mm}
    \centering
    \caption{Component ablation of ZPPO over the 10 LLM and 5 Video benchmarks at all four scales. Rows add the prompt replay buffer ($^{\dagger}$), BCQ, and NCQ to GRPO, alone or combined (\emph{Both}\,$=$\,BCQ\,$+$\,NCQ), as in \cref{tab:ablation}. Bold marks the column maximum and is shared by every row that attains it. Distillation baselines: \cref{tab:generalization} and \cref{tab:generalization_4b9b}. Benchmark abbreviations: \cref{sec:appendix:benchmarks}.}
    \label{tab:ablation_gen}
    \resizebox{\linewidth}{!}{
    \setlength{\tabcolsep}{0.9mm}
    \renewcommand{\arraystretch}{1.05}
    \begin{tabular}{lccccccccccccccccc}
    \toprule
    & \multicolumn{11}{c}{10 LLM Benchmarks} & \multicolumn{6}{c}{5 Video Benchmarks} \\
    \cmidrule(lr){2-12}\cmidrule(lr){13-18}
    Method & AIME25 & AIME26 & CEval & GPQA-D & HLE & IMO-AB & MMLU & MMLU-Pro & MMLU-Rd & MultiCh & Avg & MMVU & MVBench & VMME & VMME$^\text{S}$ & VMMMU & Avg \\
    \midrule
    Qwen3.5-0.8B & \textbf{3.3} & 0.0 & 52.3 & 12.1 & 2.2 & 3.3 & 57.7 & 41.7 & 62.5 & 16.6 & \cellcolor{gray!10}25.2 & 39.0 & 51.6 & 50.3 & 51.6 & 48.8 & \cellcolor{gray!10}48.3 \\
    \cdashline{1-18}\noalign{\vskip 0.5ex}
    $+$ GRPO & 0.0 & 0.0 & 53.9 & 20.0 & 3.2 & 5.1 & 59.4 & 44.8 & 64.5 & 19.7 & \cellcolor{gray!10}27.1 & 39.9 & 53.2 & 51.4 & 52.7 & 50.8 & \cellcolor{gray!10}49.6 \\
    $+$ GRPO$^{\dagger}$ & \textbf{3.3} & 0.0 & 54.5 & 25.5 & 3.5 & 7.0 & 59.9 & 47.1 & 65.1 & 20.8 & \cellcolor{gray!10}28.7 & 41.3 & 53.5 & 52.2 & 53.0 & 52.5 & \cellcolor{gray!10}50.5 \\
    $+$ GRPO $+$ Both & \textbf{3.3} & 0.0 & 54.8 & 25.2 & 2.6 & 5.8 & 60.4 & 45.9 & 64.2 & 20.7 & \cellcolor{gray!10}28.3 & 40.3 & 53.5 & 51.0 & 51.8 & 51.3 & \cellcolor{gray!10}49.6 \\
    $+$ GRPO$^{\dagger}$ $+$ BCQ & \textbf{3.3} & 0.0 & 56.7 & 35.8 & 4.4 & 8.9 & 62.1 & 50.0 & 67.9 & 25.9 & \cellcolor{gray!10}31.5 & 41.8 & 55.8 & 53.3 & 54.4 & 54.8 & \cellcolor{gray!10}52.0 \\
    $+$ GRPO$^{\dagger}$ $+$ NCQ & \textbf{3.3} & 0.0 & 55.2 & 27.0 & 4.1 & 7.4 & 60.6 & 47.5 & 66.0 & 22.5 & \cellcolor{gray!10}29.4 & 41.6 & 54.3 & 52.5 & 53.6 & 52.9 & \cellcolor{gray!10}51.0 \\
    \rowcolor{colorful}
    $+$ ZPPO & \textbf{3.3} & 0.0 & \textbf{57.6} & \textbf{42.4} & \textbf{5.0} & \textbf{10.5} & \textbf{63.0} & \textbf{51.7} & \textbf{69.0} & \textbf{28.6} & \cellcolor{gray!10}\textbf{33.1} & \textbf{42.5} & \textbf{56.6} & \textbf{53.9} & \textbf{54.9} & \textbf{56.4} & \cellcolor{gray!10}\textbf{52.9} \\
    \midrule
    Qwen3.5-2B & \textbf{26.7} & 30.0 & 71.1 & 53.5 & 3.5 & 11.3 & 74.6 & 66.2 & 80.4 & 35.8 & \cellcolor{gray!10}45.3 & 53.7 & 60.7 & 59.9 & 61.2 & 67.6 & \cellcolor{gray!10}60.6 \\
    \cdashline{1-18}\noalign{\vskip 0.5ex}
    $+$ GRPO & \textbf{26.7} & 33.3 & 72.0 & 55.3 & 3.8 & 15.8 & 74.9 & 66.8 & 81.1 & 37.0 & \cellcolor{gray!10}46.7 & 55.1 & 61.3 & 60.7 & 62.1 & 66.8 & \cellcolor{gray!10}61.2 \\
    $+$ GRPO$^{\dagger}$ & \textbf{26.7} & 33.3 & 72.4 & 56.5 & 4.2 & 19.3 & 75.4 & 67.3 & 81.4 & 37.3 & \cellcolor{gray!10}47.4 & 56.7 & 61.7 & 61.2 & 62.6 & 67.3 & \cellcolor{gray!10}61.9 \\
    $+$ GRPO $+$ Both & \textbf{26.7} & 33.3 & 72.7 & 56.1 & 4.3 & 18.3 & 75.4 & 66.6 & 80.4 & 36.7 & \cellcolor{gray!10}47.1 & 55.7 & 61.0 & 60.3 & 61.6 & 67.3 & \cellcolor{gray!10}61.2 \\
    $+$ GRPO$^{\dagger}$ $+$ BCQ & \textbf{26.7} & 36.7 & 73.4 & 58.7 & 4.9 & 24.6 & 76.0 & 68.0 & 82.3 & 39.3 & \cellcolor{gray!10}49.1 & 57.6 & 62.3 & 61.9 & 63.4 & 67.9 & \cellcolor{gray!10}62.6 \\
    $+$ GRPO$^{\dagger}$ $+$ NCQ & \textbf{26.7} & 36.7 & 73.0 & 57.8 & 4.6 & 22.3 & 75.8 & 67.8 & 81.9 & 38.7 & \cellcolor{gray!10}48.5 & 56.9 & 62.0 & 61.8 & 63.1 & 67.5 & \cellcolor{gray!10}62.3 \\
    \rowcolor{colorful}
    $+$ ZPPO & \textbf{26.7} & \textbf{40.0} & \textbf{74.0} & \textbf{60.6} & \textbf{5.1} & \textbf{29.5} & \textbf{76.4} & \textbf{68.2} & \textbf{82.9} & \textbf{40.6} & \cellcolor{gray!10}\textbf{50.4} & \textbf{59.0} & \textbf{62.7} & \textbf{62.4} & \textbf{63.9} & \textbf{68.1} & \cellcolor{gray!10}\textbf{63.2} \\
    \midrule
    Qwen3.5-4B & 53.3 & 63.3 & 83.6 & 69.7 & 7.8 & 35.5 & 85.3 & 79.4 & 89.5 & 48.9 & \cellcolor{gray!10}61.6 & 69.2 & \textbf{70.2} & 68.1 & 69.9 & \textbf{80.3} & \cellcolor{gray!10}71.5 \\
    \cdashline{1-18}\noalign{\vskip 0.5ex}
    $+$ GRPO & 53.3 & 63.3 & 84.1 & 70.9 & 7.5 & 38.9 & 85.3 & 78.1 & 88.6 & 50.5 & \cellcolor{gray!10}62.1 & 69.1 & 68.3 & 66.8 & 69.2 & 79.6 & \cellcolor{gray!10}70.6 \\
    $+$ GRPO$^{\dagger}$ & 53.3 & 63.3 & 84.6 & 73.5 & 7.1 & 40.6 & 85.9 & 79.2 & 88.9 & 50.7 & \cellcolor{gray!10}62.7 & 69.2 & 69.1 & 68.2 & 68.7 & 79.4 & \cellcolor{gray!10}70.9 \\
    $+$ GRPO $+$ Both & 53.3 & 63.3 & 84.2 & 72.7 & 8.2 & 39.6 & 85.5 & 79.1 & 89.7 & 51.5 & \cellcolor{gray!10}62.7 & 70.2 & 68.1 & 67.7 & 69.2 & 79.5 & \cellcolor{gray!10}70.9 \\
    $+$ GRPO$^{\dagger}$ $+$ BCQ & 53.3 & \textbf{66.7} & 85.2 & 74.9 & 8.2 & 44.4 & 85.4 & 78.9 & 89.4 & 54.3 & \cellcolor{gray!10}64.1 & 70.3 & 69.0 & 68.3 & 69.3 & 79.4 & \cellcolor{gray!10}71.3 \\
    $+$ GRPO$^{\dagger}$ $+$ NCQ & \textbf{56.7} & \textbf{66.7} & 85.5 & 75.4 & 8.2 & 44.5 & 85.6 & 79.5 & 89.7 & 54.2 & \cellcolor{gray!10}64.6 & 70.2 & 69.2 & 68.4 & 70.0 & 80.0 & \cellcolor{gray!10}71.6 \\
    \rowcolor{colorful}
    $+$ ZPPO & \textbf{56.7} & \textbf{66.7} & \textbf{86.1} & \textbf{77.3} & \textbf{8.3} & \textbf{48.0} & \textbf{86.0} & \textbf{79.7} & \textbf{90.1} & \textbf{56.0} & \cellcolor{gray!10}\textbf{65.5} & \textbf{70.6} & 69.6 & \textbf{68.5} & \textbf{70.2} & 80.2 & \cellcolor{gray!10}\textbf{71.8} \\
    \midrule
    Qwen3.5-9B & 60.0 & 70.0 & 84.0 & 77.3 & 8.9 & 38.8 & 87.3 & 81.1 & 90.5 & 50.8 & \cellcolor{gray!10}64.9 & 73.8 & 71.4 & \textbf{70.5} & 72.4 & 80.9 & \cellcolor{gray!10}73.8 \\
    \cdashline{1-18}\noalign{\vskip 0.5ex}
    $+$ GRPO & 63.3 & 70.0 & 84.3 & 77.4 & 8.9 & 41.5 & 87.5 & 80.6 & 90.9 & 52.2 & \cellcolor{gray!10}65.7 & 73.5 & 70.9 & 69.2 & 71.6 & 80.9 & \cellcolor{gray!10}73.2 \\
    $+$ GRPO$^{\dagger}$ & 63.3 & 70.0 & 85.3 & 77.8 & 8.8 & 43.6 & 87.4 & 81.3 & 90.9 & 53.0 & \cellcolor{gray!10}66.1 & 73.1 & 71.5 & 69.3 & 71.2 & 82.0 & \cellcolor{gray!10}73.4 \\
    $+$ GRPO $+$ Both & 63.3 & 70.0 & 85.3 & 77.9 & 9.2 & 42.6 & 87.6 & 81.0 & 90.8 & 52.9 & \cellcolor{gray!10}66.1 & 73.1 & 71.1 & 69.5 & 72.1 & 81.4 & \cellcolor{gray!10}73.4 \\
    $+$ GRPO$^{\dagger}$ $+$ BCQ & 66.7 & \textbf{73.3} & 86.7 & 77.8 & 9.0 & 45.8 & 87.9 & 81.8 & 91.0 & 54.5 & \cellcolor{gray!10}67.5 & 73.8 & 71.5 & 68.8 & 72.6 & 81.5 & \cellcolor{gray!10}73.6 \\
    $+$ GRPO$^{\dagger}$ $+$ NCQ & \textbf{70.0} & \textbf{73.3} & 86.9 & 78.1 & 9.5 & 48.3 & 87.8 & 81.7 & 91.6 & 55.6 & \cellcolor{gray!10}68.3 & 73.9 & 71.6 & 69.7 & 72.5 & 81.7 & \cellcolor{gray!10}73.9 \\
    \rowcolor{colorful}
    $+$ ZPPO & \textbf{70.0} & \textbf{73.3} & \textbf{87.2} & \textbf{78.8} & \textbf{9.8} & \textbf{49.8} & \textbf{88.1} & \textbf{82.0} & \textbf{91.7} & \textbf{56.8} & \cellcolor{gray!10}\textbf{68.8} & \textbf{74.2} & \textbf{71.7} & 70.2 & \textbf{72.8} & \textbf{82.1} & \cellcolor{gray!10}\textbf{74.2} \\
    \bottomrule
    \end{tabular}
    }
\end{table*}

\textbf{(i)~Neither component suffices alone.} Over the $12$ (scale, family) blocks of
\cref{tab:ablation}, \cref{tab:ablation_4b9b}, and \cref{tab:ablation_gen}, BCQ$+$NCQ
without the buffer (GRPO\,$+$\,\emph{Both}) does no better than the buffer alone
(GRPO$^{\dagger}$) in the printed averages. Without the buffer, a hard question is reformulated
only in the step it is drawn (\cref{sec:appendix:baselines}).

\textbf{(ii)~Replay and reformulation are complementary.} Pairing BCQ or NCQ with the
buffer lifts the average above GRPO$^{\dagger}$ in all $24$ cases (four scales, three
families, two reformulations).

\textbf{(iii)~BCQ and NCQ carry the gain at different scales.}
ZPPO\,$-$\,GRPO$^{\dagger}\!+\!$NCQ measures what BCQ adds, and
ZPPO\,$-$\,GRPO$^{\dagger}\!+\!$BCQ what NCQ adds. The two differences stay positive in all $12$ blocks, and
their bootstrap intervals over benchmark settings exclude zero (\cref{tab:appendix:ci_ablation}).
GRPO$^{\dagger}\!+\!$BCQ leads GRPO$^{\dagger}\!+\!$NCQ on all
three families at $0.8$B and $2$B, and GRPO$^{\dagger}\!+\!$NCQ leads at $4$B and $9$B. Full ZPPO has the highest average in every block,
so dropping any single component (BCQ, NCQ, or the buffer) lowers every family's average at
every scale.

\section{Additional empirical results}
\label{sec:appendix:additional_results}

\subsection{Buffer dynamics across student scales}
\label[appendix]{sec:appendix:zone}

\begin{figure}[!htbp]
    \centering
    \includegraphics[width=\linewidth]{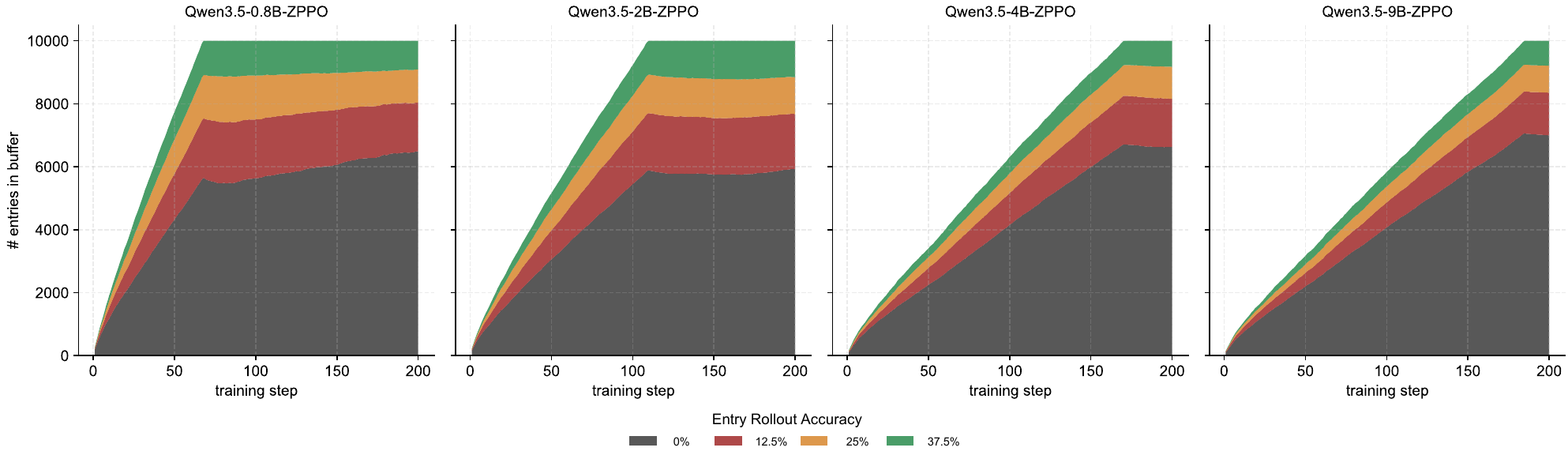}
    \caption{Buffer entries over training for ZPPO at every student scale, split by plain-rollout accuracy at admission. Each panel stacks the four admission accuracies, so the total height is the number of buffer entries.}
    \label{fig:appendix:buffer_compose}
\end{figure}

We track how the prompt replay buffer fills and how its entries leave, by graduation or
FIFO eviction, for ZPPO at each scale. Admission and graduation use only the plain rollouts (\cref{sec:method:buffer}).
\Cref{fig:appendix:buffer_compose} follows the buffer entries at each step.

\textbf{(i)~The buffer reaches capacity earlier and evicts more at smaller scales.} The
$0.8$B run reaches the $10{,}000$-entry capacity first, then $2$B, $4$B, and $9$B
(\cref{fig:appendix:buffer_compose}). In total, the
$0.8$B/$2$B/$4$B/$9$B runs graduate $2{,}334$/$2{,}627$/$2{,}645$/$2{,}455$ entries and
FIFO-evict $17{,}963$/$7{,}251$/$1{,}910$/$896$ (\cref{fig:system}b), so eviction
dominates at the small scales and graduation at the large ones.

\begin{figure}[!htbp]
    \centering
    \includegraphics[width=\linewidth]{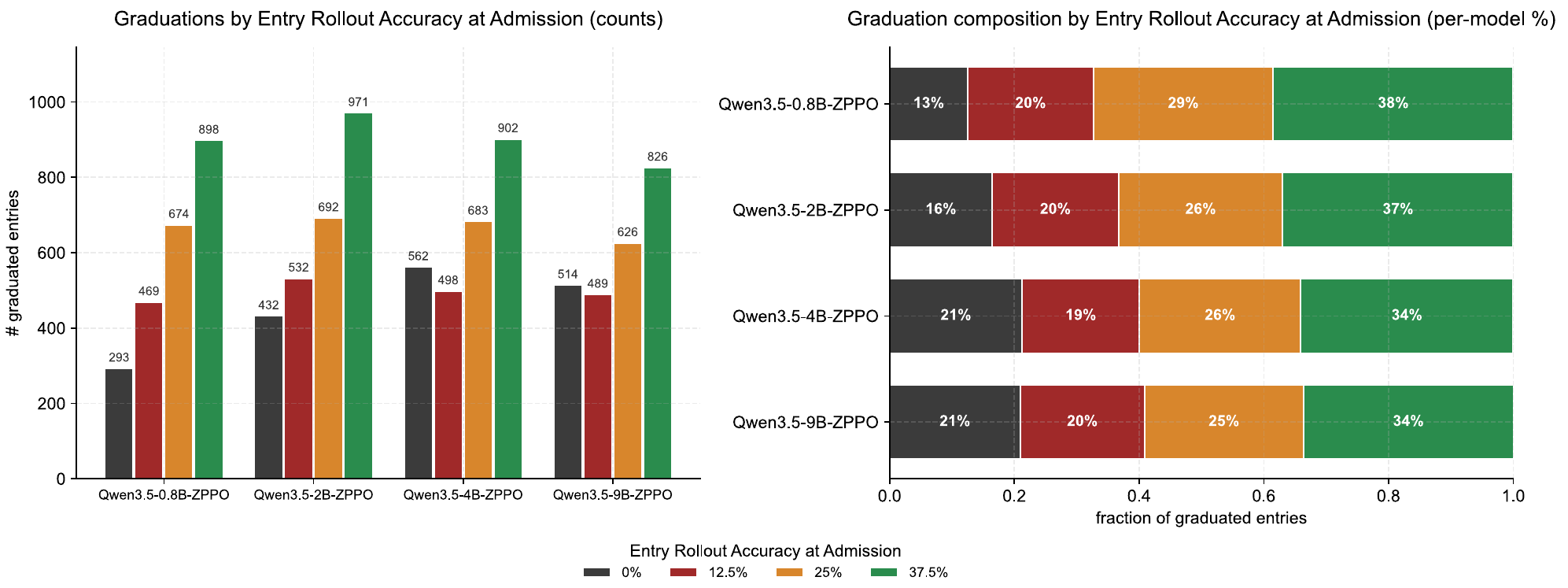}
    \caption{\textbf{(Left)}~Graduate counts per scale, split by plain-rollout accuracy at admission ($0\%$, $12.5\%$, $25\%$, $37.5\%$). \textbf{(Right)}~Share of each admission accuracy among the graduates, per scale.}
    \label{fig:appendix:grad_ratio}
\end{figure}

\textbf{(ii)~At every scale, ZPPO graduates questions admitted at $0\%$.}
\Cref{fig:appendix:grad_ratio} counts the graduates admitted at $0\%$:
$293$/$432$/$562$/$514$ at $0.8$B/$2$B/$4$B/$9$B.

\subsection{ZPPO vs.\ GRPO$^{\dagger}$ training dynamics at $2$B}
\label[appendix]{sec:appendix:dynamics_2b}

We compare ZPPO with GRPO$^{\dagger}$ at $2$B. The two share
every setting of \cref{tab:appendix:hyperparameters} and differ only in whether BCQ
and NCQ are on. Each run admits the questions on which its own plain rollouts give
$\bar{r}_x\!<\!\tau$.

\begin{figure}[!htbp]
    \centering
    \includegraphics[width=\linewidth]{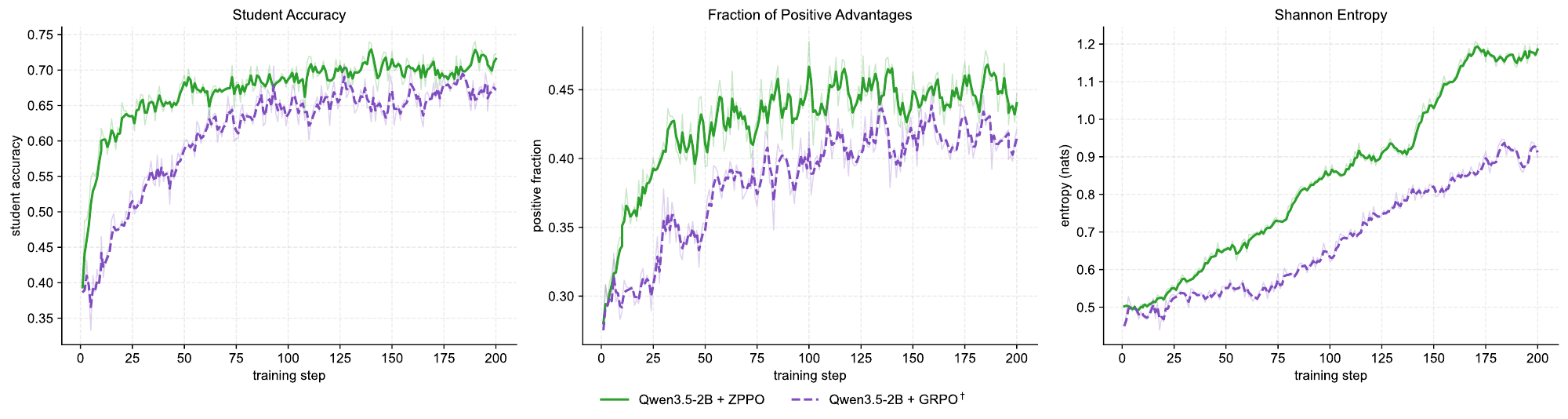}
    \caption{Training dynamics of ZPPO vs.\ GRPO$^{\dagger}$ at $2$B. \textbf{(Left)}~Plain-rollout accuracy. \textbf{(Center)}~Fraction of student rollouts with positive group-relative advantage. \textbf{(Right)}~Actor entropy.}
    \label{fig:appendix:2b_comp}
\end{figure}

\Cref{fig:appendix:2b_comp} plots per-step metrics of the two runs. The positive-advantage
fraction counts every rollout of the step, including ZPPO's BCQ and NCQ rollouts.

\begin{figure}[!htbp]
    \centering
    \includegraphics[width=0.6\linewidth]{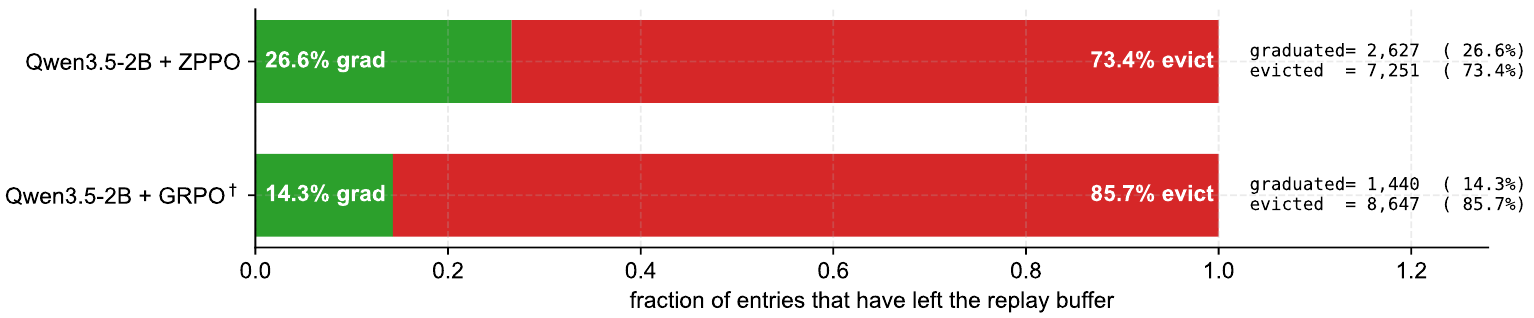}
    \caption{Buffer entries that graduated or were FIFO-evicted, for ZPPO and GRPO$^{\dagger}$ at $2$B, as in \cref{fig:system}b.}
    \label{fig:appendix:grad_evict}
\end{figure}

\textbf{(i)~ZPPO's graduation rate is higher under the same buffer settings.} By step
$200$, ZPPO has graduated $2{,}627$ entries and FIFO-evicted $7{,}251$
($26.6$/$73.4\%$), against $1{,}440$ and $8{,}647$ ($14.3$/$85.7\%$) for GRPO$^{\dagger}$
(\cref{fig:appendix:grad_evict}). \Cref{fig:zone} instead divides the graduates by the
entries admitted at each accuracy and re-sampled at least once: $432$/$1566$ vs.\
$73$/$2035$ at $0\%$ and $532$/$981$ vs.\ $160$/$1158$ at $12.5\%$, which give the $28\%$
vs.\ $4\%$ and $54\%$ vs.\ $14\%$ of \cref{sec:experiments:zone}(i). Graduation is scored
on plain rollouts, with no candidate in the prompt (\cref{sec:experiments:discussion}(iv)).

\begin{figure}[!htbp]
    \centering
    \includegraphics[width=\linewidth]{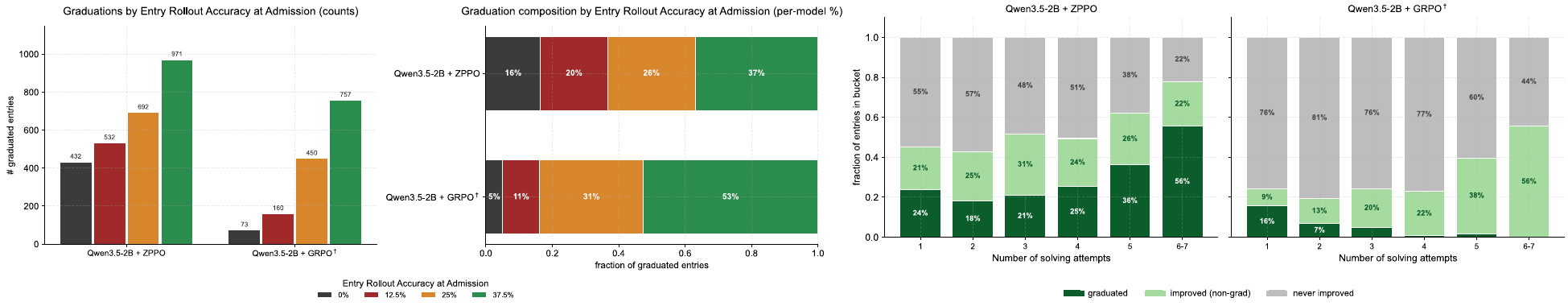}
    \caption{ZPPO vs.\ GRPO$^{\dagger}$ at $2$B. \textbf{(Left two panels)}~Graduate counts by plain-rollout accuracy at admission, and the share of each admission accuracy among each method's graduates. \textbf{(Right two panels)}~For each number of solving attempts, the fraction of entries that graduated, improved without graduating, or never improved.}
    \label{fig:appendix:grad_num_attempt}
\end{figure}

\textbf{(ii)~The largest gaps in graduate count are at the two hardest admission
accuracies.} ZPPO graduates $359$/$372$/$242$/$214$ more entries than GRPO$^{\dagger}$ at
$0\%$/$12.5\%$/$25\%$/$37.5\%$ (\cref{fig:appendix:grad_num_attempt}).

\textbf{(iii)~ZPPO's graduated share is higher at every number of solving attempts} (right two
panels of \cref{fig:appendix:grad_num_attempt}).

\subsection{Benchmark-selection robustness via paired bootstrap}
\label[appendix]{sec:appendix:stat_reliability}

Inside each (scale, family) block, we resample the reported settings $10{,}000$ times with
replacement and recompute the paired mean difference from the printed cells.
\Cref{tab:appendix:ci_main} and \cref{tab:appendix:ci_ablation} report this mean $\Delta$
and its $95\%$ interval. The mean $\Delta$ can differ slightly from the difference of two
rounded \emph{Avg} cells. Each setting is one resampling unit, so these intervals test the
choice of benchmarks.

\textbf{(i)~Every interval excludes zero.} This holds for ZPPO against GRPO$^{\dagger}$,
Off-Distill$^{\dagger}$, On-Distill$^{\dagger}$, GRPO$^{\dagger}\!+\!$BCQ, and
GRPO$^{\dagger}\!+\!$NCQ in all $12$ (scale, family) blocks. Against GRPO$^{\dagger}$, ZPPO
is higher on every VLM and Video setting, and every LLM setting where it is not higher is
an exact tie on AIME25 or AIME26 (\cref{sec:appendix:main_tables}).
\textbf{(ii)~The gain does not depend on one setting.} The ZPPO\,$-$\,GRPO$^{\dagger}$
mean stays above zero in every block when the setting with the largest gap is dropped
and, on LLM, when both AIME settings are dropped.

\begin{table}[!htbp]
    \centering
    \caption{Paired bootstrap $95\%$ intervals over benchmark settings for the
    mean difference $\Delta$ (pp) of \cref{tab:main},
    \cref{tab:generalization}, \cref{tab:main_4b9b}, and
    \cref{tab:generalization_4b9b} (\cref{sec:appendix:stat_reliability}).
    \emph{Win} counts the settings on which ZPPO is higher.}
    \label{tab:appendix:ci_main}
    \resizebox{\linewidth}{!}{%
    \renewcommand{\arraystretch}{1.10}
    \begin{tabular}{llccc}
        \toprule
        Scale & Family & ZPPO $-$ Off-Distill$^{\dagger}$ & ZPPO $-$ On-Distill$^{\dagger}$ & ZPPO $-$ GRPO$^{\dagger}$ \\
         &  & $\Delta$ [95\% CI] (Win) & $\Delta$ [95\% CI] (Win) & $\Delta$ [95\% CI] (Win) \\
        \midrule
        0.8B & VLM & +8.78 [+6.87, +10.61] (16/16) & +7.93 [+6.18, +9.63] (16/16) & +4.79 [+3.45, +6.12] (16/16) \\
        0.8B & LLM & +10.56 [+5.87, +16.47] (9/10) & +9.88 [+5.30, +15.87] (9/10) & +4.44 [+2.02, +7.77] (8/10) \\
        0.8B & Video & +7.86 [+6.80, +9.44] (5/5) & +7.04 [+5.62, +8.96] (5/5) & +2.36 [+1.54, +3.30] (5/5) \\
        \cdashline{1-5}\noalign{\vskip 0.4ex}
        2B & VLM & +4.72 [+3.69, +5.83] (16/16) & +4.02 [+3.03, +5.09] (16/16) & +2.76 [+1.89, +3.74] (16/16) \\
        2B & LLM & +7.27 [+4.40, +10.85] (9/10) & +6.63 [+4.24, +9.82] (10/10) & +3.02 [+1.36, +5.19] (9/10) \\
        2B & Video & +4.62 [+3.34, +6.18] (5/5) & +4.06 [+2.92, +5.42] (5/5) & +1.32 [+0.96, +1.84] (5/5) \\
        \cdashline{1-5}\noalign{\vskip 0.4ex}
        4B & VLM & +3.61 [+2.60, +4.84] (16/16) & +3.10 [+2.00, +4.45] (16/16) & +2.24 [+1.41, +3.19] (16/16) \\
        4B & LLM & +4.79 [+2.86, +7.11] (10/10) & +4.32 [+2.28, +6.74] (10/10) & +2.78 [+1.52, +4.23] (10/10) \\
        4B & Video & +2.08 [+1.70, +2.44] (5/5) & +1.48 [+1.16, +1.78] (5/5) & +0.90 [+0.48, +1.32] (5/5) \\
        \cdashline{1-5}\noalign{\vskip 0.4ex}
        9B & VLM & +2.50 [+1.90, +3.18] (16/16) & +2.01 [+1.39, +2.77] (16/16) & +1.54 [+0.92, +2.29] (16/16) \\
        9B & LLM & +4.06 [+2.16, +6.25] (10/10) & +3.63 [+1.65, +5.91] (10/10) & +2.61 [+1.36, +3.99] (10/10) \\
        9B & Video & +1.62 [+1.38, +1.86] (5/5) & +1.14 [+0.86, +1.32] (5/5) & +0.78 [+0.30, +1.26] (5/5) \\
        \bottomrule
    \end{tabular}%
    }
\end{table}

\begin{table}[!htbp]
    \centering
    \caption{Paired bootstrap $95\%$ intervals over benchmark settings for the
    mean difference $\Delta$ (pp) of \cref{tab:ablation},
    \cref{tab:ablation_4b9b}, and \cref{tab:ablation_gen}, formatted as in
    \cref{tab:appendix:ci_main}. ZPPO\,$-$\,GRPO$^{\dagger}\!+\!$NCQ measures what BCQ
    adds, and ZPPO\,$-$\,GRPO$^{\dagger}\!+\!$BCQ what NCQ adds.}
    \label{tab:appendix:ci_ablation}
    \resizebox{\linewidth}{!}{%
    \renewcommand{\arraystretch}{1.10}
    \begin{tabular}{llccc}
        \toprule
        Scale & Family & ZPPO $-$ GRPO$^{\dagger}$ & ZPPO $-$ GRPO$^{\dagger}\!+\!$BCQ & ZPPO $-$ GRPO$^{\dagger}\!+\!$NCQ \\
         &  & $\Delta$ [95\% CI] (Win) & $\Delta$ [95\% CI] (Win) & $\Delta$ [95\% CI] (Win) \\
        \midrule
        0.8B & VLM & +4.79 [+3.45, +6.12] (16/16) & +1.64 [+1.13, +2.16] (16/16) & +4.05 [+2.84, +5.22] (16/16) \\
        0.8B & LLM & +4.44 [+2.02, +7.77] (8/10) & +1.61 [+0.69, +2.92] (8/10) & +3.75 [+1.60, +6.80] (8/10) \\
        0.8B & Video & +2.36 [+1.54, +3.30] (5/5) & +0.84 [+0.58, +1.22] (5/5) & +1.88 [+1.16, +2.78] (5/5) \\
        \cdashline{1-5}\noalign{\vskip 0.4ex}
        2B & VLM & +2.76 [+1.89, +3.74] (16/16) & +1.21 [+0.79, +1.67] (16/16) & +1.86 [+1.26, +2.52] (16/16) \\
        2B & LLM & +3.02 [+1.36, +5.19] (9/10) & +1.34 [+0.52, +2.42] (9/10) & +1.87 [+0.81, +3.35] (9/10) \\
        2B & Video & +1.32 [+0.96, +1.84] (5/5) & +0.60 [+0.32, +1.02] (5/5) & +0.96 [+0.62, +1.54] (5/5) \\
        \cdashline{1-5}\noalign{\vskip 0.4ex}
        4B & VLM & +2.24 [+1.41, +3.19] (16/16) & +1.31 [+0.86, +1.80] (16/16) & +1.06 [+0.69, +1.47] (16/16) \\
        4B & LLM & +2.78 [+1.52, +4.23] (10/10) & +1.42 [+0.71, +2.22] (9/10) & +0.89 [+0.29, +1.63] (8/10) \\
        4B & Video & +0.90 [+0.48, +1.32] (5/5) & +0.56 [+0.32, +0.80] (5/5) & +0.26 [+0.16, +0.36] (5/5) \\
        \cdashline{1-5}\noalign{\vskip 0.4ex}
        9B & VLM & +1.54 [+0.92, +2.29] (16/16) & +0.94 [+0.62, +1.35] (16/16) & +0.44 [+0.31, +0.56] (16/16) \\
        9B & LLM & +2.61 [+1.36, +3.99] (10/10) & +1.30 [+0.54, +2.16] (9/10) & +0.47 [+0.20, +0.79] (8/10) \\
        9B & Video & +0.78 [+0.30, +1.26] (5/5) & +0.56 [+0.24, +1.00] (5/5) & +0.32 [+0.20, +0.44] (5/5) \\
        \bottomrule
    \end{tabular}%
    }
\end{table}

\section{Teacher capability on the same benchmarks}
\label{sec:appendix:teacher_capability}

We score the frozen $27$B teacher (Qwen3.5-$27$B-FP8) on the $31$ benchmarks of the main tables, with the same decoding and scoring as the students (\cref{sec:appendix:benchmarks}).

\Cref{tab:appendix:teacher_capability:llm}, \cref{tab:appendix:teacher_capability:vlm}, and \cref{tab:appendix:teacher_capability:video} give the per-benchmark scores of the four ZPPO students (from \cref{tab:main,tab:generalization,tab:main_4b9b,tab:generalization_4b9b}) and of the $27$B teacher.

\begin{table}[!htbp]
    \centering
    \caption{Per-benchmark scores on the $10$ LLM benchmarks: ZPPO students at four scales and the $27$B teacher. Bold marks the column maximum and is shared by every row that attains it. Benchmark abbreviations: \cref{sec:appendix:benchmarks}.}
    \label{tab:appendix:teacher_capability:llm}
    \resizebox{0.9\linewidth}{!}{
    \setlength{\tabcolsep}{2mm}
    \renewcommand{\arraystretch}{1.1}
    \begin{tabular}{lccccccccccc}
        \toprule
        Model & AIME25 & AIME26 & CEval & GPQA-D & HLE & IMO-AB & MMLU & MMLU-Pro & MMLU-Rd & MultiCh & Avg \\
        \midrule
        Qwen3.5-0.8B $+\,$ZPPO & $3.3$ & $0.0$ & $57.6$ & $42.4$ & $5.0$ & $10.5$ & $63.0$ & $51.7$ & $69.0$ & $28.6$ & $33.1$ \\
        Qwen3.5-2B   $+\,$ZPPO & $26.7$ & $40.0$ & $74.0$ & $60.6$ & $5.1$ & $29.5$ & $76.4$ & $68.2$ & $82.9$ & $40.6$ & $50.4$ \\
        Qwen3.5-4B   $+\,$ZPPO & $56.7$ & $66.7$ & $86.1$ & $77.3$ & $8.3$ & $48.0$ & $86.0$ & $79.7$ & $90.1$ & $56.0$ & $65.5$ \\
        Qwen3.5-9B   $+\,$ZPPO & $\mathbf{70.0}$ & $73.3$ & $87.2$ & $78.8$ & $9.8$ & $49.8$ & $88.1$ & $82.0$ & $91.7$ & $56.8$ & $68.8$ \\
        \midrule
        Qwen3.5-27B-FP8 & $\mathbf{70.0}$ & $\mathbf{76.7}$ & $\mathbf{91.2}$ & $\mathbf{83.3}$ & $\mathbf{16.0}$ & $\mathbf{51.3}$ & $\mathbf{91.6}$ & $\mathbf{86.3}$ & $\mathbf{94.3}$ & $\mathbf{57.5}$ & $\mathbf{71.8}$ \\
        \bottomrule
    \end{tabular}}
\end{table}

\begin{table}[!htbp]
    \centering
    \caption{Per-benchmark scores on the $16$ VLM benchmarks: ZPPO students at four scales and the $27$B teacher. Bold marks the column maximum and is shared by every row that attains it. Benchmark abbreviations: \cref{sec:appendix:benchmarks}.}
    \label{tab:appendix:teacher_capability:vlm}
    \resizebox{\linewidth}{!}{
    \setlength{\tabcolsep}{1.5mm}
    \renewcommand{\arraystretch}{1.1}
    \begin{tabular}{lccccccccccccccccc}
        \toprule
        Model & AI2D & BabyV & CharXiv & DynaM & EmbSp & InfoVQA & MVerse & MVision & MVista & MMMU$^\text{Pro}$ & MM-Vet & OCR$^\text{EN}$ & OCR$^\text{ZH}$ & VisP & VBlind & WeMath & Avg \\
        \midrule
        Qwen3.5-0.8B $+\,$ZPPO & $76.5$ & $13.9$ & $63.9$ & $31.1$ & $71.5$ & $75.3$ & $59.3$ & $29.2$ & $73.2$ & $37.6$ & $59.9$ & $42.5$ & $18.7$ & $35.0$ & $44.7$ & $71.7$ & $50.3$ \\
        Qwen3.5-2B   $+\,$ZPPO & $85.3$ & $18.6$ & $73.9$ & $52.7$ & $79.5$ & $84.6$ & $76.0$ & $50.5$ & $80.5$ & $53.2$ & $77.1$ & $48.8$ & $26.0$ & $42.0$ & $60.5$ & $82.6$ & $62.0$ \\
        Qwen3.5-4B   $+\,$ZPPO & $89.6$ & $27.8$ & $82.8$ & $63.9$ & $82.4$ & $88.8$ & $85.7$ & $72.2$ & $85.4$ & $67.1$ & $79.8$ & $54.6$ & $33.0$ & $54.5$ & $74.9$ & $92.6$ & $70.9$ \\
        Qwen3.5-9B   $+\,$ZPPO & $90.8$ & $26.3$ & $83.7$ & $66.7$ & $82.1$ & $91.4$ & $88.1$ & $75.9$ & $87.5$ & $70.2$ & $83.9$ & $\mathbf{56.7}$ & $\mathbf{35.6}$ & $56.4$ & $79.4$ & $92.9$ & $73.0$ \\
        \midrule
        Qwen3.5-27B-FP8 & $\mathbf{92.1}$ & $\mathbf{30.4}$ & $\mathbf{86.0}$ & $\mathbf{70.5}$ & $\mathbf{84.2}$ & $\mathbf{91.5}$ & $\mathbf{89.1}$ & $\mathbf{80.0}$ & $\mathbf{88.2}$ & $\mathbf{74.0}$ & $\mathbf{84.9}$ & $55.7$ & $33.5$ & $\mathbf{61.9}$ & $\mathbf{92.1}$ & $\mathbf{93.8}$ & $\mathbf{75.5}$ \\
        \bottomrule
    \end{tabular}}
\end{table}

\begin{table}[!htbp]
    \centering
    \caption{Per-benchmark scores on the $5$ Video benchmarks: ZPPO students at four scales and the $27$B teacher. Bold marks the column maximum and is shared by every row that attains it. Benchmark abbreviations: \cref{sec:appendix:benchmarks}.}
    \label{tab:appendix:teacher_capability:video}
    \resizebox{0.6\linewidth}{!}{
    \setlength{\tabcolsep}{2mm}
    \renewcommand{\arraystretch}{1.1}
    \begin{tabular}{lcccccc}
        \toprule
        Model & MMVU & MVBench & VMME & VMME$^\text{S}$ & VMMMU & Avg \\
        \midrule
        Qwen3.5-0.8B $+\,$ZPPO & $42.5$ & $56.6$ & $53.9$ & $54.9$ & $56.4$ & $52.9$ \\
        Qwen3.5-2B   $+\,$ZPPO & $59.0$ & $62.7$ & $62.4$ & $63.9$ & $68.1$ & $63.2$ \\
        Qwen3.5-4B   $+\,$ZPPO & $70.6$ & $69.6$ & $68.5$ & $70.2$ & $80.2$ & $71.8$ \\
        Qwen3.5-9B   $+\,$ZPPO & $74.2$ & $71.7$ & $70.2$ & $72.8$ & $82.1$ & $74.2$ \\
        \midrule
        Qwen3.5-27B-FP8 & $\mathbf{75.7}$ & $\mathbf{74.6}$ & $\mathbf{76.1}$ & $\mathbf{76.0}$ & $\mathbf{84.7}$ & $\mathbf{77.4}$ \\
        \bottomrule
    \end{tabular}
    }
\end{table}

The $27$B teacher is above all four ZPPO students on every family average. BCQ requires the teacher to succeed where the student fails. At $9$B, where BCQ candidates are scarce (\cref{fig:system}a), ZPPO beats GRPO$^{\dagger}$ on every family (\cref{tab:main_4b9b,tab:generalization_4b9b}). In the teacher-size study (\cref{fig:teacher_scale}), the released Qwen3.5-$4$B or Qwen3.5-$9$B checkpoint replaces the $27$B teacher in supplying BCQ responses and compressing every candidate, and every other setting, including the $27$B judge, stays at its default. With the $4$B/$9$B/$27$B teacher, ZPPO's mean family gain over the untrained base is $+4.4$/$+5.1$/$+7.3$~pp at $0.8$B and $+2.8$/$+3.1$/$+4.3$~pp at $2$B, against $+4.0$ and $+2.7$~pp for GRPO$^{\dagger}\!+\!$NCQ with the $27$B teacher.

\section{Additional comparisons and robustness checks}
\label{sec:appendix:additional_comparisons}

This appendix collects nine further comparisons, and none of them changes the ranking. They
vary the budget and the replay (\cref{sec:appendix:budget}), the zero-variance method
(\cref{sec:appendix:zerovariance}), the candidate content
(\cref{sec:appendix:bcq_controls}), the judge
(\cref{sec:appendix:judges}), the teacher-side method
(\cref{sec:appendix:teacher_alternatives}), the reward
(\cref{sec:appendix:mmrl}), and the policy objective and the student backbone
(\cref{sec:appendix:backbones}). Apart from what each
subsection changes, all rows share the initialization, corpus, schedule, and evaluation
setup (\cref{tab:appendix:hyperparameters,sec:appendix:benchmarks}). A row that adds a cited method changes only
what that method changes (\cref{sec:related_full}). Most such rows change one part of the GRPO$^{\dagger}$ loss,
such as a term, a reward, or the objective. The self-distillation rows OPSD$^{\dagger}$ and
SDPO$^{\dagger}$ instead use their distillation loss in place of the RL loss, as in their
papers and as On-Distill$^{\dagger}$ does (\cref{sec:appendix:teacher_alternatives}).
Every family average covers the $10$ LLM, $16$ VLM, or $5$ Video benchmarks, except in
\cref{tab:appendix:judges} (the shared-template settings) and
\cref{tab:appendix:backbone_families} ($3$ LLM, $7$ VLM, and $3$ Video settings).

\subsection{Budget controls: group size, epochs, and replay}
\label[appendix]{sec:appendix:budget}

\paragraph{Group size and epochs.}
We sweep the budget of two baselines, GRPO$^{\dagger}$ and On-Distill$^{\dagger}$, and two controls, Hint
and Prefix (\cref{tab:appendix:budget}).
Each of the four varies the student group size $G_{\rm S}\!\in\!\{8,16\}$ and the number of
epochs over ZPPO-$77$K ($1$ or $2$), and ZPPO stays at its default $(G_{\rm S}\!=\!8$, $1$ epoch$)$.

\begin{table}[!htbp]
    \centering
    \caption{Group-size and epoch sweep. Each compared method varies the student group size $G_{\rm S}$ and the number of epochs over ZPPO-$77$K, and ZPPO stays at its default $(G_{\rm S}\!=\!8$, $1$ epoch$)$. The $(8,1)$ row of each block is that method's row in \cref{tab:main}, \cref{tab:generalization}, \cref{tab:hint_methods}, or \cref{tab:prefix_methods}.}
    \label{tab:appendix:budget}
    \resizebox{0.7\linewidth}{!}{
    \setlength{\tabcolsep}{2mm}
    \renewcommand{\arraystretch}{1.05}
    \begin{tabular}{llcccccc}
        \toprule
        \multirow{2}{*}{Method} & \multirow{2}{*}{$(G_{\rm S}$, epochs$)$} & \multicolumn{3}{c}{Qwen3.5-0.8B} & \multicolumn{3}{c}{Qwen3.5-2B} \\
        \cmidrule(lr){3-5}\cmidrule(lr){6-8}
                                &            & LLM             & VLM             & Video           & LLM             & VLM             & Video           \\
        \midrule
        \multirow{4}{*}{GRPO$^{\dagger}$}
                                & $(8,1)$    & $28.7$          & $45.5$          & $50.5$          & $47.4$          & $59.2$          & $61.9$          \\
                                & $(16,1)$   & $29.0$          & $46.0$          & $50.6$          & $47.6$          & $59.7$          & $62.0$          \\
                                & $(8,2)$    & $28.6$          & $45.6$          & $50.3$          & $47.2$          & $59.4$          & $61.7$          \\
                                & $(16,2)$   & $28.9$          & $45.7$          & $50.7$          & $47.4$          & $59.6$          & $61.9$          \\
        \cdashline{1-8}\noalign{\vskip 0.5ex}
        \multirow{4}{*}{On-Distill$^{\dagger}$}
                                & $(8,1)$    & $23.2$          & $42.3$          & $45.8$          & $43.8$          & $58.0$          & $59.2$          \\
                                & $(16,1)$   & $23.6$          & $42.8$          & $46.0$          & $44.0$          & $58.4$          & $59.2$          \\
                                & $(8,2)$    & $22.4$          & $42.2$          & $45.1$          & $43.0$          & $57.9$          & $58.6$          \\
                                & $(16,2)$   & $22.9$          & $42.5$          & $45.4$          & $43.4$          & $58.2$          & $58.8$          \\
        \cdashline{1-8}\noalign{\vskip 0.5ex}
        \multirow{4}{*}{GRPO$^{\dagger}\!+\!$Hint}
                                & $(8,1)$    & $28.8$          & $47.2$          & $50.0$          & $47.6$          & $60.4$          & $61.6$          \\
                                & $(16,1)$   & $29.1$          & $47.6$          & $50.3$          & $47.9$          & $60.9$          & $61.8$          \\
                                & $(8,2)$    & $28.1$          & $47.3$          & $49.4$          & $47.0$          & $60.5$          & $61.1$          \\
                                & $(16,2)$   & $28.6$          & $47.4$          & $49.7$          & $47.4$          & $60.6$          & $61.3$          \\
        \cdashline{1-8}\noalign{\vskip 0.5ex}
        \multirow{4}{*}{GRPO$^{\dagger}\!+\!$Prefix}
                                & $(8,1)$    & $27.1$          & $45.5$          & $49.1$          & $46.3$          & $59.5$          & $61.0$          \\
                                & $(16,1)$   & $27.5$          & $45.8$          & $49.1$          & $46.7$          & $59.8$          & $61.1$          \\
                                & $(8,2)$    & $26.0$          & $45.2$          & $48.1$          & $45.3$          & $59.3$          & $60.1$          \\
                                & $(16,2)$   & $26.5$          & $45.4$          & $48.3$          & $45.8$          & $59.4$          & $60.4$          \\
        \midrule
        \rowcolor{colorful}
        ZPPO                    & $(8,1)$    & $\mathbf{33.1}$ & $\mathbf{50.3}$ & $\mathbf{52.9}$ & $\mathbf{50.4}$ & $\mathbf{62.0}$ & $\mathbf{63.2}$ \\
        \bottomrule
    \end{tabular}
    }
\end{table}

\textbf{(i)~Extra rollouts do not close the gap.} No swept row reaches ZPPO, including the
four that double $G_{\rm S}$ at one epoch.
\textbf{(ii)~More compute does not close the gap.} With $64$ H100 GPUs per run,
GRPO$^{\dagger}$ with doubled rollouts, doubled epochs, or both takes longer than ZPPO in
wall-clock time, as does On-Distill$^{\dagger}$ with doubled epochs
(\cref{tab:appendix:compute}), yet none of these runs reaches ZPPO. DAPO$^{\dagger}$, which
adds DAPO's dynamic sampling to GRPO$^{\dagger}$ and runs two epochs, also takes longer than
ZPPO and stays below it in every column of
\cref{tab:compute_matched}.

\paragraph{Extra replay from GRPO$^{\dagger}$'s own buffer.}
\label[appendix]{sec:appendix:replay_matched}
We raise GRPO$^{\dagger}$'s replay fraction $\rho_{\rm replay}$ from $0.25$ to $0.50$
(\cref{tab:appendix:replay_matched}), using its own buffer and no teacher. A
GRPO$^{\dagger}$ step then runs $384$ new and up to $192$ replayed groups, and a ZPPO step
at $\rho_{\rm replay}\!=\!0.25$ runs $384$ new, up to $96$ replayed, and up to $96$
reformulated groups (\cref{sec:appendix:algorithm}), so both are capped at $576$ groups
per step.

\begin{table}[!htbp]
    \centering
    \caption{Replay control: GRPO$^{\dagger}$ at two replay fractions $\rho_{\rm replay}$. At $\rho_{\rm replay}\!=\!0.50$, a GRPO$^{\dagger}$ step is capped at $576$ groups, the same cap as ZPPO at $0.25$. Bold marks the column maximum.}
    \label{tab:appendix:replay_matched}
    \resizebox{0.6\linewidth}{!}{
    \setlength{\tabcolsep}{2.5mm}
    \renewcommand{\arraystretch}{1.05}
    \begin{tabular}{llcccccc}
        \toprule
        \multirow{2}{*}{Method} & \multirow{2}{*}{$\rho_{\rm replay}$} & \multicolumn{3}{c}{Qwen3.5-0.8B} & \multicolumn{3}{c}{Qwen3.5-2B} \\
        \cmidrule(lr){3-5}\cmidrule(lr){6-8}
                                &            & LLM  & VLM  & Video & LLM  & VLM  & Video \\
        \midrule
        GRPO$^{\dagger}$        & $0.25$     & $28.7$ & $45.5$ & $50.5$ & $47.4$ & $59.2$ & $61.9$ \\
        GRPO$^{\dagger}$        & $0.50$     & $29.1$ & $45.7$ & $50.4$ & $47.6$ & $59.9$ & $62.1$ \\
        \midrule
        \rowcolor{colorful}
        ZPPO                    & $0.25$     & $\mathbf{33.1}$ & $\mathbf{50.3}$ & $\mathbf{52.9}$ & $\mathbf{50.4}$ & $\mathbf{62.0}$ & $\mathbf{63.2}$ \\
        \bottomrule
    \end{tabular}
    }
\end{table}

\textbf{Extra replay does not close the gap.} Raising $\rho_{\rm replay}$ to $0.50$ leaves
GRPO$^{\dagger}$ below ZPPO in all six cells.

\subsection{Zero-variance methods}
\label[appendix]{sec:appendix:zerovariance}

RL-ZVP~\citep{le2026no}, LENS~\citep{feng2025don}, ReLIFT~\citep{ma2026learning}, and
RSTG~\citep{han2026distill} start from the same problem as ZPPO: an all-wrong group has
zero advantage. They change the loss or the target, never the prompt
(\cref{sec:related_full}). We add each method to the GRPO$^{\dagger}$
pipeline (\cref{tab:appendix:zerovariance}; RL-ZVP, LENS, and ReLIFT also appear in
\cref{tab:mistakes}).

RL-ZVP and LENS are \emph{teacher-free}: \textbf{RL-ZVP} shapes the zero advantage from
the student's token-level entropy, and \textbf{LENS} penalizes a wrong answer more as the
student's confidence in it grows. ReLIFT and RSTG instead put teacher targets in the loss:
\textbf{ReLIFT} interleaves online fine-tuning on teacher solutions to the hardest
questions, and \textbf{RSTG} distills from the teacher on all-wrong groups and fine-tunes on
teacher-generated correct trajectories.

\begin{table}[!htbp]
    \centering
    \caption{Zero-variance methods on the GRPO$^{\dagger}$ pipeline. Each adds a signal or teacher target and leaves the prompt unchanged; $^{\dagger}$ denotes augmentation with the prompt replay buffer.}
    \label{tab:appendix:zerovariance}
    \resizebox{0.7\linewidth}{!}{
    \setlength{\tabcolsep}{2.5mm}
    \renewcommand{\arraystretch}{1.05}
    \begin{tabular}{lccccccc}
        \toprule
        \multirow{2}{*}{Method} & \multirow{2}{*}{Teacher target in loss} & \multicolumn{3}{c}{Qwen3.5-0.8B} & \multicolumn{3}{c}{Qwen3.5-2B} \\
        \cmidrule(lr){3-5}\cmidrule(lr){6-8}
                 &     & LLM  & VLM  & Video & LLM  & VLM  & Video \\
        \midrule
        Base                     & --      & $25.2$ & $41.0$ & $48.3$ & $45.3$ & $56.8$ & $60.6$ \\
        GRPO$^{\dagger}$         & No      & $28.7$ & $45.5$ & $50.5$ & $47.4$ & $59.2$ & $61.9$ \\
        On-Distill$^{\dagger}$   & Yes     & $23.2$ & $42.3$ & $45.8$ & $43.8$ & $58.0$ & $59.2$ \\
        \cdashline{1-8}\noalign{\vskip 0.3ex}
        RL-ZVP$^{\dagger}$       & No      & $29.0$ & $45.8$ & $50.7$ & $47.5$ & $59.5$ & $62.0$ \\
        LENS$^{\dagger}$         & No      & $29.3$ & $46.1$ & $50.9$ & $47.7$ & $59.8$ & $62.2$ \\
        ReLIFT$^{\dagger}$       & Yes     & $28.1$ & $46.9$ & $49.6$ & $47.1$ & $60.4$ & $61.5$ \\
        RSTG$^{\dagger}$         & Yes     & $27.5$ & $47.6$ & $49.1$ & $46.8$ & $60.7$ & $61.0$ \\
        \cdashline{1-8}\noalign{\vskip 0.3ex}
        \rowcolor{colorful}
        ZPPO                     & No      & $\mathbf{33.1}$ & $\mathbf{50.3}$ & $\mathbf{52.9}$ & $\mathbf{50.4}$ & $\mathbf{62.0}$ & $\mathbf{63.2}$ \\
        \bottomrule
    \end{tabular}
    }
\end{table}

\textbf{None of the four closes the gap.} The two teacher-free rows stay close to
GRPO$^{\dagger}$. The two teacher-target rows are
above it on VLM but below it on LLM and Video in all $8$ cells. ZPPO stays above all four
rows in every column.

\subsection{BCQ candidate-content controls}
\label[appendix]{sec:appendix:bcq_controls}

\Cref{tab:transfer} varies the BCQ candidate block on top of GRPO$^{\dagger}$ at $2$B, and
\cref{sec:experiments:discussion}(iv) reads the result. By default, the block holds a
shuffled, anonymized pair of teacher-compressed candidates, one correct teacher response and
one wrong student response, each ending in its boxed answer.
\emph{Correct candidate only} keeps only the teacher's correct response and drops the pair
instruction. \emph{Answer-removed} asks the teacher, before compression, to delete every sentence
related to the final answer from each source rollout, inside the \texttt{<think>} block as well as after
it, so the candidates keep their reasoning but drop the answer. \emph{Final answer
only} drops everything inside the \texttt{<think>} tags of each source rollout, so each
candidate keeps only its final answer. \emph{Non-anonymized} keeps the default
pair but tags one response as the teacher's solution and the other as the student's, so the
student no longer has to tell them apart. \emph{Unrelated candidates} shuffles the candidate
blocks built in the same batch and gives each question a block built for another question,
never its own, so the added block is unrelated to the question. Apart from these changes, the
template of \cref{fig:prompt-bcq} and the compression prompt stay at their defaults, except
that \emph{answer-removed} candidates end without the boxed final answer that the
compression prompt otherwise requires.

\paragraph{Measuring Copy.}
\emph{Copy} is the share of rollouts that GPT-6 Astra judges as copied from the candidate
block (\cref{fig:prompt-copy}): most of the reasoning comes from one candidate with at most
light rewording, or the final answer is taken without derivation. A rollout that reaches a
candidate's answer through its own derivation does not count, and the judge does not grade
correctness. For each row, we sample $500$ rollouts on the candidate-included prompt from
the last ten logged steps, regardless of correctness, and the ZPPO row uses its BCQ
rollouts; \emph{Prompt acc} is the accuracy on these prompts over the first and the last
ten logged steps. The judge sees the question, the candidates without source tags, and the
full rollout, but not the variant. An output that does not parse is retried once and otherwise
dropped.

\begin{figure}[!htbp]
\begin{prompttemplatefig}
\small
\emph{You are auditing whether a model's response was copied from reference candidates that appeared in its prompt.}

\textbf{\emph{Inputs:}}
\begin{itemize}[leftmargin=12pt,topsep=1pt,itemsep=1pt]
    \item \emph{QUESTION: the original question. It may have come with an image that you cannot see.}
    \item \emph{CANDIDATES: the reference responses appended to the model's prompt, numbered from 1. Some candidates may contain only a final answer, and some may be unrelated to the question.}
    \item \emph{RESPONSE: what the model wrote after reading the candidates, including any reasoning section such as text inside \texttt{<think>} tags.}
\end{itemize}

\emph{Label the RESPONSE as COPY if either condition holds.}
\begin{itemize}[leftmargin=16pt,topsep=1pt,itemsep=1pt]
    \item[(a)] \emph{Reasoning copy: most of the RESPONSE's substantive reasoning (its key steps, quantities, equations, visual observations, and intermediate conclusions) comes from a single candidate, in essentially the same order, with at most light rewording, reordering, or shortening.}
    \item[(b)] \emph{Answer copy: the RESPONSE takes a candidate's final answer without deriving it, for example by declaring a candidate correct and restating its answer, or by giving only a trivial justification.}
\end{itemize}

\emph{Otherwise, label it OWN. The RESPONSE is OWN when it:}
\begin{itemize}[leftmargin=12pt,topsep=1pt,itemsep=1pt]
    \item \emph{reaches the same final answer as a candidate through its own derivation, even if it mentions, compares, or checks the candidates;}
    \item \emph{shares only content that any solution would contain, such as restating the question, standard formulas or definitions, values given in the question, or the answer format;}
    \item \emph{reuses a few phrases or a single step but works out the decisive reasoning itself, for example by finding an error in a candidate or adding a step that settles the answer.}
\end{itemize}

\emph{Judge only how the RESPONSE was produced, not whether it is correct. Make no assumption about which candidate is correct.}

\emph{Output strict JSON and nothing else, in this format:}\\
\texttt{\{"label": "<COPY|OWN>", "source": <integer>, "evidence": "<string>"\}}
\begin{itemize}[leftmargin=12pt,topsep=1pt,itemsep=1pt]
    \item \emph{source: the 1-based index of the copied candidate if label is COPY, otherwise 0.}
    \item \emph{evidence: a quote of at most 30 words from the RESPONSE that supports the label.}
\end{itemize}

\textbf{\emph{QUESTION:}}\\
\emph{\texttt{\{\{ question \}\}}}

\textbf{\emph{CANDIDATES:}}\\
\emph{\texttt{\{\{ candidates \}\}}}

\textbf{\emph{RESPONSE:}}\\
\emph{\texttt{\{\{ response \}\}}}
\end{prompttemplatefig}
\caption{Copy-judge prompt for the \emph{Copy} column of \cref{tab:transfer}. The \texttt{\{\{\,\dots\,\}\}} fields are filled at call time, and \texttt{\{\{ candidates \}\}} lists each candidate as ``Candidate~$k$:'' with any source tag removed.}
\label{fig:prompt-copy}
\end{figure}

\paragraph{What the controls show.}
Of the six candidate blocks, only the default BCQ block and \emph{answer-removed} are above
GRPO$^{\dagger}$ on all three families, and both keep Copy low. \emph{Correct candidate
only} and \emph{non-anonymized} leave the student nothing to tell apart: most of their
rollouts are judged as copies, and both fall below GRPO$^{\dagger}$ on all three families.
Bare answers (\emph{final answer only}) and unrelated context (\emph{unrelated
candidates}) are not above GRPO$^{\dagger}$ on all three families either. The Hint and Prefix controls
(\cref{tab:hint_methods,tab:prefix_methods}) also give a correct cue or opening with no
wrong alternative to weigh it against, and at $2$B neither is above GRPO$^{\dagger}$ on
all three families.

\subsection{Robustness to the judge model}
\label[appendix]{sec:appendix:judges}

The judge for free-form answers is Qwen3.5-$27$B-FP8 (\cref{sec:appendix:benchmarks}). We
keep the scored student responses fixed and change only the judge on the settings that use
the shared template of \cref{fig:prompt-judge-shared}: $8/10$ LLM, $11/16$ VLM, and $4/5$
Video settings. The other eight settings ($3$ deterministic, $5$ benchmark-specific) do not
use this template. \Cref{tab:appendix:judges} recomputes each family average over the swept settings
under the default judge and seven others.

\begin{table}[!htbp]
    \centering
    \caption{Robustness to the judge model. Each average covers the $8/10$ LLM, $11/16$ VLM, and $4/5$ Video settings scored with the shared judge template. Only the judge changes across rows, and the scored responses stay fixed. The first row of each block is the default judge, and we query the other seven through their providers' APIs.}
    \label{tab:appendix:judges}
    \resizebox{\linewidth}{!}{
    \setlength{\tabcolsep}{1.8mm}
    \renewcommand{\arraystretch}{1.05}
    \begin{tabular}{lccccccccc}
        \toprule
        \multirow{2}{*}{Judge} & \multicolumn{3}{c}{ZPPO} & \multicolumn{3}{c}{GRPO$^{\dagger}$} & \multicolumn{3}{c}{On-Distill$^{\dagger}$} \\
        \cmidrule(lr){2-4}\cmidrule(lr){5-7}\cmidrule(lr){8-10}
                                    & LLM             & VLM             & Video           & LLM             & VLM             & Video           & LLM             & VLM             & Video           \\
        \midrule
        \multicolumn{10}{l}{\emph{Qwen3.5-0.8B student}} \\
        Qwen3.5-27B-FP8 (default)   & $37.2$          & $49.4$          & $55.5$          & $32.8$          & $43.5$          & $52.8$          & $26.8$          & $40.4$          & $48.1$          \\
        Gemini 3.1 Pro              & $37.0$          & $48.8$          & $55.6$          & $32.6$          & $43.0$          & $52.7$          & $26.5$          & $39.6$          & $48.1$          \\
        Gemini 3.5 Flash            & $38.0$          & $50.0$          & $55.9$          & $33.4$          & $43.9$          & $53.2$          & $27.7$          & $40.9$          & $48.7$          \\
        Claude Sonnet 5             & $37.5$          & $48.7$          & $55.7$          & $33.1$          & $43.0$          & $53.0$          & $27.1$          & $39.5$          & $48.3$          \\
        GPT-5.6 Luna                & $37.6$          & $49.5$          & $55.6$          & $33.3$          & $43.6$          & $52.9$          & $27.2$          & $40.6$          & $48.2$          \\
        GPT-5 Nano                  & $37.8$          & $49.8$          & $55.9$          & $33.4$          & $43.7$          & $53.3$          & $27.5$          & $41.0$          & $48.6$          \\
        Kimi-K2.6                   & $36.8$          & $48.6$          & $55.4$          & $32.5$          & $42.8$          & $52.6$          & $26.1$          & $39.4$          & $47.8$          \\
        GLM 5.2                     & $37.3$          & $48.9$          & $55.8$          & $32.9$          & $43.1$          & $53.2$          & $26.9$          & $40.0$          & $48.4$          \\
        \cdashline{1-10}\noalign{\vskip 0.4ex}
        Mean $\pm$ pop. std         & $37.4$\tiny{$\pm0.4$} & $49.2$\tiny{$\pm0.5$} & $55.7$\tiny{$\pm0.2$} & $33.0$\tiny{$\pm0.3$} & $43.3$\tiny{$\pm0.4$} & $53.0$\tiny{$\pm0.2$} & $27.0$\tiny{$\pm0.5$} & $40.2$\tiny{$\pm0.6$} & $48.3$\tiny{$\pm0.3$} \\
        \midrule
        \multicolumn{10}{l}{\emph{Qwen3.5-2B student}} \\
        Qwen3.5-27B-FP8 (default)   & $57.3$          & $61.9$          & $64.3$          & $54.0$          & $58.7$          & $63.2$          & $50.0$          & $57.8$          & $60.9$          \\
        Gemini 3.1 Pro              & $56.7$          & $61.7$          & $63.9$          & $53.5$          & $58.6$          & $62.8$          & $49.3$          & $57.4$          & $60.4$          \\
        Gemini 3.5 Flash            & $57.9$          & $62.6$          & $64.5$          & $54.4$          & $59.3$          & $63.3$          & $50.6$          & $58.6$          & $61.2$          \\
        Claude Sonnet 5             & $56.5$          & $61.8$          & $63.8$          & $53.3$          & $58.7$          & $62.7$          & $49.1$          & $57.5$          & $60.4$          \\
        GPT-5.6 Luna                & $57.4$          & $62.4$          & $64.2$          & $54.1$          & $59.1$          & $63.1$          & $50.1$          & $58.3$          & $60.8$          \\
        GPT-5 Nano                  & $57.8$          & $62.5$          & $64.4$          & $54.4$          & $59.2$          & $63.3$          & $50.6$          & $58.6$          & $61.1$          \\
        Kimi-K2.6                   & $56.4$          & $61.6$          & $63.7$          & $53.3$          & $58.5$          & $62.7$          & $49.0$          & $57.3$          & $60.0$          \\
        GLM 5.2                     & $56.8$          & $62.3$          & $64.0$          & $53.6$          & $59.0$          & $62.9$          & $49.5$          & $58.2$          & $60.6$          \\
        \cdashline{1-10}\noalign{\vskip 0.4ex}
        Mean $\pm$ pop. std         & $57.1$\tiny{$\pm0.5$} & $62.1$\tiny{$\pm0.4$} & $64.1$\tiny{$\pm0.3$} & $53.8$\tiny{$\pm0.4$} & $58.9$\tiny{$\pm0.3$} & $63.0$\tiny{$\pm0.2$} & $49.8$\tiny{$\pm0.6$} & $58.0$\tiny{$\pm0.5$} & $60.7$\tiny{$\pm0.4$} \\
        \bottomrule
    \end{tabular}
    }
\end{table}

\textbf{All eight judges keep the ordering.} Every judge gives ZPPO $>$
GRPO$^{\dagger}$ $>$ On-Distill$^{\dagger}$ on every family at both scales. Outside the sweep, ZPPO is also above
GRPO$^{\dagger}$ in all $20$ (scale, setting) cells of the five benchmark-specific
settings and in all $12$ cells of the three deterministic settings
(\cref{tab:main,tab:generalization,sec:appendix:main_tables}).

\subsection{Teacher-side alternatives: self-distillation and hybrid RL$+$distillation}
\label[appendix]{sec:appendix:teacher_alternatives}

\Cref{tab:appendix:channel} groups its rows by where teacher knowledge enters training.
In the two self-teacher rows, the teacher is the student itself, given privileged context
that the student's prompt lacks: a compressed correct rollout of the frozen $27$B teacher in
\textbf{OPSD}~\citep{zhao2026selfdistilled} and rollout feedback in
\textbf{SDPO}~\citep{hubotter2026reinforcement}. Both distill this teacher's per-token
distributions on the student's own rollouts in place of the RL loss, as in their papers.
ZPPO instead puts its context in the student's own prompt and keeps the RL loss.
Like On-Distill$^{\dagger}$, three hybrid
RL$+$distillation rows feed the frozen $27$B teacher back into the gradient:
\textbf{KDRL}~\citep{xu2025kdrl} adds a reverse-KL term on the student's rollouts,
\textbf{RLKD}~\citep{xu2026rlkd} adds a reward from the teacher's reasoning structure, and
\textbf{RLAD}~\citep{zhang2026reinforcement} anchors its trust-region ratio to a mixture of
the old student policy and the teacher.

\begin{table}[!htbp]
    \centering
    \caption{Teacher-side methods on the GRPO$^{\dagger}$ pipeline, grouped by where teacher knowledge enters training; $^{\dagger}$ denotes augmentation with the prompt replay buffer. The self-teacher rows, as in their papers, and On-Distill$^{\dagger}$ use distillation in place of the RL loss, and the hybrid RL$+$distillation rows add a reverse-KL term or a reasoning-structure reward, or anchor the trust-region ratio to a student--teacher mixture. The untrained \emph{Base} and GRPO$^{\dagger}$ are the reference rows.}
    \label{tab:appendix:channel}
    \resizebox{0.6\linewidth}{!}{
    \setlength{\tabcolsep}{3mm}
    \renewcommand{\arraystretch}{1.05}
    \begin{tabular}{lcccccc}
        \toprule
                                     & \multicolumn{3}{c}{Qwen3.5-0.8B}                    & \multicolumn{3}{c}{Qwen3.5-2B}                      \\
        \cmidrule(lr){2-4}\cmidrule(lr){5-7}
        Method                       & LLM             & VLM             & Video           & LLM             & VLM             & Video           \\
        \midrule
        Base                         & $25.2$          & $41.0$          & $48.3$          & $45.3$          & $56.8$          & $60.6$          \\
        \cdashline{1-7}\noalign{\vskip 0.5ex}
        \multicolumn{7}{l}{\emph{Self-teacher}} \\
        $+$OPSD$^{\dagger}$          & $21.8$          & $41.2$          & $44.6$          & $42.4$          & $57.0$          & $58.1$          \\
        $+$SDPO$^{\dagger}$          & $22.1$          & $40.9$          & $44.9$          & $42.7$          & $56.7$          & $58.4$          \\
        \cdashline{1-7}\noalign{\vskip 0.5ex}
        \multicolumn{7}{l}{\emph{External-teacher target, term, or reward in the loss}} \\
        $+$On-Distill$^{\dagger}$    & $23.2$          & $42.3$          & $45.8$          & $43.8$          & $58.0$          & $59.2$          \\
        $+$KDRL$^{\dagger}$          & $27.8$          & $46.0$          & $49.8$          & $46.8$          & $60.0$          & $61.3$          \\
        $+$RLKD$^{\dagger}$          & $27.5$          & $45.7$          & $49.6$          & $46.5$          & $59.6$          & $61.1$          \\
        $+$RLAD$^{\dagger}$          & $29.0$          & $46.3$          & $50.7$          & $47.9$          & $60.3$          & $62.1$          \\
        \cdashline{1-7}\noalign{\vskip 0.5ex}
        \multicolumn{7}{l}{\emph{No teacher}} \\
        $+$GRPO$^{\dagger}$          & $28.7$          & $45.5$          & $50.5$          & $47.4$          & $59.2$          & $61.9$          \\
        \cdashline{1-7}\noalign{\vskip 0.5ex}
        \multicolumn{7}{l}{\emph{Teacher in the prompt}} \\
        \rowcolor{colorful}
        $+$ZPPO                      & $\mathbf{33.1}$ & $\mathbf{50.3}$ & $\mathbf{52.9}$ & $\mathbf{50.4}$ & $\mathbf{62.0}$ & $\mathbf{63.2}$ \\
        \bottomrule
    \end{tabular}
    }
\end{table}

ZPPO is above every row in every column.

\subsection{Alternative RL rewards: multimodal and dense process rewards}
\label[appendix]{sec:appendix:mmrl}

BCQ and NCQ change the prompt and leave the reward alone; \cref{tab:appendix:mmrl} instead
changes the reward or credit assignment on GRPO$^{\dagger}$. In the top block,
\textbf{Perception-R1}~\citep{xiao2026perception} adds a visual-perception reward scored by
the shared judge (\cref{sec:appendix:benchmarks}), \textbf{GRPO-CARE}~\citep{chen2026grpo} adds an adaptive consistency bonus, and
\textbf{SRPO}~\citep{jiang2026structured} weights the advantage by token role (perception
vs.\ reasoning). In the bottom block, four dense process rewards give partial credit:
\textbf{PRIME}~\citep{cui2025process} learns an implicit process reward model online from
outcome labels, \textbf{StepGRPO}~\citep{zhang2025r1vl} adds rule-based step-wise rewards,
\textbf{PROGRS}~\citep{rezaei2026process} ranks the wrong rollouts of each group with a
centered process bonus, and \textbf{ProcessThinker}~\citep{wu2026processthinker} scores
each step by the success rate of continuations sampled from it.

\begin{table}[!htbp]
    \centering
    \caption{Reward and credit-assignment methods on the GRPO$^{\dagger}$ pipeline. The top block adds a perception reward, a consistency bonus, or role-aware token weights. The bottom block adds a dense process reward beside the sparse correctness reward. $^{\dagger}$ denotes augmentation with the prompt replay buffer.}
    \label{tab:appendix:mmrl}
    \resizebox{0.6\linewidth}{!}{
    \setlength{\tabcolsep}{3mm}
    \renewcommand{\arraystretch}{1.05}
    \begin{tabular}{lcccccc}
        \toprule
                                     & \multicolumn{3}{c}{Qwen3.5-0.8B}                    & \multicolumn{3}{c}{Qwen3.5-2B}                      \\
        \cmidrule(lr){2-4}\cmidrule(lr){5-7}
        Method                       & LLM             & VLM             & Video           & LLM             & VLM             & Video           \\
        \midrule
        GRPO$^{\dagger}$             & $28.7$          & $45.5$          & $50.5$          & $47.4$          & $59.2$          & $61.9$          \\
        \cdashline{1-7}\noalign{\vskip 0.5ex}
        Perception-R1$^{\dagger}$    & $28.1$          & $47.0$          & $49.7$          & $46.5$          & $60.3$          & $61.0$          \\
        GRPO-CARE$^{\dagger}$        & $28.2$          & $46.2$          & $50.4$          & $46.8$          & $59.9$          & $61.7$          \\
        SRPO$^{\dagger}$             & $28.0$          & $47.2$          & $49.9$          & $46.6$          & $60.6$          & $61.2$          \\
        \cdashline{1-7}\noalign{\vskip 0.5ex}
        PRIME$^{\dagger}$            & $28.8$          & $48.6$          & $50.8$          & $48.1$          & $60.7$          & $61.8$          \\
        StepGRPO$^{\dagger}$         & $30.2$          & $47.2$          & $49.9$          & $47.3$          & $60.0$          & $61.5$          \\
        PROGRS$^{\dagger}$           & $29.4$          & $46.6$          & $50.0$          & $47.8$          & $59.5$          & $61.3$          \\
        ProcessThinker$^{\dagger}$   & $30.7$          & $47.6$          & $50.7$          & $47.5$          & $60.4$          & $61.9$          \\
        \midrule
        \rowcolor{colorful}
        ZPPO                         & $\mathbf{33.1}$ & $\mathbf{50.3}$ & $\mathbf{52.9}$ & $\mathbf{50.4}$ & $\mathbf{62.0}$ & $\mathbf{63.2}$ \\
        \bottomrule
    \end{tabular}
    }
\end{table}

The three rows of the top block are above GRPO$^{\dagger}$ on VLM at both scales but below
it on LLM and Video in all $12$ (method, scale, family) cells. ZPPO stays above all three rows in every column.

The four rows of the bottom block are also above GRPO$^{\dagger}$ on VLM in all $8$ cells
and, unlike the top block, on LLM in $7$ of $8$, while on Video they stay close to it. ZPPO stays above all four rows in every column.

\subsection{Alternative policy objectives and student backbones}
\label[appendix]{sec:appendix:backbones}

\paragraph{Policy objectives.}
We keep BCQ, NCQ, and the buffer fixed and change only the policy objective of Step~6 in
\cref{alg:zppo} (\cref{tab:appendix:backbones}).
\textbf{GSPO}~\citep{zheng2025groupsequencepolicyoptimization} uses one sequence-level
importance ratio per rollout, \textbf{CISPO}~\citep{chen2025minimax} clips the importance
weight instead of the update, and \textbf{DISPO}~\citep{karaman2026dispo} clips the
importance weight with separate bounds for correct and wrong responses. We measure
ZPPO's gain within each objective.

\begin{table}[!htbp]
    \centering
    \caption{ZPPO with GSPO~\citep{zheng2025groupsequencepolicyoptimization}, CISPO~\citep{chen2025minimax}, or DISPO~\citep{karaman2026dispo} as the Step~6 objective, with BCQ, NCQ, and the prompt replay buffer fixed. $\Delta$ is ZPPO minus the same objective without BCQ and NCQ, and $^{\dagger}$ denotes augmentation with the prompt replay buffer.}
    \label{tab:appendix:backbones}
    \resizebox{0.8\linewidth}{!}{
    \setlength{\tabcolsep}{2mm}
    \renewcommand{\arraystretch}{1.05}
    \begin{tabular}{llccccccccc}
        \toprule
        \multirow{2}{*}{Scale} & \multirow{2}{*}{Objective} & \multicolumn{3}{c}{Objective alone} & \multicolumn{3}{c}{ZPPO on that objective} & \multicolumn{3}{c}{$\Delta$} \\
        \cmidrule(lr){3-5}\cmidrule(lr){6-8}\cmidrule(lr){9-11}
                               &                          & LLM    & VLM    & Video  & LLM    & VLM    & Video  & LLM        & VLM        & Video      \\
        \midrule
        \multirow{4}{*}{$0.8$B}
                               & GRPO$^{\dagger}$ (default) & $28.7$ & $45.5$ & $50.5$ & $33.1$ & $50.3$ & $52.9$ & \dpos{4.4} & \dpos{4.8} & \dpos{2.4} \\
                               & GSPO$^{\dagger}$         & $29.4$ & $46.1$ & $51.5$ & $33.7$ & $50.8$ & $53.9$ & \dpos{4.3} & \dpos{4.7} & \dpos{2.4} \\
                               & CISPO$^{\dagger}$        & $29.9$ & $46.3$ & $51.0$ & $34.5$ & $51.3$ & $53.5$ & \dpos{4.6} & \dpos{5.0} & \dpos{2.5} \\
                               & DISPO$^{\dagger}$        & $29.7$ & $46.7$ & $51.2$ & $34.2$ & $51.4$ & $53.8$ & \dpos{4.5} & \dpos{4.7} & \dpos{2.6} \\
        \midrule
        \multirow{4}{*}{$2$B}
                               & GRPO$^{\dagger}$ (default) & $47.4$ & $59.2$ & $61.9$ & $50.4$ & $62.0$ & $63.2$ & \dpos{3.0} & \dpos{2.8} & \dpos{1.3} \\
                               & GSPO$^{\dagger}$         & $48.1$ & $60.0$ & $62.9$ & $51.0$ & $62.7$ & $64.1$ & \dpos{2.9} & \dpos{2.7} & \dpos{1.2} \\
                               & CISPO$^{\dagger}$        & $48.7$ & $60.2$ & $62.5$ & $51.9$ & $63.1$ & $63.9$ & \dpos{3.2} & \dpos{2.9} & \dpos{1.4} \\
                               & DISPO$^{\dagger}$        & $48.4$ & $60.6$ & $62.7$ & $51.4$ & $63.2$ & $64.2$ & \dpos{3.0} & \dpos{2.6} & \dpos{1.5} \\
        \bottomrule
    \end{tabular}
    }
\end{table}

ZPPO gives a positive $\Delta$ in all $24$ (objective, scale, family) cells.

\paragraph{Two student backbones outside Qwen3.5.}
\label[appendix]{sec:appendix:backbone_families}
We run ZPPO on Ovis2.5-2B~\citep{lu2025ovis2} and InternVL3.5-2B~\citep{wang2025internvl3},
each with a larger teacher from its own family (Ovis2.5-$9$B and InternVL3.5-$38$B,
respectively; \cref{tab:appendix:backbone_families}). Both students differ from the Qwen3.5 student in the vision encoder, the modality connector, and the
training recipe. Both released checkpoints are already RL-tuned. Each student uses its
released decoding defaults:
Ovis2.5-2B uses its Hugging Face generation config (\texttt{temperature=0.6}, \texttt{top-p=0.95}, \texttt{top-k=20},
\texttt{repetition-penalty=1.05}), and InternVL3.5-2B uses its recommended thinking mode
(its thinking system prompt and \texttt{temperature=0.6}); both use
\texttt{max-new-tokens=12288}. Benchmarks and scoring follow
\cref{sec:appendix:benchmarks}.

\begin{table}[!htbp]
    \centering
    \caption{ZPPO on Ovis2.5-2B~\citep{lu2025ovis2} and InternVL3.5-2B~\citep{wang2025internvl3} with Ovis2.5-$9$B and InternVL3.5-$38$B teachers, respectively, over $3$ LLM, $7$ VLM, and $3$ Video settings (abbreviations in \cref{sec:appendix:benchmarks}). Bold marks the column maximum inside each backbone block, and $\Delta$ is ZPPO minus GRPO$^{\dagger}$.}
    \label{tab:appendix:backbone_families}
    \resizebox{\linewidth}{!}{
    \setlength{\tabcolsep}{1mm}
    \renewcommand{\arraystretch}{1.05}
    \begin{tabular}{ll*{16}{c}}
        \toprule
        & & \multicolumn{4}{c}{3 LLM Benchmarks} & \multicolumn{8}{c}{7 VLM Benchmarks} & \multicolumn{4}{c}{3 Video Benchmarks} \\
        \cmidrule(lr){3-6}\cmidrule(lr){7-14}\cmidrule(lr){15-18}
        Backbone & Method & AIME25 & CEval & MMLU & Avg & AI2D & DynaM & MVerse & MVision & MVista & MM-Vet & WeMath & Avg & MVBench & VMME & VMME$^\text{S}$ & Avg \\
        \midrule
        \multirow{4}{*}{Ovis2.5-2B}
                 & Base                & $40.0$ & $60.3$ & $60.9$ & \cellcolor{gray!10}$53.7$ & $85.5$ & $26.5$ & $64.8$ & $37.4$ & $81.4$ & $67.9$ & $50.8$ & \cellcolor{gray!10}$59.2$ & $66.7$ & $60.2$ & $63.6$ & \cellcolor{gray!10}$63.5$ \\
                 & $+$GRPO$^{\dagger}$ & $43.3$ & $61.2$ & $62.2$ & \cellcolor{gray!10}$55.6$ & $85.2$ & $29.6$ & $67.0$ & $40.7$ & $81.9$ & $68.8$ & $53.1$ & \cellcolor{gray!10}$60.9$ & $68.3$ & $62.3$ & $65.3$ & \cellcolor{gray!10}$65.3$ \\
                 & \cellcolor{colorful}$+$ZPPO & \cellcolor{colorful}$\mathbf{46.7}$ & \cellcolor{colorful}$\mathbf{62.6}$ & \cellcolor{colorful}$\mathbf{64.1}$ & \cellcolor{gray!10}$\mathbf{57.8}$ & \cellcolor{colorful}$\mathbf{87.6}$ & \cellcolor{colorful}$\mathbf{32.8}$ & \cellcolor{colorful}$\mathbf{69.3}$ & \cellcolor{colorful}$\mathbf{43.5}$ & \cellcolor{colorful}$\mathbf{83.2}$ & \cellcolor{colorful}$\mathbf{71.3}$ & \cellcolor{colorful}$\mathbf{56.1}$ & \cellcolor{gray!10}$\mathbf{63.4}$ & \cellcolor{colorful}$\mathbf{70.2}$ & \cellcolor{colorful}$\mathbf{64.2}$ & \cellcolor{colorful}$\mathbf{66.9}$ & \cellcolor{gray!10}$\mathbf{67.1}$ \\
        \cdashline{2-18}\noalign{\vskip 0.3ex}
                 & $\Delta$            & \dpos{3.4} & \dpos{1.4} & \dpos{1.9} & \cellcolor{gray!10}\dpos{2.2} & \dpos{2.4} & \dpos{3.2} & \dpos{2.3} & \dpos{2.8} & \dpos{1.3} & \dpos{2.5} & \dpos{3.0} & \cellcolor{gray!10}\dpos{2.5} & \dpos{1.9} & \dpos{1.9} & \dpos{1.6} & \cellcolor{gray!10}\dpos{1.8} \\
        \midrule
        \multirow{4}{*}{InternVL3.5-2B}
                 & Base                & $43.3$ & $61.0$ & $61.9$ & \cellcolor{gray!10}$55.4$ & $78.8$ & $31.5$ & $53.4$ & $42.8$ & $71.8$ & $71.7$ & $48.5$ & \cellcolor{gray!10}$56.9$ & $65.9$ & $58.4$ & $61.9$ & \cellcolor{gray!10}$62.1$ \\
                 & $+$GRPO$^{\dagger}$ & $46.7$ & $62.1$ & $63.1$ & \cellcolor{gray!10}$57.3$ & $80.2$ & $34.7$ & $56.0$ & $45.7$ & $71.4$ & $72.8$ & $50.8$ & \cellcolor{gray!10}$58.8$ & $67.7$ & $59.9$ & $63.2$ & \cellcolor{gray!10}$63.6$ \\
                 & \cellcolor{colorful}$+$ZPPO & \cellcolor{colorful}$\mathbf{50.0}$ & \cellcolor{colorful}$\mathbf{63.5}$ & \cellcolor{colorful}$\mathbf{64.7}$ & \cellcolor{gray!10}$\mathbf{59.4}$ & \cellcolor{colorful}$\mathbf{81.5}$ & \cellcolor{colorful}$\mathbf{38.6}$ & \cellcolor{colorful}$\mathbf{58.8}$ & \cellcolor{colorful}$\mathbf{49.4}$ & \cellcolor{colorful}$\mathbf{74.0}$ & \cellcolor{colorful}$\mathbf{74.3}$ & \cellcolor{colorful}$\mathbf{53.9}$ & \cellcolor{gray!10}$\mathbf{61.5}$ & \cellcolor{colorful}$\mathbf{69.3}$ & \cellcolor{colorful}$\mathbf{62.0}$ & \cellcolor{colorful}$\mathbf{64.9}$ & \cellcolor{gray!10}$\mathbf{65.4}$ \\
        \cdashline{2-18}\noalign{\vskip 0.3ex}
                 & $\Delta$            & \dpos{3.3} & \dpos{1.4} & \dpos{1.6} & \cellcolor{gray!10}\dpos{2.1} & \dpos{1.3} & \dpos{3.9} & \dpos{2.8} & \dpos{3.7} & \dpos{2.6} & \dpos{1.5} & \dpos{3.1} & \cellcolor{gray!10}\dpos{2.7} & \dpos{1.6} & \dpos{2.1} & \dpos{1.7} & \cellcolor{gray!10}\dpos{1.8} \\
        \bottomrule
    \end{tabular}
    }
\end{table}

On both backbones, ZPPO is above the base and GRPO$^{\dagger}$ on all $13$ settings.

\subsection{Scope beyond reasoning correctness}
\label{sec:appendix:scope_beyond}

ZPPO post-trains an already-built compact VLM and is orthogonal to how such small models
are first constructed, whether via architectural designs that compress visual--linguistic
computation into small
backbones~\citep{lee-etal-2024-collavo,10.1007/978-3-031-72967-6_16,NEURIPS2024_473a9a75,lee-etal-2024-trol,lee2024phantom}
or via large-to-small distillation of strong
teachers~\citep{10.1007/978-3-032-37281-9_9,lee2025building,yu2026hide,zhao2026we}, so it could in principle be
stacked on any of these as a later post-training stage. Several other axes are
deliberately out of scope. Extending the recipe to multi-step and agentic multimodal
reasoning~\citep{Lee_2026_CVPR_Recursive,kang2026agent,cho2026spatialclaw,kang2026mid,shi2026edgewise}
is open; we neither evaluate nor train for robustness, which carries its own substantial
literature on robust and causal
training~\citep{lee2021towards,NEURIPS2021_8e5e15c4,Lee_2022_CVPR,Kim_2023_CVPR,Lee_2023_ICCV};
we do not consider inference-time efficiency techniques such as visual-token
pruning~\citep{10.1007/978-3-032-37281-9_15} that are orthogonal to the training recipe; we do not target
conversational or social competence~\citep{leeenhancing}; we do not touch the upstream
perception stacks~\citep{lee2020training} that compact models ultimately depend on for
embodied deployment~\citep{yu2026switch,lee2026imitation}; and our evaluation targets single-turn reasoning accuracy rather than
complementary axes such as multi-sensor perception~\citep{yu2024spark,chung2024vision}, multi-turn
dialogue~\citep{Lee_2025_ICCV}, or iterative self-refinement~\citep{lee2026refinebench}.
Combining ZPPO with these complementary directions is left to future work.

\end{document}